\documentclass[11pt]{article}

\usepackage[final]{acl}

\usepackage{amsmath}
\usepackage{bm}
\usepackage{times}
\usepackage{latexsym}
\usepackage{booktabs}
\usepackage{multirow}
\usepackage[T1]{fontenc}
\usepackage[utf8]{inputenc}

\usepackage{microtype}

\usepackage{inconsolata}

\usepackage{graphicx}
\usepackage{comment}
\usepackage{subcaption}
\usepackage{alphalph}
\usepackage{placeins}
\usepackage{float}
\usepackage{caption}
\usepackage[htt]{hyphenat}  % allow line breaks at existing hyphens inside \texttt
\usepackage{soul}           % \hl for in-line highlighting in expert-annotated vignette (Appendix)
\definecolor{trigcol}{HTML}{F8E29B}
\definecolor{memcol}{HTML}{B9D8E5}
\definecolor{appcol}{HTML}{CDB5DA}
\definecolor{stratcol}{HTML}{C2E2BD}
\definecolor{threatcol}{HTML}{F3B6B0}
\newcommand{\hltrig}[1]{{\sethlcolor{trigcol}\hl{#1}}}
\newcommand{\hlmem}[1]{{\sethlcolor{memcol}\hl{#1}}}
\newcommand{\hlapp}[1]{{\sethlcolor{appcol}\hl{#1}}}
\newcommand{\hlstrat}[1]{{\sethlcolor{stratcol}\hl{#1}}}
\newcommand{\hlthreat}[1]{{\sethlcolor{threatcol}\hl{#1}}}

\title{Generating Clinical Vignettes that Preserve Cognitive Formulations}%Generating Personalized PTSD Vignettes Grounded in Cognitive Behavioural Models}

\author{
Amit Oren\textsuperscript{1} \quad
Nimrod Hertz-Palmor\textsuperscript{2} \quad
Dean Ariel\textsuperscript{3,4} \quad
Guy Laban\textsuperscript{1,5,6}\thanks{Corresponding author: \texttt{laban@bgu.ac.il}}
\\[0.6em]
\textsuperscript{1}Department of Industrial Engineering and Management, Ben-Gurion University of the Negev, Beer Sheva, Israel
\\
\textsuperscript{2}MRC Cognition and Brain Sciences Unit, University of Cambridge, United Kingdom
\\
\textsuperscript{3}Clalit Health Services, Israel
\\
\textsuperscript{4}School of Public Health, Tel Aviv University, Israel
\\
\textsuperscript{5}School of Brain Sciences and Cognition, Ben-Gurion University of the Negev, Beer Sheva, Israel
\\
\textsuperscript{6}Azrieli National Center for Autism and Neurodevelopment Research, Beer Sheva, Israel
}

\usepackage{listings}
\begin{document}
\maketitle
\begin{abstract}
Large language models can generate fluent clinical case vignettes, but fluency alone does not ensure fidelity to a specifiable clinical structure. We introduce \textsc{Forma}, a theory-grounded framework that compiles a cognitive model of a disorder into a directed weighted graph, samples a person-specific configuration of that graph, and validates whether the generated vignette preserves the specified components and causal links. We instantiate \textsc{Forma} on Posttraumatic Stress Disorder using the Ehlers and Clark cognitive model, generating 16{,}500 vignettes across 500 personas, 11 generation models, and three ablation conditions. Evaluation combines an external edge-recovery probe, two clinical experts, a scaled LLM judge, and a clinician user study with 100 licensed practitioners. The cognitive graph is recoverable from full-condition vignettes ($\mathrm{MCC}=+0.41$, $\mathrm{AUC}=0.70$) but not from zero-shot generation ($\mathrm{MCC}=+0.01$, $\mathrm{AUC}=0.50$). Experts rate full vignettes substantially higher than zero-shot alternatives, and clinicians perceive them to be human-written 85\% of the time, compared with 22\% for zero-shot. \textsc{Forma} also reduces demographic disparity in perceived quality by 1.5--7$\times$. These results show that cognitive formulation can serve as an auditable specification for scalable synthetic clinical text generation.
A repository with the data and code is available online: \small{\url{https://github.com/Amit-Oren/FORMA}}.
\end{abstract}

\section{Introduction}

Clinical case vignettes, short narrative descriptions of a patient's presentation, are a workhorse of mental-health training, research, and assessment \citep{evans2015vignette,sheringham2021use}, used to teach diagnostic reasoning \citep{rosenquist2000best,st2021developing}, standardise experimental stimuli, and benchmark the clinical capabilities of language models \citep{benoit2023chatgpt,yanagita2024can}. Across all three uses the same constraint binds: vignettes must look like real clinical material, in numbers and along axes of variation that a hand-curated bank cannot deliver. Curated banks are small and skewed toward prototypical cases \citep{evans2015vignette,heverly1984constructing}; real clinical records resist redistribution at scale and rarely come annotated with the clinical structure researchers want to vary. Synthetic generation with large language models (LLMs) is the natural alternative \citep{frayling2024zero,reichenpfader2024simulating,bhate2023zero}, but a generated vignette is only useful insofar as it is \emph{faithful} to a specifiable clinical picture rather than fluent in the average. Given a diagnostic label, a competent LLM produces text that reads as plausible while averaging over the population of patients with the diagnosis and clustering around trauma-type templates rather than the individual processes that maintain the disorder.

We address this by grounding generation in a \emph{cognitive formulation} of the disorder: a standard tool in cognitive-behavioural therapy for case conceptualisation \citep{persons2008case} specifying a disorder by a directed, person-specific network of psychological components whose interactions explain how the disorder is maintained in this patient, rather than merely using a symptom checklist.  %they specify the disorder not as a symptom checklist but as a directed, person-specific network of psychological components (triggers, appraisals, memories, threat perceptions, coping strategies) whose interactions explain how the disorder is maintained in this patient. 
They are mechanistic, individualised, and clinically standard, and formal cognitive models with this structure exist for many disorders including depression \citep{beck1979cognitive}, OCD \citep{salkovskis1985obsessional}, social anxiety \citep{clark1995cognitive}, and Posttraumatic Stress Disorder (PTSD) \citep[][]{ehlers2000cognitive}, with recent work showing these models can be empirically recovered from patient data \citep{hertzpalmor2025quantifying}. Concretely, we compile the cognitive model of a disorder into a directed, weighted graph over its components, sample a concrete graph configuration for each persona, and use the configuration as a structural specification that the generating LLM must satisfy. A closed-loop validator then verifies that each specified component and directed causal link is realised in the text, inserting targeted patches where it is not. The resulting framework, \textsc{Forma}\footnote{From Latin \emph{forma} (``shape, structure''): the cognitive graph supplies the form, the LLM the realisation.}, is disorder-agnostic.

\begin{figure}[t]
    \centering
    \includegraphics[width=\columnwidth,trim=1 43 238 42,clip]{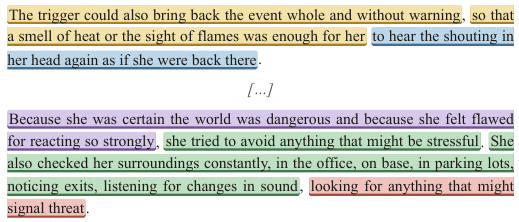}
    \\[0.5em]
    \includegraphics[width=0.55\columnwidth,trim=299 12 2 32,clip]{persona14_figure.pdf}
\caption{\footnotesize Excerpt from a full-condition vignette (Persona 14, \texttt{gpt-5.4}) and the
persona's specified cognitive graph. Highlighted spans mark clinician-identified
realizations of each Ehlers and Clark component. The complete, unabridged vignette is
given in Appendix~\ref{app:expert-annotation}.}
\label{fig:persona14-intro}
\end{figure}

We demonstrate it on PTSD, which has a well-validated cognitive theory \citep[][]{ehlers2000cognitive} operationalising the disorder as a graph over five components (traumatic memory, negative appraisals, current threat, re-experiencing triggers, maladaptive strategies) linked by directed causal edges. PTSD's symptom heterogeneity is enormous: the DSM-5 criteria admit over 600,000 distinct profiles \citep{galatzer2013636}, making it a stress test for whether a generation framework can produce \emph{individuated} rather than \emph{averaged} cases. Figure~\ref{fig:persona14-intro} shows an excerpt beside the graph it was generated from, clinician-marked by component; the full case is in Appendix~\ref{app:expert-annotation}.

This paper makes three contributions.
\begin{enumerate}\setlength\itemsep{1pt}
\item \textbf{A theory-grounded generation framework} (\textsc{Forma}) that compiles a cognitive model into a directed weighted graph, samples configurations as personas, and uses a closed-loop validator to enforce structural fidelity (\S\ref{sec:methodology}), together with a three-condition ablation design (full / no-formulation / zero-shot) that isolates the marginal contribution of the graph from the persona's self-report items and the demographic specification (\S\ref{sec:conditions}).
\item \textbf{A multi-tier evaluation protocol} triangulating structural validity (judge versus specified graph, corroborated by two clinical raters annotating all $20$ directed edges on $150$ vignettes), geometric structure of the generated set (cross-model convergence and trauma-type stereotyping in embedding space), expert validity (two clinical experts), scale (an LLM judge on $16{,}500$ vignettes from $11$ models), ecological validity (a clinician user study with $N{=}100$ licensed practitioners), and demographic fairness (\S\ref{sec:rating-schema}--\ref{sec:fairness}).
\item \textbf{An empirical case that the cognitive graph is a methodological intervention, not a stylistic input variable.} 
Across the evaluation suite, full-formulation vignettes: (i) preserve the specified directed edges (MCC $+0.41$ vs $+0.01$ in zero-shot; AUC of edge weight as a predictor of edge presence $0.70$ vs $0.50$); (ii) in embedding space, make different generation models converge on a shared portrayal of the same persona ($d_z = 1.52$) while breaking trauma-type stereotyping; (iii) are rated higher by clinical experts (Cohen's $d_z$ up to $+1.98$) and pass as human-written to clinicians $85\%$ of the time (vs $22\%$ in zero-shot); (iv) elicit monotonically higher inter-rater agreement. Effects replicate across $11$ models from $5$ families (\S\ref{sec:results}).
\end{enumerate}
The released dataset\footnote{Data + code: \url{https://github.com/Amit-Oren/FORMA}; interactive data browser: \url{https://clinical-vignettes.streamlit.app/}.} comprises $16{,}500$ vignettes ($500$ personas $\times$ $11$ models $\times$ $3$ conditions), a $330$-vignette evaluation subsample with $2$ expert and $3$ LLM-judge ratings, and $1{,}600$ ratings from the clinician user study. Full release details in \S\ref{sec:reproducibility-release}.

\section{Related Work}
\label{sec:related-work}

Case vignettes have been used in clinical training, research, and assessment for decades \citep{heverly1984constructing,evans2015vignette,sheringham2021use}, and \citet{mohan2014validating} document how unvalidated vignettes distort clinical reasoning when used in training. Recent LLM-based generation efforts produce fluent but structurally shallow output: \citet{benoit2023chatgpt} and \citet{yanagita2024can} use zero-shot chat prompts on diagnostic labels; \citet{bhate2023zero} and \citet{frayling2024zero} report similar surface-fluency-without-faithfulness on clinical notes; \citet{reichenpfader2024simulating} use persona conditioning for demographic diversity; \citet{suhas2025thousand} release a large-scale dataset of Prolonged-Exposure therapy conversations from synthetic clients. %whose demographic profiles are sampled to match U.S.\ population distributions \citep{uscensus2025quickfacts}. 
None of this work uses a formal cognitive model of the disorder as the generation specification or evaluates whether the generated text preserves directed causal structure.

The theoretical move that would close this gap is already standard in clinical psychology. Case formulation represents a disorder not as a symptom checklist but as a person-specific directed network of psychological components whose interactions explain how the disorder is maintained \citep{persons2008case,beck1979cognitive,salkovskis1985obsessional,clark1995cognitive,ehlers2000cognitive}. %; this view has been formalised in psychometrics as the network theory of mental disorders \citep{borsboom2017network} and recently shown to be empirically recoverable from ecological momentary assessment data on real patients \citep{blomfield2025mapping,hertzpalmor2025quantifying}.
Generating text conditioned on a formal structure is a recognised problem family with established graph-to-text \citep{koncel2019text} and knowledge-grounded generation \citep{yu2022survey} machinery, but it has not been applied to clinical case generation where the ``\textit{knowledge}'' is a mechanistic cognitive model of the disorder. 
%A recent wave of clinical 
Recent NLP work has begun to bring theoretic structure into LLM generation of mental-health text but stops short of a formal graph. \citet{wang2024patient} simulate %psychiatric
patients for CBT training by conditioning generation on a free-form cognitive conceptualisation diagram; \citet{louie2024roleplay} elicit domain-expert principles as the scaffold for client simulation; \citet{sharma2024cognitive} use LLMs to rewrite negative thoughts via cognitive-reframing operators, and \citet{Laban2025AReappraisal} used an LLM-based architecture for cognitive reappraisal in on-going interventions. %; and \citet{chen2023soulchat} fine-tune for empathetic multi-turn support.
More specifically related, \citet{tu2025trust} built an LLM dialogue system for structured PTSD diagnostic assessment; \citet{wang2025annaagent} simulate counselling seekers with an emotion-modulator and multi-session memory; and \citet{wu2026ecas} ground counselling-dialogue generation in multiple psychological theories simultaneously. The structure these systems use is either a free-form natural-language summary, a single-operator transformation, or a dialogue-act schema for assessment rather than for content generation and there is no formal directed graph of the disorder's components that the generated text can be audited against, edge by edge. We test this trajectory by compiling a %published
cognitive model into a directed weighted graph, sample concrete graph configurations as personas, and audit the generated text against the specified edges with an external structural probe (\S\ref{sec:res-structural}), treating the cognitive formulation as an auditable computational specification rather than a free-form text scaffold.
%LLM-as-judge evaluation \citep{zheng2023judging,li2025judgment} is one component of our multi-tier protocol (\S\ref{sec:rating-schema}); we treat the specified graph itself as an independent gold standard for calibrating the judges, rather than relying on judge--human agreement alone.

\section{Methodology}
\label{sec:methodology}
Following \citet{hertzpalmor2025quantifying}, we operationalise the \citet{ehlers2000cognitive} cognitive model of PTSD as a directed graph over five components: trauma memory, re-experiencing triggers, sense of current threat, negative appraisals, and maladaptive cognitive/behavioural strategies (definitions and operationalisation in Appendix~\ref{app:ec-components}). For each persona we sample a graph configuration and matched self-report items from a $403$-item clinical pool derived from real-patient disclosures to the five EC components in a clinical sample \citep{anon2026clinical}; these were parameterised into per-component item distributions, and each persona's items are drawn at random from these distributions and then verified by the Vignette Validator (\S\ref{sec:vignette-validation}). Full pool and sampling procedure in Appendix~\ref{sec:appendix-items} (pool released in the repository).

\subsection{Persona Construction}
\label{sec:persona-construction}
\begin{figure*}[h!]
    \centering
    \includegraphics[width=.83\linewidth]{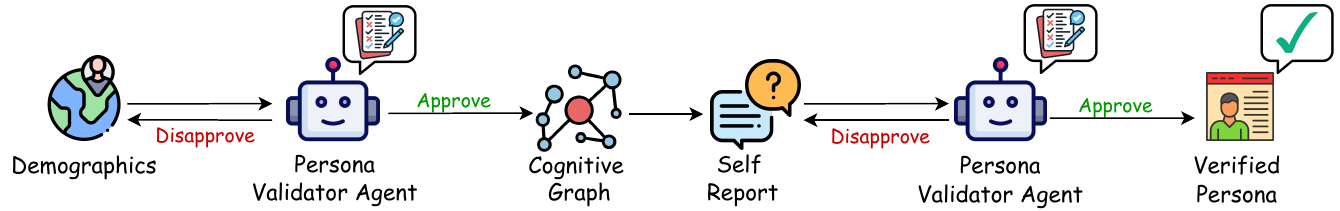}
    \caption{Persona construction. Demographics and a cognitive formulation graph are sampled independently; three self-report items are drawn per active component. A Persona Validator checks plausibility; failures are routed to a Persona Crafter for targeted repair.}
    \label{fig:persona-pipeline}
\end{figure*}

Each persona is built in three sampled-and-validated stages (Figure~\ref{fig:persona-pipeline}).
\emph{Demographics} (age, gender, ethnicity, occupation, trauma type, PCL-5 severity, relationship status; schema following \citealp{uscensus2025quickfacts}) are sampled from category-uniform distributions (relationship status conditioned on age) and checked by a \textit{Persona Validator Agent} for clinical plausibility (e.g., a 19-year-old is unlikely to be widowed); failed fields are resampled, up to six attempts.
\emph{Cognitive model.} Each of the five Ehlers~\&~Clark (EC) nodes is independently activated with probability $p_\text{node}=0.7$ and each directed edge between active nodes with $p_\text{edge}=0.7$; active edges receive a continuous weight from $\mathrm{Uniform}(0.01,1.0)$, inactive edges weight zero.
\emph{Self-report items.} For each active component we sample three items from the clinical item pool. A \textit{Persona Validator Agent} checks internal coherence, trauma-type consistency, and cross-node contradictions; failed items are replaced by a \textit{Persona Crafter Agent}, again up to six attempts. Both persona-level agents use \texttt{g-5.4}; the same 500 personas are reused across every (model, condition) cell in \S\ref{sec:gen-set}.

\subsection{Vignette Generation}
The \textit{Vignette Crafter Agent} receives the persona's demographics, the sampled self-report items, and the cognitive graph (active components and weighted causal links). Its task is to write a 500--700 word third-person clinical narrative reflecting the persona faithfully. Edge weights map to four narrative-prominence levels: $>0.5$ explicit, $0.1$--$0.5$ implicit, $<0.1$ passing, zero-weight components if they appear at all are described in isolation with no causal language.

\subsection{Vignette Validation}
\label{sec:vignette-validation}
\begin{figure*}[h!]
    \centering
    \includegraphics[width=.8\linewidth]{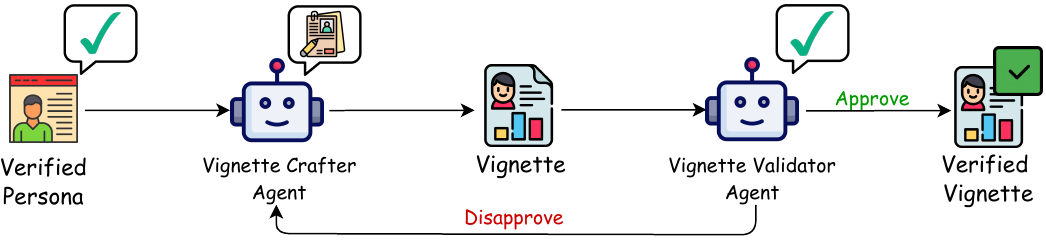}
    \caption{Vignette generation and validation loop. For each unsatisfied edge, the Validator passes violations to the Crafter, which retries up to five times.}
    \label{fig:crafter-loop}
\end{figure*}

The Vignette Validator and Crafter share an LLM and form a closed feedback loop (Figure~\ref{fig:crafter-loop}). The Validator checks each active component and each weight-positive edge $A\to B$: an edge is satisfied if some paragraph mentions both endpoints (or synonyms) and conveys that $A$ influences $B$. For each unsatisfied edge the Crafter writes a targeted \emph{patch} (anchor sentence copied verbatim plus a new sentence inserted after it) instead of rewriting; up to five retries. Most vignettes pass on the first attempt (Appendix~\ref{app:iterations}); the sole exception is \texttt{gpt-4o-mini}, whose runs retain a median of $4$ unsatisfied edges because it consistently fails to emit valid patch-block markers (a markup-compliance issue, not content), so we exclude it from per-condition scaled-judge and GT-recovery analyses but retain it for rating-schema-structure analyses.

\subsection{Ablations}
\label{sec:conditions}
We run every persona under three conditions to isolate each layer of structure. \textit{Full}: demographics + self-report items + cognitive graph, structural validation on active edges. \textit{No-formulation}: demographics + self-report items only, no graph; validator checks consistency with self-report. \textit{Zero-shot}: demographics only, no validation.

\section{Experimental Setup}
\label{sec:experimental-setup}
\subsection{Persona and Vignette Set}
\label{sec:personas}
\label{sec:gen-set}
\label{sec:eval-set}
We generated $500$ personas spanning ages $18$--$80$ ($M = 48.5$, $SD = 18.2$), approximately balanced gender, ethnicity uniform across eight world regions, and $29$ trauma categories (interpersonal violence, childhood trauma, war and collective violence, accidental and life-threatening events, loss/medical). PCL-5 severity ($M = 57.8$, $SD = 13.9$) placed every persona above the clinical threshold for probable PTSD. Chi-square tests confirm that demographic and clinical attributes are mutually independent in the realised set (Appendix~\ref{app:chi_square}); full distributional breakdowns are in Appendix~\ref{sec:appendix-demographics}. Each persona is generated under the three conditions of \S\ref{sec:conditions} with eleven generation models from five families (Anthropic, DeepSeek, Google, OpenAI, Alibaba; full model list and per-model token counts in Appendix~\ref{app:tokens}), yielding $16{,}500$ vignettes. Tokens per vignette range from $\sim900$ (small open-weight, zero-shot) to $\sim13{,}000$ (\texttt{gemini-2.5-pro}, full). For inter-judge analysis we drew a balanced subsample of $330$ vignettes ($10$ personas $\times$ $11$ models $\times$ $3$ conditions), stratified by trauma category, gender, age, and PCL-5; the same $10$ personas appear under every (model, condition) cell to hold persona-level variance constant.

\subsection{Rating Schema, Raters, and Metrics}
\label{sec:rating-schema}
\label{sec:metrics}
Every vignette is scored on $18$ items in four groups with distinct roles: three CVI items (clarity, relevance, importance; $1$--$3$; \citealt{st2021developing}) for perceived quality, four construction guidelines (grounding, narrativity, diagnostic salience, clinical economy; $1$--$3$; \citealt{evans2015vignette}) for narrative quality, six DSM-5 PTSD criteria \citep{apa2013dsm} (binary) as a diagnostic-floor sanity check, and five EC components (binary) as the locus of the formulation effect. Five raters score the $330$-vignette subsample: two clinical experts (R1, R2; both MD general-practice residents trained to read clinical cognitive formulations, blind to model and condition) and three LLM judges (\texttt{claude-opus-4-5}, \texttt{deepseek-v4-pro}, \texttt{deepseek-v4-flash}); Flash additionally scores the full $16{,}500$-vignette set. R1 and R2 first calibrated against six training vignettes (not part of the evaluation subsample), discussed any disagreements with the authors until convergence on the rating standard, then completed the $330$-vignette scoring blind over approximately four days each. We report Gwet's AC1 (binary) and AC2 with quadratic weights (ordinal) for inter-rater agreement (six items have base rate $\geq 0.85$ in at least one condition. %, where Gwet's chance term is more robust than Cohen's $\kappa$). 
Condition contrasts use omnibus RM-ANOVA (ordinal, $df = (2, 218)$) or Cochran's Q \citep{cochran1950comparison} (binary), then Bonferroni-corrected paired $t$-tests / proportion contrasts with paired Cohen's $d_z$. For ground-truth recovery (rater's binary EC labels vs the persona's specified \texttt{active\_nodes}, \S\ref{sec:res-structural}) we report precision, recall, F1, balanced accuracy, MCC, and Gwet's AC1. Generation and judging hyperparameters, total compute, and the full system prompts for the Crafter, Validator, Edge Probe, and Judge agents are in Appendices~\ref{app:hyperparams} and \ref{app:instructions}; Appendix~\ref{app:expert-annotation} reproduces a representative full-condition vignette with its specified graph (Figure~\ref{fig:persona14-graph}) and an expert annotation of each EC-relevant span.

\section{Results}
\label{sec:results}

\subsection{Structural Fidelity}%: Components and Edges}
\label{sec:res-structural}
We test whether the cognitive graph leaves a structural imprint on the text, starting from the strongest gold (two clinical experts) and scaling out.

\textbf{Component recovery.}
The two experts (R1, R2) score each persona's vignette for the presence of each EC component on the $330$-subsample. Treating the persona's specified \texttt{active\_nodes} as gold, in full all five components recover above chance (R1: AC1 $0.64$--$0.86$; R2: $0.66$--$0.85$), and recovery collapses to chance under zero-shot (e.g., R1 on \emph{Threat} $-0.52$ AC1; full per-EC per-rater in Appendix~\ref{app:gt-recovery-330}). The three LLM judges show the same direction and rank-order across components (Appendix~\ref{app:gt-recovery-330}). An expert qualitative annotation of a representative full-condition vignette, in which a licensed clinical psychologist tags each EC-relevant span span-by-span, is in Appendix~\ref{app:expert-annotation}.
%\textbf{Component recovery (scaled).}
Restricted to the $15{,}000$-vignette generation set, the scaled judge (\texttt{deepseek-v4-flash}) reproduces the pattern at scale: all five EC components recover above chance in full (AC1 $0.51$--$0.75$); the gap between full and zero-shot is the size of the formulation's contribution (Appraisals MCC $.59 \!\to\! .01$, Strategies $.41 \!\to\! -.01$; Appendix~\ref{app:judge-vs-specification}).

\textbf{Edge recovery.}
We then ask whether the specified \emph{directed edges} between components also appear. An external probe (the scaled judge, blind to the persona's active set and condition) reads every vignette in the $330$-subsample and returns a binary judgement on each of the twenty directed EC pairs $A\to B$, yielding $6{,}600$ paired cells. Table~\ref{tab:edge-recovery-by-cond} reports the metrics: in full, the judge recovers specified edges above chance (MCC $= +0.41$, AUC $= 0.70$); in no-formulation, recovery is at chance (MCC $= +0.03$); in zero-shot, at chance (MCC $= +0.01$). The judge's ability to read the specified graph off the text is created by the formulation.

The probe cannot read back an input it never receives, being given only the vignette prose, never the graph, the self-report items, or the condition (\S\ref{sec:rating-schema}). The stronger control is no-formulation, which passes through the identical generation-and-validation loop with the same self-report content and differs from full only in whether the graph is supplied, yet recovers at chance. Recovery therefore tracks the specification rather than the pipeline or the validator loop, whose own limits we note in the \nameref{sec:limitations}.

%The graph behaves as a near-binary on/off switch: 
In full, specified-active edges are judged present $75\%$ of the time vs $35\%$ for specified-absent ($40.8$-pp gap; $+3.0$/$+0.5$ pp in no-form/zero-shot). Within the active set recovery is flat across weight bands ($0.74$/$0.75$/$0.84$), so weight matters as a presence cue more than as a prominence dial.
\begin{table}[h]
\centering
\small
\setlength{\tabcolsep}{4pt}
\caption{External edge-recovery probe (\texttt{deepseek-v4-flash}) on the $330$-vignette evaluation subsample. Top block: classification metrics treating ``specified active'' (weight $>0$) as the gold label and the judge's per-edge binary as the prediction. Bottom block: $P(\text{judge says edge present})$ by the persona's edge-weight band. The $40.8$-percentage-point gap between specified-active and specified-absent rates in the full condition (vs $0.5$ pp in zero-shot) is the signature of the formulation doing structural work. Each column aggregates $20$ directed edges across $110$ vignettes ($n=2{,}200$).}
\label{tab:edge-recovery-by-cond}
\begin{tabular}{l ccc}
\toprule
& \textbf{Full} & \textbf{No-form.} & \textbf{Zero-shot} \\
\midrule
\multicolumn{4}{l}{\textit{Classification metrics (active vs absent)}} \\
Precision       & \textbf{0.67}   & $0.51$  & $0.49$ \\
Recall          & \textbf{0.75}   & $0.34$  & $0.30$ \\
F1              & \textbf{0.71}   & $0.41$  & $0.37$ \\
MCC             & $\bm{+0.41}$    & $+0.03$ & $+0.01$ \\
AC1             & $\bm{+0.41}$    & $+0.07$ & $+0.06$ \\
AUC             & $\bm{0.70}$     & $0.52$  & $0.50$ \\
\midrule
\multicolumn{4}{l}{\textit{$P$(judge says edge present), by weight band}} \\
high ($>0.5$)        & $0.74$        & $0.35$ & $0.30$ \\
med ($0.1$--$0.5$)   & $0.75$        & $0.30$ & $0.27$ \\
low ($<0.1$)         & $0.84$        & $0.47$ & $0.44$ \\
zero ($=0$)          & $\bm{0.35}$   & $0.31$ & $0.30$ \\
\bottomrule
\end{tabular}
\end{table}

\textbf{Every directed edge type recovers above chance under the formulation.}
All $20$ directed EC edges have positive MCC against the specified gold in full ($+0.25$ to $+0.66$); the strongest are the directed pairs the cognitive model itself emphasises as maintaining the disorder (Maladaptive Strategies~$\to$~Threat: MCC $+0.66$; Triggers~$\to$~Negative Appraisals: $+0.63$; Threat~$\to$~Memory: $+0.57$; Threat~$\to$~Negative Appraisals: $+0.56$), and the weakest are edges with high judge-marginal rates such as Triggers~$\to$~Threat (recall $1.00$, precision $0.47$; the judge tends to read this connection from any narrative that mentions both triggers and threat). The full per-edge table is in Appendix~\ref{app:edge-recovery-per-edge}. Figure~\ref{fig:edge-recovery-graph} draws the same pattern as the EC graph once per condition, each edge weighted and coloured by the judge's recovery probability. The structure is clearly visible in the full panel, while no-formulation and zero-shot are uniformly washed-out, reflecting near-flat recovery across the graph.

\begin{figure*}[h!]
    \centering
    \includegraphics[width=0.75\textwidth]{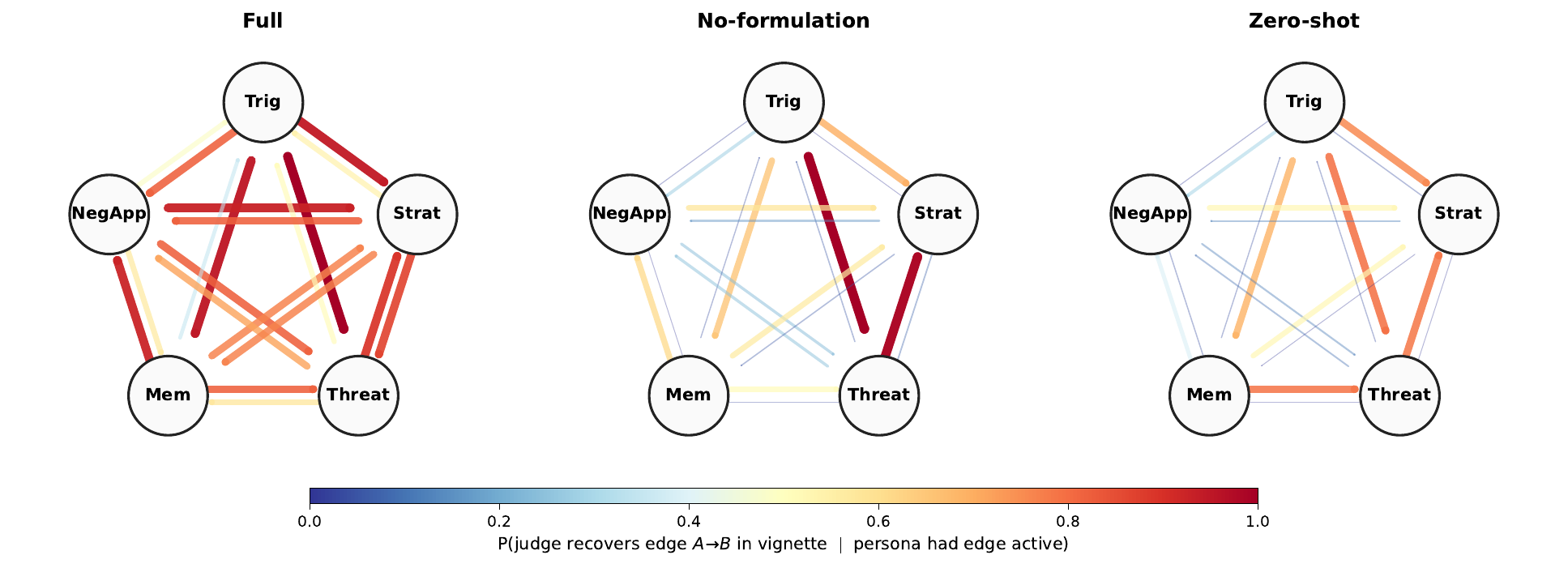}
    \captionsetup{font=small}
    \caption{\footnotesize Edge-level recovery of the specified cognitive graph by condition ($n = 110$ vignettes per panel, $1{,}980$--$2{,}300$ directed-edge judgements aggregated). Each panel shows the five-node EC graph (Trig=Triggers, NegApp=Negative Appraisals, Mem=Memory, Threat=Sense of threat, Strat=Maladaptive Strategies). Each directed edge $A\to B$ is coloured and thickness-weighted by $P(\textit{judge marks }A\to B\textit{ present}\mid\textit{persona had this edge active})$. Edge structure is clearly visible only under Full; no-formulation and zero-shot panels show uniform near-flat recovery, indicating that without the graph in the input the text contains no signal about which edges the persona was constructed with.}
    \label{fig:edge-recovery-graph}
\end{figure*}

\subsection{Human Validation of Directed-Edge Recovery}
\label{sec:res-human-edge}
The edge-recovery result above is produced by an LLM probe, and the two experts validate EC \emph{components} rather than the directed links themselves. We therefore replicated the edge analysis under human annotation at comparable scale. Two clinical raters independently annotated all $20$ directed EC edges on a $150$-vignette sample ($50$ personas $\times$ $3$ conditions, balanced across the ten retained generation models), blind to condition and to the persona's specified graph, yielding $3{,}000$ directed-edge judgements per rater. Each edge was scored both for presence and on a graded $0$--$1$ prominence scale. The identical schema was applied to the same $150$ vignettes by two of the paper's own LLM judges (\texttt{deepseek-v4-flash}, \texttt{deepseek-v4-pro}), giving a four-rater panel of two humans and two models. Protocol and full pairwise statistics are in Appendix~\ref{app:human-edge-validation}.

\textbf{The probe is indistinguishable from an additional human annotator.} On the presence scale every rater pair agrees at Gwet's AC1 between $0.63$ and $0.75$, and human-to-LLM agreement (mean AC1 $= 0.67$) is at least as high as agreement between the two humans themselves (AC1 $= 0.63$). The graded scale behaves the same way (human-to-LLM ICC $0.67$--$0.72$; human-to-human ICC $= 0.67$). Agreement with the probe is therefore not lower than the agreement two clinicians reach with each other.

\textbf{Humans recover the specified graph.} All four raters independently reproduce the recovery gradient reported above (Table~\ref{tab:human-edge-validation}). In full, MCC against the specified graph ranges from $+0.32$ to $+0.41$ and the AUC of the specified weight predicting detection from $0.65$ to $0.69$, against approximately zero MCC and approximately $0.50$ AUC in both no-formulation and zero-shot. The graded strength ratings carry the same formulation-dependent signal, correlating with the specified weight under full (Spearman $0.34$--$0.43$ across raters) and at chance in both ablations, so the weights leave a trace that appears only when the graph is supplied. Because the human raters apply the causal standard by hand, without sight of the graph, the recovery reported in \S\ref{sec:res-structural} is a property of the generated text rather than an artefact of the probe and the in-loop validator sharing a rubric.

\begin{table}[h]
\centering
\small
\setlength{\tabcolsep}{4pt}
\caption{\footnotesize Human validation of directed-edge recovery on the $150$-vignette sample ($50$ personas $\times$ $3$ conditions; $20$ directed edges; $3{,}000$ judgements per rater). H1 and H2 are the two clinical raters; \textsc{Flash} and \textsc{Pro} are \texttt{deepseek-v4-flash} and \texttt{deepseek-v4-pro} scoring the identical schema, blind to condition and to the specified graph. All four raters recover the specified graph under full and sit at chance under both ablations, reproducing Table~\ref{tab:edge-recovery-by-cond} under blind human annotation.}
\label{tab:human-edge-validation}
\begin{tabular}{l ccc}
\toprule
& \textbf{Full} & \textbf{No-form.} & \textbf{Zero-shot} \\
\midrule
\multicolumn{4}{l}{\textit{MCC vs specified graph}} \\
H1                  & $\bm{+0.41}$ & $+0.06$ & $+0.04$ \\
H2                  & $\bm{+0.32}$ & $+0.04$ & $+0.01$ \\
\textsc{Flash}      & $\bm{+0.32}$ & $+0.04$ & $-0.02$ \\
\textsc{Pro}        & $\bm{+0.38}$ & $-0.01$ & $-0.02$ \\
\midrule
\multicolumn{4}{l}{\textit{AUC (specified weight $\to$ detection)}} \\
H1                  & $\bm{0.69}$ & $0.54$ & $0.53$ \\
H2                  & $\bm{0.65}$ & $0.53$ & $0.51$ \\
\textsc{Flash}      & $\bm{0.65}$ & $0.52$ & $0.49$ \\
\textsc{Pro}        & $\bm{0.67}$ & $0.50$ & $0.49$ \\
\midrule
\multicolumn{4}{l}{\textit{Spearman (rated strength, specified weight)}} \\
H1                  & $\bm{+0.43}$ & $+0.08$ & $+0.05$ \\
H2                  & $\bm{+0.36}$ & $+0.06$ & $+0.02$ \\
\textsc{Flash}      & $\bm{+0.34}$ & $+0.03$ & $-0.01$ \\
\textsc{Pro}        & $\bm{+0.39}$ & $+0.01$ & $-0.02$ \\
\bottomrule
\end{tabular}
\end{table}

\subsection{Vignettes' Geometric Structure}
\label{sec:res-geometric}
Two embedding-space analyses test whether the formulation acts as a shared structural specification across models. To match the structural and clinical-content analyses (\S\ref{sec:res-structural}, \S\ref{sec:fairness}), the geometric analysis uses the $10$ generation models that successfully validate; \texttt{gpt-4o-mini} is excluded for the validator-loop pathology documented in Appendix~\ref{app:iterations}. To guard against an encoder-specific artefact, we re-embed these $15{,}000$ vignettes with two sentence encoders trained on different objectives and corpora: MPNet \citep{song2020mpnet,reimers2019sentence} (\texttt{all-mpnet-base-v2}), used as the primary encoder in the body, and BGE-base \citep{xiao2024c} (\texttt{BAAI/bge-base-en-v1.5}) used %as an alternative encoder
for replication. %We report MPNet numbers in the body and cross-validate every claim against BGE.

\textbf{Cross-model convergence.} The within-persona cross-model cosine similarity (MPNet) is highest in full ($\mu = 0.792$), intermediate in no-formulation ($0.767$), lowest in zero-shot ($0.737$); all three pairwise differences are significant (paired $t$, $n=500$, $p<10^{-48}$), full-vs-zero-shot is the largest contrast ($d_z = 1.52$). BGE preserves the ordering and the effect size ($\mu$: $0.837 / 0.816 / 0.800$; full-vs-zero-shot $d_z = 1.61$; Appendix~\ref{app:sens-embedding}, Figure~\ref{fig:sens-embedding}). The two encoders disagree only on absolute level (BGE places all vignettes closer in cosine space than MPNet), but the conditional contrast that the claim rests on is encoder-invariant. The formulation acts as a cross-model anchor: with the graph in the input, the ten retained generation models converge on a shared portrayal under either encoder.

\textbf{Trauma-type stereotyping.} A complementary $t$-SNE analysis (Appendix~\ref{app:fig-trauma-stereotyping}) shows that under zero-shot, vignettes of the same trauma type cluster tightly in latent space: the generator writes from a trauma-type template rather than from the persona's individual content; the same coloured points are dispersed under full and no-formulation. Appendix~\ref{app:sensitivity} quantifies this shift (cognitive-signature/trauma-category cohesion ratio rises from $0.07$ in zero-shot to $0.48$ in full, a $\sim 7\times$ shift) and replicates the $t$-SNE pattern under BGE and across model capability tiers. Absolute cluster cohesion is lower under BGE than under MPNet, but the conditional gradient %(zero-shot $\to$ no-formulation $\to$ full)
and the tier $\times$ condition interaction reproduce, ruling out an encoder-specific artefact.

\subsection{Inter-Rater Agreement}
\label{sec:res-judges}

Five raters score the 330-vignette subsample using the schema in \S\ref{sec:rating-schema}: two clinical experts (R1, R2) and three LLM judges (Opus, DSpro, Flash). Table~\ref{tab:judge-consensus} reports inter-rater agreement (R1--R2, mean LLM--LLM, multi-rater All 5) per condition. Three patterns repeat across metric families: LLM--LLM agreement is uniformly high (median $\geq 0.93$) in every condition; R1--R2 is near-perfect on CVI, Construction, and DSM items but only strong on the EC items in the full condition ($0.82$--$0.95$), dropping to moderate in no-formulation (Threat $0.43$, Strategies $0.54$) and zero-shot (Strategies $0.36$); and the multi-rater All-5 collapses on the EC items in zero-shot (Threat $0.10$) even though LLM--LLM remains near-ceiling there, indicating that the loss of agreement is driven by human--LLM divergence on EC content (the six cross pairs and full 10-pair matrix in Appendix~\ref{app:full-pairwise}).

\begin{table}[h]
\centering
\small
\setlength{\tabcolsep}{3pt}
\caption{\footnotesize Inter-rater agreement on the 330-vignette evaluation subsample, per condition ($n = 110$ each), aggregated to three cluster summaries. AC1 (Gwet's) for binary; AC2 (quadratic weights) for ordinal. R1--R2: the two human experts. LLM: mean over the three LLM--LLM pairs (Opus--Pro, Opus--Flash, Pro--Flash). All 5: multi-rater AC across R1, R2, and the three LLM judges. The full 10-pair matrix, including the six human--LLM cross pairs, is in Appendix~\ref{app:full-pairwise}.}
\label{tab:judge-consensus}
\resizebox{\linewidth}{!}{%
\begin{tabular}{l rrr rrr rrr}
\toprule
& \multicolumn{3}{c}{\textbf{Full}} & \multicolumn{3}{c}{\textbf{No-form.}} & \multicolumn{3}{c}{\textbf{Zero-shot}} \\
\cmidrule(lr){2-4} \cmidrule(lr){5-7} \cmidrule(lr){8-10}
\textbf{Metric} & R1--R2 & LLM & All5 & R1--R2 & LLM & All5 & R1--R2 & LLM & All5 \\
\midrule
\multicolumn{10}{l}{\textit{Ordinal (AC2, quadratic weights)}} \\
Clarity     & 0.974 & 0.950 & 0.896 & 0.941 & 0.978 & 0.857 & 0.937 & 0.987 & 0.754 \\
Relevance   & 0.973 & 0.976 & 0.945 & 0.956 & 0.980 & 0.901 & 0.923 & 0.979 & 0.775 \\
Importance  & 0.977 & 0.963 & 0.951 & 0.903 & 0.947 & 0.888 & 0.902 & 0.863 & 0.783 \\
Grounded    & 0.976 & 0.935 & 0.871 & 0.962 & 0.971 & 0.839 & 0.937 & 0.962 & 0.657 \\
Narrative   & 0.983 & 0.976 & 0.951 & 0.960 & 0.945 & 0.859 & 0.944 & 0.848 & 0.707 \\
Explicit    & 0.983 & 0.998 & 0.975 & 0.961 & 1.000 & 0.913 & 0.971 & 0.998 & 0.766 \\
Focused     & 0.993 & 0.899 & 0.869 & 0.940 & 0.927 & 0.830 & 0.908 & 0.917 & 0.701 \\
\midrule
\multicolumn{10}{l}{\textit{DSM-5 binary (AC1)}} \\
Trauma exposure     & 1.000 & 0.895 & 0.925 & 1.000 & 0.797 & 0.842 & 0.855 & 0.927 & 0.867 \\
Intrusion           & 1.000 & 0.937 & 0.887 & 0.981 & 0.935 & 0.869 & 0.933 & 0.975 & 0.916 \\
Avoidance           & 1.000 & 0.931 & 0.902 & 0.991 & 0.852 & 0.861 & 0.831 & 0.915 & 0.803 \\
Negative cognitions & 1.000 & 0.975 & 0.983 & 1.000 & 0.969 & 0.982 & 0.952 & 0.797 & 0.792 \\
Hyperarousal        & 1.000 & 0.973 & 0.966 & 1.000 & 0.952 & 0.945 & 0.961 & 0.988 & 0.966 \\
Impairment          & 1.000 & 0.981 & 0.989 & 1.000 & 0.994 & 0.996 & 1.000 & 0.981 & 0.985 \\
\midrule
\multicolumn{10}{l}{\textit{Ehlers \& Clark binary (AC1)}} \\
Threat        & 0.820 & 0.753 & 0.723 & 0.427 & 0.555 & 0.348 & 0.674 & 0.553 & 0.096 \\
Appraisals    & 0.877 & 0.896 & 0.801 & 0.734 & 0.707 & 0.432 & 0.580 & 0.642 & 0.336 \\
Memory        & 0.946 & 0.750 & 0.697 & 0.886 & 0.644 & 0.418 & 0.710 & 0.543 & 0.277 \\
Strategies    & 0.952 & 0.946 & 0.882 & 0.535 & 0.580 & 0.426 & 0.356 & 0.253 & 0.108 \\
Triggers      & 0.902 & 0.944 & 0.853 & 0.814 & 0.896 & 0.723 & 0.761 & 0.915 & 0.677 \\
\bottomrule
\end{tabular}
}
\end{table}

\noindent The conditional gradient is clearest on the EC items: \textit{Threat} falls from All-5 AC $= 0.72$ in full to $0.10$ in zero-shot, \textit{Strategies} from $0.88$ to $0.11$, \textit{Memory} from $0.70$ to $0.28$, and \textit{Appraisals} from $0.80$ to $0.34$ (Table~\ref{tab:judge-consensus}, rightmost columns within each block; visualised in Figure~\ref{fig:multirater-by-condition}).
This is the same conditional gradient the scaled judge reports for EC \emph{coverage} in \S\ref{sec:res-by-condition}: the cognitive formulation does not merely make raters score higher on EC presence; it makes raters agree on what they see, because the components are actually there in the text. Without the formulation specification, EC content is sparse and inconsistent across vignettes, and raters disagree on whether it is present at all.

\subsection{Clinical-Content Quality}%: Experts and Scaled Judge}
\label{sec:res-quality}
\label{sec:res-expert}
\label{sec:res-by-condition}
%Does the structural lift translate to clinical content quality? 
Two experts rate the $330$-vignette subsample (blind) and \texttt{deepseek-v4-flash} scales the schema to all $15{,}000$ retained-model vignettes ($n = 5{,}000$ per condition). Pairing is at the (persona $\times$ generation model) unit. %Full results in
See Appendices~\ref{app:expert-by-condition}, \ref{app:flash-by-condition}.

\textbf{Diagnostic validity.} Before any condition contrast, the vignettes clear the diagnostic floor. Coverage of the six DSM-5 PTSD criteria is at ceiling in all three conditions ($0.85$--$1.00$), independently for the two clinical experts (Appendix~\ref{app:expert-by-condition}) and for the scaled judge across all $15{,}000$ vignettes (Appendix~\ref{app:flash-by-condition}). Every condition therefore yields recognisable PTSD presentations, and the effects reported below are differences in cognitive-formulation structure between vignettes that are already diagnostically valid, not differences in whether the text depicts PTSD at all.

\textbf{Experts.} Omnibus reaches $p<.001$ on all seven ordinal items, three of six DSM-5 items, four of five EC items. \textit{Importance} shows the strongest contrast (full $2.94$ vs zero-shot $2.23$, $d_z = +1.98$); on EC items the gap is larger than the scaled judge's: Threat $0.86\!\to\!0.19$ ($d_z = +1.58$), Appraisals $0.85\!\to\!0.40$ ($+0.88$), Memory $0.91\!\to\!0.53$ ($+0.71$), Strategies $0.85\!\to\!0.44$ ($+0.76$). Non-significant omnibi cluster where the condition is structurally invariant (DSM-5 at ceiling, Triggers narratively familiar).

\textbf{Scaled judge.} (i) EC coverage is where the formulation pays: Strategies (full $0.91$ vs zero-shot $0.49$), Appraisals ($0.85$ vs $0.52$), Memory ($0.60$ vs $0.42$) show the largest gaps; Triggers saturates at $\sim 0.93$. (ii) Narrative ($2.89$ vs $2.47$) and Importance ($2.82$ vs $2.54$) follow. Clarity slightly favours the no-formulation condition ($d = -0.09$), an LLM-judge preference for unconstrained prose that the experts overturn on the same item ($d_z = +0.81$; per-rater means in Appendix~\ref{app:judge-means}). Vignette lengths are comparable across conditions (full $\mu=753$ vs zero-shot $\mu=682$ words, $\sim 10\%$ grand-mean difference); the EC-Strategies and EC-Appraisals effects survive length-matched within-persona pairing ($d_z=+0.72$ and $+0.53$ at median $16$-word length difference; Appendix~\ref{app:sens-length}).

\subsection{Clinician User Study}
\label{sec:user-study}
\label{sec:us-method}
\label{sec:us-results}

\begin{figure*}[!h]
    \centering
    \includegraphics[width=0.85\textwidth]{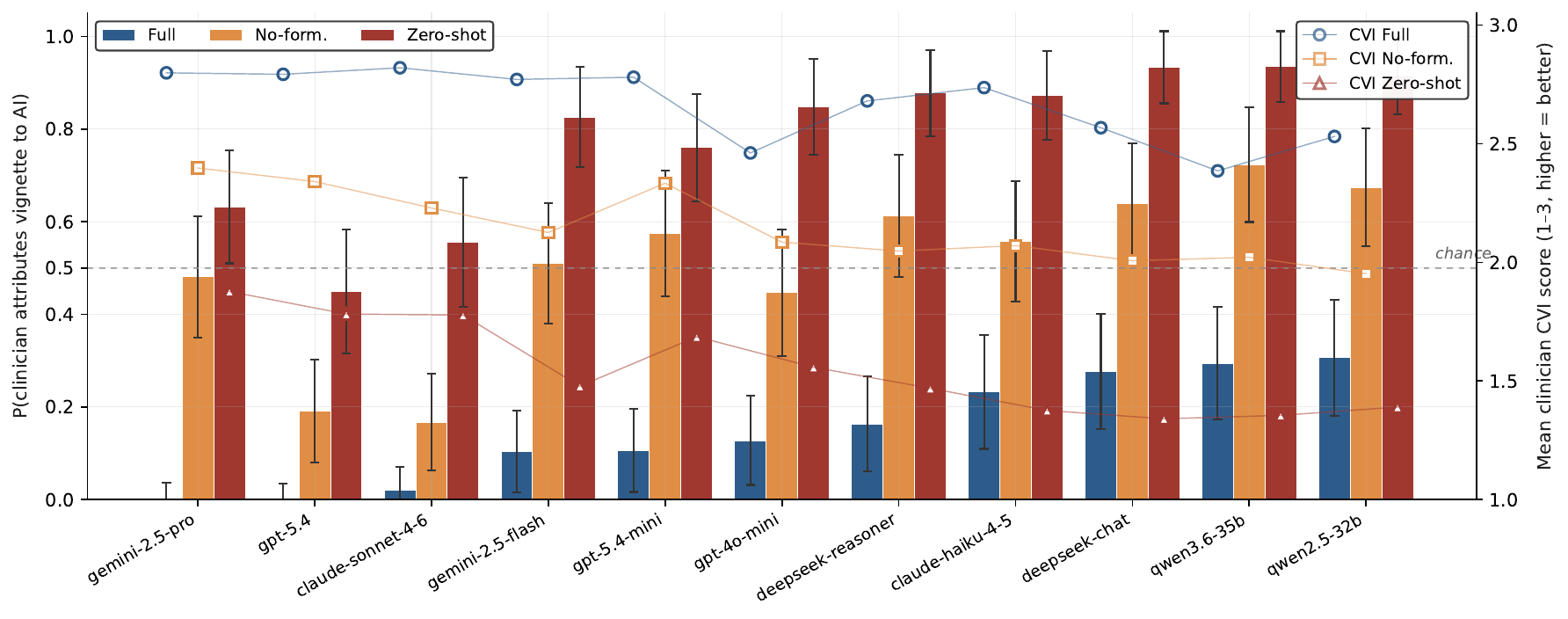}
    \caption{\footnotesize Clinician AI-authorship attribution rate (bars, left axis) and mean CVI quality score (markers, right axis) by generation model and condition. $N = 100$ clinicians. Models ordered by ascending full-condition AI rate. Bars are Wilson 95\% CI. Two large proprietary models (\texttt{gpt-5.4}, \texttt{gemini-2.5-pro}) reach $0\%$ AI-attribution in the full condition while scoring near-ceiling CVI; smaller open-weight models remain recognisable as AI ($28$--$31\%$) but still receive higher CVI in full than in zero-shot.}
    \label{fig:us-ai-detection}
\end{figure*}
Expert and judge evaluations test content; they do not test whether practising clinicians find the vignettes plausible. We recruited $N = 100$ licensed clinicians with PTSD-patient experience (psychologists, psychiatrists, MD GPs, nurses; Appendix~\ref{app:us-sample}) via an online panel, paid \pounds$7.50$/hour pro-rated; ethics approval, informed consent, and debriefing in Appendix~\ref{app:us-sample}. Each rated $17$--$18$ randomly assigned vignettes plus an attention check, blind to model and condition ($1{,}600$ ratings, $43$--$57$ per cell). They rated the three CVI items and perceived AI authorship (binary). We fit LMMs for the ordinal outcomes and GEE logistic for the binary \citep{liang1986longitudinal}, clustering on participant.

\textbf{Perceived quality.} The condition effect is monotone on every item. Full vignettes are rated $> 1$ point higher on the $1$--$3$ scale than zero-shot by the same clinician: Clarity $b{=}{+}1.087$, Relevance $b{=}{+}1.109$, Importance $b{=}{+}1.111$ (all $t>34$, $p<10^{-250}$); the no-formulation contrast is half the size on every metric ($p<10^{-58}$). Participant ICC is $\sim 0.08$; the condition effect is roughly an order of magnitude larger than between-participant variance (Appendix~\ref{app:us-mixed-quality}). Clinicians spent $31\%$ longer reading full than zero-shot vignettes.

\textbf{Perceived AI authorship.} %Although every vignette was LLM-generated, 
Clinicians perceive AI authorship in $14.7\%$ of full vignettes, $50.8\%$ in no-formulation, $77.8\%$ in zero-shot. GEE %logistic 
confirms all %three 
contrasts (full-vs-zero-shot OR $= 0.05$, $z = {-}18.5$; full-vs-no-formulation OR $= 0.17$; no-form-vs-zero-shot OR $= 0.30$; all $p < 10^{-15}$; Appendix~\ref{app:us-ai-gee}). Per-model (Figure~\ref{fig:us-ai-detection}; Appendix~\ref{app:us-by-model}): \texttt{gpt-5.4} and \texttt{gemini-2.5-pro} reach $0\%$ AI-attribution in full; %smaller 
open-weight models remain perceivable at $28$--$31\%$; the formulation lift is large for %every 
all models but does not equalise model quality. 

\subsection{Fairness by Demographic and Condition}
\label{sec:fairness}
We test if the framework's output is demographically equivariant across persona gender, ethnicity, age, occupation, and trauma category, using Statistical Parity Difference \citep[SPD; ][]{dwork2012fairness}, Disparate Impact Ratio \citep[DIR; ][]{feldman2015certifying}, and pairwise Cohen's $d$ \citep{cohen1988statistical} alongside omnibus $F$-tests and demographic-by-condition interactions (Appendix~\ref{app:fairness-metrics}). The formulation collapses demographic disparity in scaled-judge perceived quality: every demographic SPD falls $1.5$--$7\times$ from zero-shot to full, full-condition DIRs exceed the four-fifths-rule threshold, $d_{\max} \leq 0.12$. EC coverage is invariant on gender, ethnicity, and age, with a residual content-fit disparity on trauma category (childhood-trauma narratives engage all five EC components more naturally than %loss/medical 
other narratives).%; the demographic-by-condition interaction is significant on trauma category. %Clinician ratings show no detectable demographic bias on quality or AI authorship.

\section{Discussion and Conclusion}
\label{sec:discussion}
\label{sec:conclusion}
We introduced \textsc{Forma}, a theory-grounded framework that compiles a cognitive model of a disorder into a directed weighted graph, samples a configuration for each persona, and conditions LLM generation on the resulting structural specification. The four evaluation tiers converge on a single finding: this changes the structural content of the generated vignettes, not only their surface style. The specified directed graph is recoverable from full-condition vignettes but not from zero-shot or no-formulation vignettes; the formulation makes generation models converge on a shared persona portrayal and breaks zero-shot trauma-type stereotyping; experts rate full vignettes higher than zero-shot; clinicians read full vignettes as human-written far more often than zero-shot vignettes; and the framework reduces demographic disparity in perceived quality on every demographic axis tested.

Across our evaluation tiers, LLM judges mostly "\textit{agreed}" on clinical text regardless of what the text contained, and on stylistic items their consensus ran counter to clinical experts. Inter-rater agreement on its own is therefore not evidence that an LLM-generated artefact is faithful to its specification. When the specification is itself structured, as a cognitive-formulation graph is here, it can serve as an external gold for the generated text; in this work the prompted graph is what makes the formulation effect identifiable. %We expect this form of structural calibration to be useful wherever the generator's input is itself auditable.

Compiling a cognitive model into an auditable directed graph turns synthetic case material into something a clinician can read for clinical plausibility and theoretical structure, and should generalise wherever a disorder admits a formal model. %We release the dataset, the four-tier evaluation protocol, and the code as a starting point for theory-grounded synthetic clinical text in disorders beyond PTSD.

\section*{Limitations}
\label{sec:limitations}
First, the in-loop Vignette Validator shares the same LLM model as the Vignette Crafter, so the per-model iteration counts in Appendix~\ref{app:iterations} reflect both generation ability and self-critique; the \texttt{qwen} rows, which pass on the first attempt with zero violations, illustrate the risk of fluent self-approval. We address this with an external edge probe (\S\ref{sec:res-structural}), but a stronger structural check with multiple independent validators is left to future work.
Second, both demographics and cognitive formulation graphs are sampled programmatically from real patient data rather than representing real patients. However, working with cognitive formulations often calls for similar practices. The pipeline should be read as a controlled stress test of whether LLMs can populate a clinically structured template, not as a replacement for real clinical material; the framework's faithfulness to actual clinical heterogeneity remains an empirical question we cannot answer with this dataset alone.
Third, all prompts, self-report items, and vignettes are in English, and the framework's behaviour under translation or culturally-specific persona sampling is untested.
Fourth, the framework instantiates one cognitive theory \citep{ehlers2000cognitive}. Cognitive theories of mental disorders are themselves contested; pinning vignette generation to one theory propagates that theory's assumptions and gaps to every vignette generated. We see this as a feature (the structural choices are explicit) but it should not be conflated with a generic claim about ``what PTSD looks like.''
Fifth, the fairness analysis (\S\ref{sec:fairness}) tests output equivariance only and does not test for stereotyping in content, representation harms in surface prose, or intersectional disparities; it also relies on a small number of unique persona profiles per demographic cell on the edge-probe subsample.
Sixth, the clinician user study was conducted with a panel-recruited sample of 100 licensed practitioners, predominantly female, predominantly U.S./U.K./Canada/Australia. Generalisability to clinicians outside this profile and to other patient-care contexts is an empirical question.

\section{Ethical Considerations}
\label{sec:ethics}
\textbf{No real patients.}
The dataset contains no real patient material. All personas, demographics, self-report items, and vignettes are LLM-generated from the sampling distributions documented in \S\ref{sec:personas}. The Persona Validator and the cognitive-model parameters are drawn from published clinical theory and from prior work that re-derived the EC graph from ecological momentary assessment data on real patients \citep{hertzpalmor2025quantifying}; no individual-level data was ingested. This is the primary ethical advantage of synthetic case generation, and the basis on which we believe the dataset can be released openly. The clinician user study, the only component involving human participants, was reviewed and approved by the ethics committee of the Department of Industrial Engineering and Management, Ben-Gurion University of the Negev, and participants gave informed consent before rating (Appendix~\ref{app:us-sample}).

\textbf{Mis-attribution and downstream use.}
The clinician user study finding that full-condition vignettes pass as human-written $85\%$ of the time is itself an ethical concern as well as a methodological one. If the dataset is incorporated into clinical training material or downstream LLM-evaluation benchmarks without clear provenance, end users may take it for material derived from real patients. We recommend that any release-time distribution of the dataset prominently and persistently label every vignette as synthetic and as not representing a real individual. A second risk runs the other way. Used as training or evaluation data for downstream clinical models, these vignettes propagate the framework's theoretical commitments and the generators' stylistic regularities into those models, amplifying the assumptions of a single cognitive theory and one generation pipeline; we take up that case below.

\textbf{Limits of theory-grounded generation.}
A vignette that is faithful to a cognitive model is still only as accurate as that cognitive model. Pinning generation to one theory of a disorder propagates that theory's framings, its blind spots, and any populations the theory underrepresents. The EC model \citep{ehlers2000cognitive} is well-validated for the cognitive maintenance of PTSD in Western clinical samples; it is not a theory of culturally-specific trauma response, of complex PTSD, or of disorders co-morbid with PTSD. Our vignettes inherit this scope.

\textbf{Bias and fairness.}
\S\ref{sec:fairness} reports an algorithmic-fairness analysis showing that the framework does not introduce demographic disparities in EC coverage and substantially reduces them in perceived quality. We emphasise the limitations of that analysis: it tests output equivariance on the variables the dataset records, not deeper representation harms or stereotyping in the textual content. We therefore caution against using the dataset to evaluate \emph{clinical} bias of downstream LLMs until those further axes have been audited.

\textbf{Use in training mental-health LLMs.}
Synthetic case material is a tempting training corpus for clinical LLMs because it bypasses privacy constraints. We urge restraint. The model that generates the training data shapes the model that consumes it; downstream LLMs trained on \textsc{Forma} vignettes will inherit our generators' rhetorical patterns and our framework's theoretical commitments. We see appropriate uses as standardised stimuli for clinician training (where the synthetic origin is disclosed), benchmark sets for clinical capabilities of LLMs (where the gold structure is the audited contribution, not the prose), and as a probe-set for inter-rater reliability research. We do not see appropriate uses as training data for clinical decision-support LLMs deployed in patient care.

\textbf{Reproducibility and release.}\label{sec:reproducibility-release}
The full dataset and code are released at \url{https://github.com/Amit-Oren/FORMA}, under the MIT License for the code and CC BY 4.0 for the datasets. We additionally provide an interactive web browser, \url{https://clinical-vignettes.streamlit.app/}, where readers can inspect any of the $500$ full-condition vignettes alongside its persona cognitive graph and the corresponding self-report items, making the vignette--graph correspondence visible end-to-end. We release the full $16{,}500$-vignette dataset, the $330$-vignette evaluation subsample with ratings from two clinical experts and three LLM judges, the user-study ratings ($N = 1{,}706$ total: $1{,}600$ vignette ratings + $106$ attention-check responses), all generation and evaluation code, and the cognitive-formulation graphs sampled per persona. Generation hyperparameters, total compute, and provider routing are documented in Appendix~\ref{app:hyperparams}; the full system prompts for the Crafter, Validator, Edge Probe, and Judge are in Appendix~\ref{app:instructions}; Appendix~\ref{app:expert-annotation} reproduces a representative full-condition vignette so that readers without the released code can see what the system produces, with an expert annotation of each EC-relevant span and the persona's cognitive graph (Figure~\ref{fig:persona14-graph}). Sensitivity to analytical choices (edge-weight threshold, judge identity, encoder, cluster operationalisation) is reported in Appendix~\ref{app:sensitivity}. The user-study data are released in aggregate; participant identifiers are randomised and demographic combinations that would re-identify individual respondents are coarsened or suppressed.

\textbf{AI assistant disclosure.} LLMs are the primary research artifact of this paper; their use as generators, validators, and judges is documented in \S\ref{sec:methodology}, \S\ref{sec:rating-schema}, \S\ref{sec:results}, and Appendix~\ref{app:instructions}. Separately, AI assistants (LLMs) were used for light copy-editing and \LaTeX{} formatting during manuscript preparation; all scientific claims, analyses, and figures were authored and verified by the human authors.

\section*{Acknowledgments}
Nimrod Hertz-Palmor is supported by the Gates Cambridge Trust (\#OPP1144).

\bibliography{custom}

\appendix
\renewcommand{\thesection}{\AlphAlph{\value{section}}}

% ===== EC components =====

\section{The Five Ehlers \& Clark Components}
\label{app:ec-components}

The cognitive graph used throughout this paper is the five-component formulation of \citet{ehlers2000cognitive}, the dominant cognitive model of PTSD in UK clinical practice and the basis for NICE-recommended trauma-focused CBT. The model proposes that PTSD is maintained by the interaction of five components rather than by any single symptom cluster; below we summarise each and note how it is operationalised in our graph and in the rater anchors of the schema judge (Appendix~\ref{app:instructions}).

\paragraph{Trauma memory.} The autobiographical memory of the traumatic event(s), characterised in the model by poor elaboration, fragmentation, and a strong sensory and ``here-and-now'' quality, such that retrieval is not experienced as remembering past danger but as re-living it. Operationally, we encode it as one of the five graph nodes; the schema judge scores it $1$ when the vignette shows fragmented, disorganised, or present-tense intrusive recall (rather than ordinary autobiographical recall of the trauma).

\paragraph{Re-experiencing triggers.} Specific internal or external cues (sights, sounds, sensations, contexts, internal states) that reliably reactivate the trauma memory and accompanying distress. Cues may be incidental and not obviously linked to the trauma. Operationally, this node is scored $1$ when the vignette describes identifiable cues that set off intrusion, hyperarousal, or avoidance.

\paragraph{Sense of current threat.} The pervasive subjective experience that danger is ongoing rather than past, manifesting as hypervigilance, startle, and bodily threat-response. This is the model's central maintaining mechanism (the model's title concept). Operationally, scored $1$ when the vignette portrays the person as feeling currently in danger, not merely cautious or unsettled.

\paragraph{Negative appraisals.} Catastrophic or overly negative interpretations of the trauma, of the person's reactions to it, and of its sequelae (e.g.\ ``I am permanently damaged'', ``the world is unsafe'', ``it was my fault''). Appraisals are the cognitive content the model targets for therapeutic change. Operationally, scored $1$ when explicit catastrophic or self-blaming beliefs are stated or shown.

\paragraph{Maladaptive cognitive/behavioural strategies.} Strategies the person uses to manage symptoms that, while affording short-term relief, prevent elaboration of the trauma memory and disconfirmation of the negative appraisals, thereby maintaining the disorder. Canonical examples: thought suppression, rumination, situational avoidance, safety behaviours, substance use. Operationally, scored $1$ when the vignette shows such a strategy in action with implied or explicit symptom-maintaining function.

\paragraph{Directed edges between components.} The model posits a small number of directed causal links between these components (e.g.\ Triggers $\to$ Memory $\to$ Threat $\to$ Maladaptive Strategies $\to$ Negative Appraisals $\to$ Threat). We do not hand-fix the edge set: each persona's directed weighted graph is sampled from per-edge weight distributions calibrated against the EMA-derived networks of \citet{hertzpalmor2025quantifying} (sampling procedure in \S\ref{sec:persona-construction}), and the external edge probe (\S\ref{sec:res-structural}, Appendix~\ref{app:edge-recovery-per-edge}) evaluates whether each of the $20$ possible directed pairs is present in the generated text given that it was active in the persona's specification.

% ===== Dataset and Personas =====

\section{Self-Report Item Pool}
\label{sec:appendix-items}
The item pool contains 403 items total, distributed across the five Ehlers \& Clark components (\S\ref{sec:persona-construction}, Step~3). Each item is a short clinical sentence reflecting a specific cognitive, behavioural, or experiential feature of the component. The pool derives from real-patient disclosures to the five EC components in a clinical sample \citep{anon2026clinical}: patient disclosures were parameterised into per-component item distributions, and each persona's items are drawn at random from these distributions. The Vignette Validator (\S\ref{sec:vignette-validation}) then verifies that the sampled set is realised in the generated text. For Maladaptive Strategies the pool merges three sub-pools (core maladaptive strategies, distraction strategies, and avoidance strategies) into a single sampling distribution. A small sample per component is shown below; the full pool is available in the code release.

\paragraph{Pool size per component.}
\begin{itemize}\setlength\itemsep{0pt}
  \item Triggers: 137 items
  \item Maladaptive Strategies: 130 items (48 core + 39 distraction + 43 avoidance)
  \item Negative Appraisals: 58 items
  \item Memory characteristics: 39 items
  \item Current threat appraisals: 39 items
\end{itemize}

\paragraph{Example items, Triggers.}
\begin{itemize}\setlength\itemsep{0pt}
  \item \textit{Olfactory (Alcohol)}: ``Alcohol smell''
  \item \textit{Olfactory Danger Cue}: ``Burnt smell''
  \item \textit{Auditory (Babies)}: ``Babies''
  \item \textit{Auditory (Cars)}: ``Car sounds''
  \item \textit{Auditory/Threat}: ``Gunshot noise''
\end{itemize}
\paragraph{Example items, Maladaptive Strategies.}
\begin{itemize}\setlength\itemsep{0pt}
  \item \textit{Suppress Thoughts}: ``try hard not think about the trauma''
  \item \textit{Mental Distraction}: ``keep mind occupied all the time''
  \item \textit{Alcohol Use}: ``drink alcohol''
  \item \textit{Drug Use}: ``take drugs''
  \item \textit{Avoid All Feelings}: ``avoid anything that could cause negative or positive feelings''
\end{itemize}
\paragraph{Example items, Negative Appraisals.}
\begin{itemize}\setlength\itemsep{0pt}
  \item \textit{Self-Blame}: ``The event happened because of the way I acted''
  \item \textit{Self-Doubt}: ``I can't trust that I will do the right thing''
  \item \textit{Weakness Belief}: ``I am a weak person''
  \item \textit{Fear of Rage}: ``I will not be able to control my anger and will do something terrible''
  \item \textit{Low Coping}: ``I can't deal with even the slightest upset''
\end{itemize}
\paragraph{Example items, Memory characteristics.}
\begin{itemize}\setlength\itemsep{0pt}
  \item \textit{Visual Memories}: ``My memories of the frightening event are mostly pictures or images.''
  \item \textit{Nonverbal Memory}: ``I can't seem to put the frightening event into words.''
  \item \textit{Auditory Recall}: ``When I have memories of what happened I sometimes hear things in my head that I heard during the frightening event.''
  \item \textit{Reliving Event}: ``When I remember the frightening event I feel like it is happening right now.''
  \item \textit{Olfactory Recall}: ``When I think about the frightening event I can sometimes smell things that I smelt when the frightening event happened.''
\end{itemize}
\paragraph{Example items, Current threat appraisals.}
\begin{itemize}\setlength\itemsep{0pt}
  \item \textit{Flashbacks}: ``Suddenly feeling or acting as if the stressful experience were actually happening again''
  \item \textit{Physical Reactivity}: ``Having strong physical reactions when something reminded you of the stressful experience''
  \item \textit{Negative Emotions}: ``Having strong negative feelings such as fear, horror, anger, guilt, or shame''
  \item \textit{Irritability}: ``Irritable behavior, angry outbursts, or acting aggressively''
  \item \textit{Risky Behavior}: ``Taking too many risks or doing things that could cause you harm''
\end{itemize}

\section{Validation Loop Iteration Counts}
\label{app:iterations}

Table~\ref{tab:iterations} reports the distribution of validation-loop iterations per generation model in the full condition. Most models pass on the first attempt; the \texttt{qwen} rows reflect lenient self-validation (Crafter and Validator share the same LLM), and \texttt{gpt-4o-mini} hits the iteration cap due to a markup-compliance failure (\S\ref{sec:vignette-validation}).

\begin{table}[h]
\centering
\small
\caption{Distribution of validation loop iterations per model in the full condition ($n=500$ each). 
Columns show the percentage of vignettes that terminated after exactly 1, exactly 2, or 3 or more iterations, and the largest number of iterations observed. 
Since Crafter and Validator are the same model, the qwen rows reflect lenient self-validation rather than error-free generation.}
\label{tab:iterations}
\resizebox{\linewidth}{!}{%
\begin{tabular}{lrrrr}
\toprule
\textbf{Model} & \textbf{1 iter} & \textbf{2 iters} & \textbf{3+ iters} & \textbf{max} \\
\midrule
qwen2.5-32b           & 100.0\% &  0.0\% &  0.0\% & 1 \\
qwen3.6-35b           & 100.0\% &  0.0\% &  0.0\% & 1 \\
claude-sonnet-4-6     &  97.0\% &  3.0\% &  0.0\% & 2 \\
deepseek-chat         &  94.6\% &  4.6\% &  0.8\% & 5 \\
claude-haiku-4-5      &  85.8\% &  8.0\% &  6.2\% & 5 \\
gpt-5.4-mini          &  85.4\% & 10.6\% &  4.0\% & 6 \\
gpt-5.4               &  84.4\% & 11.4\% &  4.2\% & 6 \\
gemini-2.5-pro        &  66.4\% & 23.8\% &  9.8\% & 4 \\
gemini-2.5-flash      &  60.6\% & 29.6\% &  9.8\% & 5 \\
deepseek-reasoner     &  54.4\% & 32.6\% & 13.0\% & 6 \\
gpt-4o-mini           &   3.8\% &  0.6\% & 95.6\% & 6 \\
\bottomrule
\end{tabular}
}
\end{table}

% ===== Sensitivity Analyses =====

\section{Demographic Independence Tests}
\label{app:chi_square}
The persona pipeline (\S\ref{sec:persona-construction}) samples demographic attributes (gender, ethnicity, age, occupation, relationship status) and clinical attributes (trauma type, PCL-5 severity) from independent categorical distributions, so that downstream condition and fairness comparisons are not confounded by demographic--clinical dependencies that would arise if, say, a particular trauma type were over-represented among one ethnicity. In real clinical samples these variables are correlated (interpersonal-violence histories are unevenly distributed across genders; combat trauma is unevenly distributed across age and occupation), and a generation framework that reproduces those correlations would make condition contrasts uninterpretable: any difference between conditions could be attributable to a confound between demographics and trauma content rather than to the formulation.

We therefore tested every pair $($demographic, clinical$)$ for statistical independence in the realised persona set using Pearson's chi-square test of independence, treating each persona as an observation and binning continuous variables (age) into deciles. Table~\ref{tab:chi_square} reports the test statistic, degrees of freedom, and $p$-value for each pair. No association reaches $p<.05$; the smallest $p$-value (age $\times$ trauma type, $p=.063$) is consistent with the null at this sample size. We conclude that the persona set realises the intended factorial structure: every demographic stratum is approximately equally populated by every trauma type, so condition contrasts within strata in the fairness analysis (\S\ref{sec:fairness}) cannot be confounded by clinical-content imbalance.

A caveat: independence at the marginal level does not guarantee independence at the joint level. Intersectional cells (e.g., ``female~$\times$~South-Asian~$\times$~childhood-trauma'') contain a small number of personas ($\sim 1$--$3$) and we do not test higher-order interactions because they would be underpowered. Reported fairness metrics are accordingly limited to single-axis disparity.

\begin{table}[ht]
\centering
\small
\setlength{\tabcolsep}{6pt}
\caption{Chi-square tests of independence between demographic and clinical variables. No association is significant at $p<.05$.}
\label{tab:chi_square}
\begin{tabular}{l r r r}
\toprule
\textbf{Association} & $\bm{\chi^2}$ & \textbf{\textit{df}} & \textbf{\textit{p}} \\
\midrule
Gender $\times$ Trauma type         & 58.36  & 56  & .389 \\
Gender $\times$ Occupation          &  1.89  &   2 & .388 \\
Ethnicity $\times$ Trauma type      & 186.10 & 196 & .683 \\
Ethnicity $\times$ Occupation       &  5.29  &   7 & .625 \\
Age (binned) $\times$ Trauma type   & 135.77 & 112 & .063 \\
Occupation $\times$ Trauma type     &  27.97 &  28 & .466 \\
\bottomrule
\end{tabular}
\end{table}

\section{Persona Demographic Distributions}
\label{sec:appendix-demographics}
This section gives the full distributional breakdown of the $500$-persona set summarised in \S\ref{sec:personas}. We report exact frequencies because they are the demographic universe over which every downstream condition contrast, fairness analysis, and inter-rater agreement statistic is computed: they fix the population that any disparity claim is defined relative to. Table~\ref{tab:demographics-appendix} consolidates every persona-level variable in a single view; the notes below the table cover sampling-design details that the numbers alone do not communicate.

\begin{table*}[h]
\centering
\small
\captionsetup{font=small}
\caption{Persona demographic distributions across the $500$-persona set ($N = 500$). Continuous variables: mean, SD, range. Categorical variables: counts and percentages over $N = 500$. The eight-region ethnicity scheme is coarse by design (see notes); the occupation bucketing follows the fairness-analysis taxonomy in \S\ref{sec:fairness} (\texttt{occupation\_bucket} in \texttt{code/analysis/fairness\_analysis.py}); the six-category trauma roll-up corresponds to \texttt{trauma\_bucket}. The 29-category trauma list (raw) is summarised below the table.}
\label{tab:demographics-appendix}
\renewcommand{\arraystretch}{1.05}
\begin{tabular}{@{}l l r r@{}}
\toprule
\textbf{Variable} & \textbf{Level} & \textbf{$n$} & \textbf{\%} \\
\midrule
\multicolumn{4}{@{}l}{\textit{Continuous variables}} \\
Age (years)               & $M = 48.5$, $SD = 18.2$, range $18$--$80$              & 500 & 100.0 \\
PCL-5 severity            & $M = 57.8$, $SD = 13.9$, range $33$--$80$              & 500 & 100.0 \\
\midrule
\multicolumn{4}{@{}l}{\textit{Age band}} \\
                          & 18--29                                                 &  95 & 19.0 \\
                          & 30--44                                                 & 125 & 25.0 \\
                          & 45--59                                                 & 121 & 24.2 \\
                          & 60--80                                                 & 159 & 31.8 \\
\midrule
\multicolumn{4}{@{}l}{\textit{Gender}} \\
                          & Male                                                   & 258 & 51.6 \\
                          & Female                                                 & 238 & 47.6 \\
                          & Non-binary                                             &   4 &  0.8 \\
\midrule
\multicolumn{4}{@{}l}{\textit{Ethnicity (8 world regions)}} \\
                          & South-Eastern Asian                                    &  71 & 14.2 \\
                          & European                                               &  70 & 14.0 \\
                          & North American                                         &  68 & 13.6 \\
                          & African                                                &  62 & 12.4 \\
                          & South Asian                                            &  60 & 12.0 \\
                          & Latin American                                         &  59 & 11.8 \\
                          & Oceanian                                               &  57 & 11.4 \\
                          & Middle Eastern                                         &  53 & 10.6 \\
\midrule
\multicolumn{4}{@{}l}{\textit{Relationship status}} \\
                          & Married                                                & 239 & 47.8 \\
                          & Single                                                 &  95 & 19.0 \\
                          & Divorced                                               &  72 & 14.4 \\
                          & In a relationship                                      &  48 &  9.6 \\
                          & Widowed                                                &  46 &  9.2 \\
\midrule
\multicolumn{4}{@{}l}{\textit{Occupation (6-bucket roll-up; $135$ unique free-text titles)}} \\
                          & Professional / tech                                    &  68 & 13.6 \\
                          & Trades                                                 &  63 & 12.6 \\
                          & Healthcare                                             &  34 &  6.8 \\
                          & Uniformed (police/fire/military)                       &  16 &  3.2 \\
                          & Education                                              &  15 &  3.0 \\
                          & Other / unclassified                                   & 304 & 60.8 \\
\midrule
\multicolumn{4}{@{}l}{\textit{Trauma category (6-bucket roll-up of 29 raw categories)}} \\
                          & Interpersonal violence (abuse, assault, stalking)      & 104 & 20.8 \\
                          & War / collective violence (combat, terrorism, torture) &  95 & 19.0 \\
                          & Accident / disaster                                    &  75 & 15.0 \\
                          & Loss / medical                                         &  49 &  9.8 \\
                          & Childhood (non-abuse)                                  &  16 &  3.2 \\
                          & Other / mixed                                          & 161 & 32.2 \\
\bottomrule
\end{tabular}
\end{table*}

\paragraph{Design notes on Table~\ref{tab:demographics-appendix}.}
Categorical demographic sampling follows the reasoning of \citet{suhas2025thousand}: gender, ethnicity and relationship-status weights are grounded in U.S.\ population distributions reported by the \citet{uscensus2025quickfacts}, then perturbed lightly toward category-uniformity so that downstream per-cell fairness tests retain power. \textit{Age} is sampled uniformly over $18$--$80$ rather than from a population-realistic distribution because the goal is a stress test of the framework, not epidemiological representativeness. \textit{Gender} has a slight male majority ($51.6\%$) as an artefact of the categorical sampler — we did not enforce exact parity because that would have over-constrained the joint with other demographics. \textit{Ethnicity} uses an eight-region scheme that conflates national-origin, racial, and cultural categories; we use it because (a) it is the granularity at which downstream LLMs reliably condition on ``ethnicity'' as a generation cue, and (b) finer-grained sampling would render per-cell fairness analyses underpowered. \textit{Relationship status} is sampled conditional on age (e.g., widowed unlikely under $30$) and validated by the Persona Validator agent.

\paragraph{Occupation and trauma roll-ups.} Occupation is sampled as a free-text job title (135 unique titles across the 500 personas); for fairness analysis we bucket those titles into the six categories above using the keyword-based \texttt{occupation\_bucket} routine. The \emph{Other / unclassified} bucket is large ($60.8\%$) because the keyword list is intentionally conservative — it only labels a title with a category if at least one anchor keyword is matched (e.g., a ``Dancer'' or ``Park Ranger'' lands in \emph{Other}). Per-cell fairness tests are run on the five labelled buckets and reported in \S\ref{sec:fairness}.

\paragraph{Trauma categories.} Trauma is sampled across $29$ raw categories spanning interpersonal violence (e.g., physical and sexual abuse, stalking), childhood trauma, war and collective violence (e.g., combat, terrorism, torture), accidental and life-threatening events, and loss/medical trauma. The five most frequent raw categories are physical abuse by a romantic partner ($28$, $5.6\%$), and assault, civilian exposure in a war zone, physical abuse by a caregiver in childhood, and accidentally causing serious injury to another (each $23$, $4.6\%$); the remaining $24$ categories each contribute $14$--$22$ personas. The list of $29$ categories is adapted from the trauma classifications used in the Life Events Checklist (LEC-5) supplemented with childhood-specific items from the ACE inventory \citep{felitti1998relationship}; we widened the LEC-5 set because its $17$-item form aggregates childhood maltreatment into a single ``physical abuse / sexual abuse'' bucket that under-represents the cognitive heterogeneity of early trauma cases. For fairness analysis we roll the $29$ raw categories into the six buckets in the table above.

\paragraph{PCL-5 severity.} The PCL-5 is the standard $20$-item self-report measure of PTSD severity with a clinical cutoff of $33$ \citep{blevins2015posttraumatic}; every persona in our set exceeds this cutoff by design, so the dataset represents a probable-PTSD population rather than a general one.

\section{Mean Token Counts by Model and Condition}
\label{app:tokens}
Table~\ref{tab:tokens} reports the mean total tokens (prompt $+$ completion) per generated vignette by model and condition. Token counts encode two distinct generator behaviours that the body discussion does not separate. First, the formulation graph adds roughly $2{,}000$--$4{,}000$ tokens of specification content (component list, edge list, weights, self-report items) that every model has to process, and the validation loop appends Crafter-generated patches until structural coverage is achieved, so completion tokens also grow. Second, generation models differ widely in how verbose their first-pass vignette is: \texttt{gemini-2.5-flash} and \texttt{gemini-2.5-pro} produce $2{,}700$+ tokens even in zero-shot, while \texttt{qwen2.5-32b} and \texttt{qwen3.6-35b} produce roughly one-third that. The cross-product of these two factors explains the $\sim 14\times$ range in full-condition tokens across models.

The \texttt{gpt-4o-mini} row stands out at $29{,}234$ tokens in the full condition: this is not a verbose generation, but a refinement-loop pathology. Because \texttt{gpt-4o-mini} fails to emit valid patch-block markers (\S\ref{sec:vignette-validation}), the Validator continues to flag the same edge violations across iterations, and each iteration re-sends the prompt plus the accumulated context to the Crafter. The token count thus reflects an interrupted control flow rather than a structurally rich vignette, which is one of the reasons we exclude this model from per-condition analyses.

The token count is not a quality measure: \texttt{gemini-2.5-flash} produces $13{,}651$-token full vignettes and rates highly on both expert and clinician scales, while \texttt{qwen2.5-32b} produces $3{,}284$-token full vignettes and rates substantially lower. We report token counts for reproducibility (downstream users budgeting inference cost) and for transparency about the cost of the cognitive-graph specification; we do not use token counts to filter or weight the per-condition analyses.

\begin{table}[h]
\centering
\small
\caption{Mean total tokens per vignette by model and condition.}
\label{tab:tokens}
\resizebox{\columnwidth}{!}{%
\begin{tabular}{lrrr}
\toprule
\textbf{Model} & \textbf{Zero-shot} & \textbf{No-formulation} & \textbf{Full} \\
\midrule
claude-haiku-4-5    & 1{,}007 & 5{,}768 &  6{,}602 \\
claude-sonnet-4-6   & 1{,}062 & 4{,}214 &  5{,}062 \\
deepseek-chat       & 1{,}050 & 3{,}479 &  4{,}159 \\
deepseek-reasoner   & 1{,}443 & 6{,}108 & 11{,}205 \\
gemini-2.5-flash    & 2{,}779 & 6{,}075 & 13{,}651 \\
gemini-2.5-pro      & 2{,}739 & 6{,}379 & 11{,}024 \\
gpt-4o-mini         & 1{,}091 & 3{,}238 & 29{,}234 \\
gpt-5.4             & 1{,}255 & 3{,}539 &  5{,}187 \\
gpt-5.4-mini        & 1{,}111 & 3{,}472 &  5{,}144 \\
qwen2.5-32b         &    918 & 2{,}790 &  3{,}284 \\
qwen3.6-35b         &    910 & 2{,}919 &  3{,}479 \\
\bottomrule
\end{tabular}%
}
\end{table}

% ===== Generation Hyperparameters and Compute =====

\section{Generation Hyperparameters and Compute}
\label{app:hyperparams}

\textbf{Sampling.} All eleven generation models and the five rating agents (two human-blind expert raters do not apply here; the three LLM judges, the persona Crafter/Validator, the vignette Crafter/Validator, and the external edge probe) use \emph{temperature} $= 0.7$ with provider-default \texttt{top\_p} and provider-default \texttt{max\_tokens}; we did not set explicit nucleus or top-k truncation. Three model families behave slightly differently from the default: (i) the \texttt{claude} family does not accept a \texttt{temperature} parameter via the Anthropic SDK, so for these models we use the API default; (ii) the \texttt{deepseek-reasoner} model uses its built-in reasoning trace and ignores nucleus parameters; (iii) the open-source \texttt{qwen} models are served through an internal lab inference endpoint and use the same temperature setting. Generation seeds are not fixed: we report effect sizes across $500$ personas $\times$ $11$ models $\times$ $3$ conditions, so model-level stochasticity is absorbed into the design rather than controlled out. The validation loop retries up to $5$ times (six iterations total); persona construction retries each demographic field up to $6$ times.

\textbf{Architecture and routing.} Generation requests are routed through provider-specific LangChain wrappers (\texttt{ChatOpenAI}, \texttt{ChatAnthropic}, \texttt{ChatGoogleGenerativeAI}, a custom \texttt{DeepSeekChatModel} pointing at \texttt{api.deepseek.com}, and \texttt{ChatOllama}/custom open-source-LLM clients for \texttt{qwen2.5-32b}, \texttt{qwen3.6-35b}). Structured outputs for the Validator and Judge use LangChain's Pydantic-typed structured-output decoder; we manually retry one time on JSON parse failure before logging the call as a structural-validation failure.

\textbf{Compute.} All eleven generation models, the five LLM judges, and the external edge probe were accessed through commercial APIs (OpenAI, Anthropic, Google, DeepSeek) or through an internal lab inference endpoint (Alibaba \texttt{qwen}); we ran no LLM inference locally for generation or judging. Total token consumption across the $16{,}500$-vignette generation set, the $5{,}000$-vignette scaled-judge pass, the $330$-vignette $3$-LLM-judge pass, and the $6{,}600$-cell edge-probe pass is reported in Appendix~\ref{app:tokens} (per-cell averages). %Approximate end-to-end wall-clock generation time was $\sim 6$ hours per generation-model pass, parallelised at $4$--$8$ concurrent requests.%; total generation cost across all models was on the order of \$$1{,}500$ in API spend. 
Embedding analyses (MPNet, BGE-base) ran on a single NVIDIA RTX 3050 Ti Laptop GPU (4 GB VRAM, CUDA 12.6), $\sim 90$ minutes total wall-clock for the $16{,}500$-vignette BGE pass and a comparable CPU pass for MPNet (used in the body). %The two clinical-expert raters scored the $330$-vignette evaluation subsample in approximately $25$ hours each. The $N=100$-clinician user study had a median per-participant completion time of $70.1$ minutes (Appendix~\ref{app:us-sample}).

% ===== System Prompts =====

\section{System Prompts}
\label{app:instructions}

This section reproduces the system prompts used by the four LLM agents that govern the structural part of the framework: the Vignette Crafter (\S\ref{sec:methodology}; condition: \emph{full}), the Vignette Validator (\S\ref{sec:vignette-validation}), the external Edge Probe (\S\ref{sec:res-structural}), and the EC-coverage Judge (\S\ref{sec:res-structural}). The user-specification templates that wrap these system instructions at inference time are released with the code; the persona Crafter/Validator instructions and the no-formulation / zero-shot variants are deferred to the released repository for space.

\subsection*{Vignette Crafter, full condition}
\begin{lstlisting}
You are a clinical psychologist writing psychological case
vignettes grounded in Ehlers & Clark's (2000) cognitive model
of PTSD.

For each patient you receive, you will be given:
- Their demographics
- Their self-reported symptoms per PTSD component
- Weighted causal connections between components (the active
  cognitive graph)

Core constraints that apply to every vignette:
- Only include active components.
- Do NOT invent or infer causal links not present in the
  active graph.
- For EACH required causal connection (A -> B), ensure the
  vignette conveys -- within a paragraph -- that A influences
  or leads to B.
- Ground every clinical detail in the patient's reported items.
- The causal connections are weighted (0 to 1). These weights
  govern narrative prominence:
    > 0.5  : these connections form the narrative backbone
    0.1-0.5: present and integrated but not load-bearing
    < 0.1  : mentioned briefly and in passing
    = 0    : the component may appear but must be fully isolated
             from all causal language

Output format -- write 500-700 words of continuous third-person
prose. No headers, no numbered sections, no bullet points.
Tell the patient's story chronologically; the dominant weighted
connections (weight > 0.6) should form the narrative backbone
of the second half of the vignette.
\end{lstlisting}

\subsection*{Vignette Validator, full condition}
\begin{lstlisting}
You are a clinical validator checking whether a PTSD case
vignette accurately reflects a patient's required cognitive
connections.

CAUSAL STANDARD
A required edge A->B is SATISFIED if, within the same paragraph:
  - Both A and B (or their synonyms) appear, AND
  - The paragraph conveys -- through any language -- that A
    influences, drives, or leads to B.
This includes direct causal sentences, sequential structure,
or connecting phrases like "because of", "this meant", "so",
"which led", "as a result", "consequently", "in turn".
A required edge is MISSING only if no paragraph conveys the
A->B relationship at all.

SECTION 1 -- COMPONENT CHECK
Confirm each active component appears as a described feature
of the patient's presentation. (Internal check only -- do NOT
add component results to violations.)

SECTION 2 -- REQUIRED EDGES
For each edge, apply the causal standard above.
- If satisfied: do nothing -- only failures are reported.
- If missing: add it to violations and explain in one sentence
  what causal relationship is absent.
\end{lstlisting}

\subsection*{External Edge Probe}
\begin{lstlisting}
You are evaluating a PTSD case vignette for the presence of
causal connections between five Ehlers & Clark cognitive
components: Triggers, Negative Appraisals, Memory, Threat,
and Maladaptive Strategies.

For each of the 20 ordered, directed (A -> B) pairs of distinct
components, decide whether the vignette conveys, in some
paragraph, that A influences, drives, or leads to B. Apply the
same causal standard as the structural validator.

CAUSAL STANDARD
An edge A -> B is PRESENT (return 1) if, within the same
paragraph:
  (i) both A and B (or their synonyms) appear; AND
  (ii) the paragraph conveys that A influences / drives / leads
       to B, through direct causal language, sequential
       structure, or connecting phrases like "because of",
       "this meant", "so", "which led", "as a result",
       "consequently", "in turn".

You will return a binary judgement for each of the 20 ordered
pairs. Be strict: do not assume an edge based on mere
co-occurrence; the paragraph must actually convey directional
influence.
\end{lstlisting}

\subsection*{Schema Judge (used by Opus, DSpro, Flash)}
The schema judge that scores the $18$-item rating protocol (\S\ref{sec:rating-schema}) issues one structured call per item, sharing a common rubric and reading one vignette at a time. The shared header is:
\begin{lstlisting}
You are a critical reviewer of PTSD clinical vignettes for a
research study. Rate the vignette on ONE dimension only, using
the anchors below. Score based solely on what is written; do
not infer content that is absent. Before scoring, identify a
specific sentence that justifies your rating.

Guidelines for a well-written vignette:
  1. Derive from the literature and/or clinical experience
  2. Be clear, well-written, and carefully edited
  3. 500-700 words: enough to show a complex picture without
     being too long to read closely
  4. Follow a narrative, story-like progression
  5. Be culturally and socio-economically neutral
  6. Resemble real people, not a personification of symptoms
  7. Be relatable, relevant, and plausible to clinicians
  8. Avoid red herrings, misleading details, and bizarre content
  9. Highlight the key clinical variables of interest
  10. Allow vague or ambiguous elements that invite reflection
\end{lstlisting}

Per-item anchors are appended to this header. We give one representative item from each of the four families; the remaining $14$ are released with the code.

\paragraph{Clarity (CVI, ordinal $1$--$3$).}
\begin{lstlisting}
Dimension: **Clarity** (1-3) -- easy to understand, free from
confusion, ambiguity, or obstruction.

1 = hard to follow. A short vignette can still be unclear.
    Example: "He was haunted by what happened. The past was a
    shadow that followed him, a weight that pressed down on his
    chest, a storm that raged inside him."
2 = readable but relies on vague or generic clinical language
    without specific detail.
    Example: "He experienced hypervigilance that made things
    hard. He avoided reminders of the trauma and had trouble
    sleeping."
3 = precise and specific: clear what the person does, thinks,
    and feels, and why.
    Example: "He lives with a constant sense of being on guard,
    sits with his back to walls, watches people's hands. When
    triggered, he feels panic mixed with anger."
\end{lstlisting}

\paragraph{Grounded narrative (Construction guideline g1, ordinal $1$--$3$).}
\begin{lstlisting}
Dimension: **g1 -- Grounded in realistic clinical experience**
(1-3).

1 = generic phrasing, no specific detail, could describe any
    PTSD patient.
2 = some realistic detail but described from the outside.
3 = written from inside this person's specific experience: the
    exact situations, paradoxes, and reason they sought help.
\end{lstlisting}

\paragraph{DSM-5 Criterion C (binary, avoidance).}
\begin{lstlisting}
Score 1 only if the criterion is clearly and explicitly present
in the text; not merely inferable. Score 0 if absent or only
vaguely implied.

Dimension: **DSM-5 Criterion C** -- avoidance of trauma-related
thoughts/feelings, or of external reminders.

Score 1 example: "He declines team events in busy restaurants,
takes alternate routes to avoid the coastal expressway, and has
stopped answering automated weather alerts."
Score 0 example: "He preferred quieter environments and did not
like to dwell on the past."
\end{lstlisting}

\paragraph{Sense of current threat (Ehlers \& Clark, binary).}
\begin{lstlisting}
Score 1 only if the component is explicitly shown; not just
consistent with having PTSD.

Dimension: **Sense of current threat** -- the person feels the
trauma is still happening or still dangerous now, not merely a
memory of past danger.

Score 1 example: "He wakes some mornings certain the floor will
give way: not a memory of past danger but a present conviction
that the ground itself is unsafe."
Score 0 example: "He remained cautious in general and found it
hard to fully relax."
\end{lstlisting}

\paragraph{Item user message.} The same one-line user message wraps each item call: \verb|Rate the following clinical vignette:| \verb|{vignette}|. Each call returns a single structured score; the $18$ calls are independent and parallelised per vignette.

\section{Expert-Annotated Vignette (Persona 14)}
\label{app:expert-annotation}

To complement the quantitative evaluation with a qualitative one, a licensed clinical psychologist with formal training in the Ehlers \& Clark cognitive model of PTSD and a clinical specialism in PTSD case formulation annotated a full-condition vignette span by span, tagging each clinically meaningful passage with the EC component it instantiates. The annotation is reproduced verbatim below in the annotator's colour scheme. The exercise is illustrative, not statistical: it shows what the structural recovery results of \S\ref{sec:res-structural} look like at the level of clinical reading.

\paragraph{Persona (sampled inputs).} A $54$-year-old female military combat engineer; trauma type: military / route-clearance explosion; PCL-5 = $62$. Active components: \{Triggers, Negative~Appraisals, Memory, Threat, Maladaptive~Strategies\}. Vignette generated by \texttt{gpt-5.4} under the full condition.

\begin{figure}[H]
    \centering
    \includegraphics[width=\columnwidth]{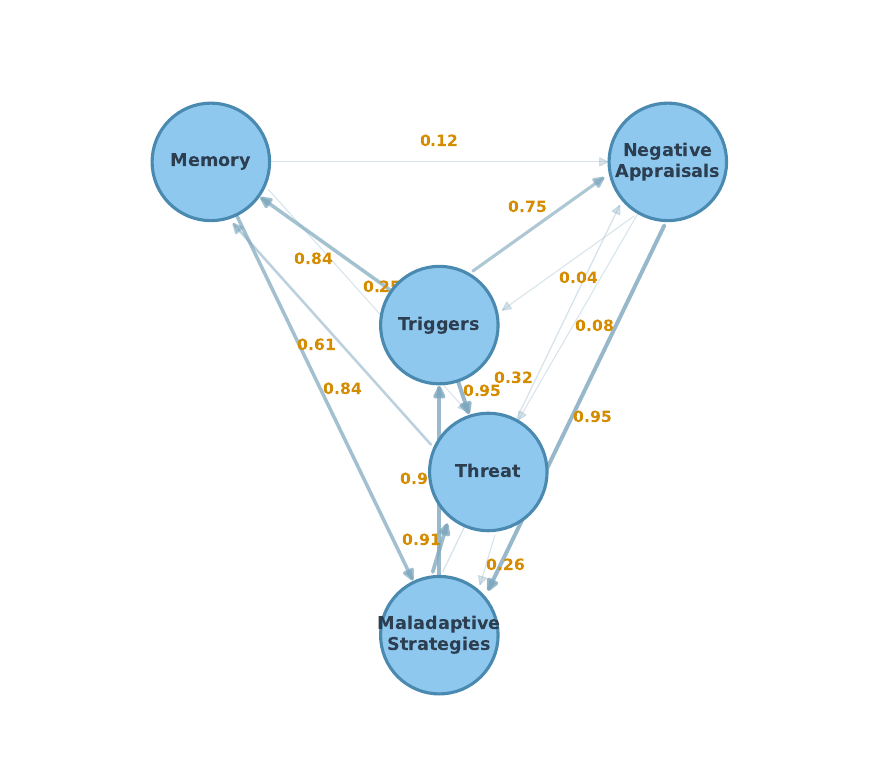}
    \caption{Persona 14 cognitive graph. Active components: \{Triggers, Negative~Appraisals, Memory, Threat, Maladaptive~Strategies\}. }
    \label{fig:persona14-graph}
\end{figure}

\begin{table}[ht]
\centering
\footnotesize
\caption{Self report items per component of persona 14}
\label{tab:self report persona 14}
\setlength{\tabcolsep}{4pt}
\renewcommand{\arraystretch}{1.1}
\begin{tabular}{p{2.4cm} p{4.8cm}}
\hline
\textbf{Component} & \textbf{Items} \\
\hline
\multirow{3}{2.4cm}{Triggers}
 & \textbullet\ Fire or flames \\
 & \textbullet\ A specific temperature on the skin \\
 & \textbullet\ Shouting \\
\hline
\multirow{3}{2.4cm}{Maladaptive Strategies}
 & \textbullet\ Avoid anything that could be stressful \\
 & \textbullet\ I numb my feelings \\
 & \textbullet\ Constantly scan environment for signs of threat \\
\hline
\multirow{3}{2.4cm}{Threat}
 & \textbullet\ My body feels constantly tense and ready to react to danger \\
 & \textbullet\ Unusual silence makes me feel something dangerous is about to happen \\
 & \textbullet\ Having difficulty concentrating \\
\hline
\multirow{3}{2.4cm}{Memory}
 & \textbullet\ Something in my surroundings can instantly bring back the full memory without warning \\
 & \textbullet\ When I have memories of what happened I sometimes hear things in my head that I heard during the frightening event \\
 & \textbullet\ I cannot tell a clear story of what happened even to myself \\
\hline
\multirow{3}{2.4cm}{Negative Appraisals}
 & \textbullet\ There is something wrong with me as a person \\
 & \textbullet\ What happened to me was deeply unfair and nothing has been done about it \\
 & \textbullet\ The world is a dangerous place \\
\hline
\end{tabular}
\end{table}

\paragraph{Legend.} \hltrig{Triggers} \quad \hlmem{Memory} \quad \hlapp{Negative Appraisals} \quad \hlstrat{Maladaptive Strategies} \quad \hlthreat{Sense of current Threat}

\begin{quote}\small\raggedright
At $54$, Marisol Alvarez had spent most of her adult life in the Army as a combat engineer, and she still talked about deployment in fragments rather than as a story with a beginning and end. The event she circled around happened during a route-clearance mission overseas, when an explosion ignited a vehicle ahead of hers. \hlmem{She remembered the sudden bloom of flames}, \hlmem{the searing heat striking the exposed skin of her face and forearms}, \hlmem{and men shouting over one another in the confusion}. \hlmem{Those moments were preserved in sharp pieces rather than a narrative}. \hlmem{She could not put them into a clear sequence even in the privacy of her own mind}; \hlmem{what returned instead were flashes and sounds}, especially the shouted commands and panic she heard that day.

Back home, those same sensory details became potent in ordinary life. At work, if she passed a training burn pit or caught sight of fire during demolition exercises, \hlthreat{her body reacted before she could think}. \hltrig{A burst of heat from industrial equipment against her skin or a soldier yelling across} \hlthreat{the motor pool could make her go rigid and instantly ready for danger}. What unnerved her just as much was the silence that sometimes followed routine activity; \hlthreat{an unusual quiet in a work area made her feel that something terrible was about to break loose}. In those moments she was not only tense but unable to focus on what was in front of her. \hltrig{The trigger could also bring back the event whole and without warning}, \hltrig{so that a smell of heat or the sight of flames was enough for her} \hlmem{to hear the shouting in her head again as if she were back there}.

\hlapp{Those sudden returns of memory shaped what she believed about herself and the world}. Each time the scene intruded so forcefully, \hlapp{Marisol took it as proof that something was wrong with her}, that other people could move on from war and she could not. \hlapp{She had come to believe the world was fundamentally dangerous}, \hlapp{not just in combat but everywhere}, \hlapp{and the randomness of what happened felt intolerably unfair to her}. \hltrig{Seeing fire or hearing shouting did not simply startle her}; \hlthreat{it seemed to confirm that danger was always near} \hlapp{and that she had been left to carry an injustice no one had put right}. \hltrig{Over time she became sensitized to those cues themselves}, \hltrig{scanning for heat}, noise, \hltrig{and changes in the atmosphere} \hlapp{as if her mind were teaching itself where the next threat would come from}.

Her beliefs then began to govern how she lived. \hlapp{Because she was certain the world was dangerous and because she felt flawed for reacting so strongly}, \hlstrat{she tried to avoid anything that might be stressful}. \hlstrat{She turned down optional field exercises}, \hlstrat{lingered in administrative tasks when she could}, \hlstrat{and at home steered away from family cookouts where open flames and loud voices were likely}. \hlstrat{She worked hard to numb her feelings}, \hlstrat{speaking flatly when her husband asked what was wrong and trying to shut down whatever rose up inside her}. \hlstrat{She also checked her surroundings constantly, in the office, on base, in parking lots, noticing exits, listening for changes in sound}, \hlstrat{looking for anything that might signal threat}. \hlapp{That constant management of herself made her feel even more unlike the person she used to be}, \hlapp{reinforcing her conviction that there was something defective in her}.

\hlmem{The memories themselves fed this pattern}. \hlmem{Because they arrived without warning and brought the shouting back so vividly}, \hlstrat{Marisol began organizing her days around not being caught off guard}. \hlstrat{She avoided stressors and kept herself emotionally blunted because she feared that any strong feeling would open the door to the full memory}. \hlthreat{Yet the more she lived in this narrowed, watchful way, the more tense her body remained}. \hlthreat{She was on alert so much of the time that concentration became difficult even during routine engineering planning}, and once she was already keyed up, \hltrig{the next silence}, shout, \hltrig{or wave of heat struck with even more force}. \hlthreat{Hypervigilance made} \hlmem{the memories easier to set off}; \hlmem{the memories intensified} \hlthreat{her sense of present threat}.

Over months this tightened into the life she now found herself living. \hlstrat{Her constant scanning meant she noticed every flare}, \hltrig{every raised voice}, \hltrig{every change in temperature}, \hltrig{which made those triggers feel omnipresent}. \hlstrat{Avoiding stress and numbing herself had shrunk her world} \hlthreat{but had not made it feel safer}. \hlapp{Instead, the restrictions seemed to prove that the world really was dangerous and that she really was damaged by what had happened}. \hlapp{Her marriage had grown strained by her distance}, \hlapp{and at work she was ashamed that she could no longer concentrate as she once had in the very environment where she was expected to lead calmly}.

Marisol finally came for help after \hlthreat{freezing during a stateside training exercise} \hltrig{when flames rose unexpectedly and someone shouted for equipment}. \hlstrat{She recovered quickly enough that others may not have understood what happened}, \hlthreat{but she spent the rest of the day convinced she could hear the old voices in her head}. \hlthreat{What frightened her most was not just the memory itself but the sense that her life had become organized around waiting for it}. She said she wanted, more than anything, \hlthreat{to be able to stand in her own work environment without feeling that danger was already there}, \hlapp{and to understand whether the part of her that felt permanently wrong could still be reached}.
\end{quote}

\paragraph{What the expert annotation shows.} All five EC components are concretely instantiated, including spans the quantitative judge can recognise on its own: explicit \emph{Triggers} (heat, flame, shouting) in paragraph $1$; \emph{Memory} characterised as fragmented and present-tense rather than narrative (paragraphs $1$--$2$); catastrophic \emph{Negative Appraisals} about the self and the world (paragraph $3$); \emph{Maladaptive Strategies} of avoidance and emotional numbing (paragraphs $4$--$5$); and \emph{Sense of current Threat} as bodily readiness for danger and hypervigilance (paragraphs $2$, $5$). The annotator also marks the maintenance cycle the \citet{ehlers2000cognitive} model predicts: \emph{Triggers} reactivate \emph{Memory}, \emph{Memory} feeds \emph{Negative Appraisals}, \emph{Negative Appraisals} drive \emph{Maladaptive Strategies}, and \emph{Strategies} ($+$ hypervigilance) sustain \emph{Sense of current Threat}, which in turn keeps the system primed for the next \emph{Trigger}.
Figure~\ref{fig:persona14-graph} makes the edge structure concrete. The highest-weight edges have explicit textual counterparts.
 \textit{"a burst of heat\ldots could make her go rigid and instantly ready for danger''} grounds Triggers$\to$Threat ($w{=}0.95$); 
 \textit{"the trigger could also bring back the event whole and without warning\ldots as if she were back there''} grounds Triggers$\to$Memory ($w{=}0.84$); 
 \textit{"those sudden returns of memory shaped what she believed about herself and the world''} grounds Triggers$\to$Negative Appraisals ($w{=}0.75$);
\textit{"because she was certain the world was dangerous and because she felt flawed for reacting so strongly''} grounds Negative Appraisals$\to$Maladaptive Strategies ($w{=}0.95$); and 
\textit{"her constant scanning meant she noticed every flare''} closes the maintenance loop via Maladaptive Strategies$\to$Triggers ($w{=}0.93$).
By contrast, the one edge-span whose source component carries a low ground-truth weight -- \textit{looking for anything that might signal threat''} annotated as Negative Appraisals$\to$Threat ($w{=}0.08$) -- is a borderline reading: the hypervigilance it describes is more directly attributable to the Threat node itself than to an appraisal driving threat.
The clinically informed reading converges with the structural-fidelity and edge-recovery results in \S\ref{sec:res-structural}.

% ===== Geometric Structure of the Generated Vignettes =====

\section{Per-Rater EC-Component Recovery on the 330-Vignette Subsample}
\label{app:gt-recovery-330}
The per-rater EC-recovery analysis extends the scaled-judge analysis (Appendix~\ref{app:judge-vs-specification}) to all five raters on the $330$-vignette evaluation subsample ($n = 110$ per condition; $11$ generation models $\times$ $30$ personas per model). It answers two questions the body cannot: (a) whether the recovery gradient is judge-specific or replicates across human and LLM raters, and (b) whether human and LLM raters disagree systematically on which components are present.

The structure of Table~\ref{tab:gt-recovery-330} is: rows are EC components, columns are the five raters (R1, R2 are clinical experts; Opus = \texttt{claude-opus-4-5}; DSpro = \texttt{deepseek-v4-pro}; Flash = \texttt{deepseek-v4-flash}), and each rater contributes five metrics (P, R, F1, MCC, AC1). The three blocks (full / no-formulation / zero-shot) are stacked vertically.

Two patterns repeat across raters. First, recovery in the full condition is positive and substantial for every rater on every component: every full-block MCC is above zero, AC1 ranges $0.60$--$0.92$, and the human experts (R1, R2) recover the components on par with or above the LLM judges on most components (e.g., Memory: R1 AC1 $= 0.85$, Opus $= 0.82$; Threat: R1 AC1 $= 0.86$, Opus $= 0.92$). Second, recovery in the zero-shot block collapses for every rater: human-expert AC1 falls into negative territory on Memory ($\sim -0.05$ to $+0.18$), Threat ($\sim -0.53$), and Appraisals ($\sim -0.12$ to $+0.06$), while LLM judges retain a slight positive AC1 on Triggers (a base-rate effect: LLMs almost always say Triggers are present) but otherwise also collapse. The MCC pattern is the same and more interpretable because MCC is invariant to base-rate inflation.

These patterns are useful for two reasons. First, they show that the full-condition recovery is not a property of a single LLM judge: humans and LLMs both find the specified components when they are explicitly cued into the text, and both fail to find them when they are not. The recovery effect is a property of the text, not of the rater. Second, the relative ordering within the full block (e.g., Memory recovery: R1 AC1 $= 0.85 >$ Opus $= 0.82 >$ Flash $= 0.72$) gives a noisy but informative ranking of rater sensitivity that we use to calibrate the scaled-judge analysis on the full $5{,}000$-vignette set.

Caveats: the no-formulation block on the two human experts is mixed (some components recover, others do not), which we attribute to the experts' weighting of explicit-mention versus implicit-context evidence in the rating instructions; LLM raters use the explicit-mention criterion more strictly.

\begin{table*}[h]
\centering
\scriptsize
\setlength{\tabcolsep}{2.5pt}
\caption{Recovery of specified EC components on the 330-vignette subsample, per rater and per condition ($n = 110$ per condition). Columns: P = precision, R = recall, F1, MCC, AC1 (Gwet's). The zero-shot block is a negative control: the specified GT was not communicated to the Crafter in that condition, so recovery should be at chance.}
\label{tab:gt-recovery-330}
\resizebox{\linewidth}{!}{%
\begin{tabular}{l rrrrr rrrrr rrrrr rrrrr rrrrr}
\toprule
& \multicolumn{5}{c}{\textbf{R1}} & \multicolumn{5}{c}{\textbf{R2}} & \multicolumn{5}{c}{\textbf{Opus}} & \multicolumn{5}{c}{\textbf{DSpro}} & \multicolumn{5}{c}{\textbf{Flash}} \\
\cmidrule(lr){2-6} \cmidrule(lr){7-11} \cmidrule(lr){12-16} \cmidrule(lr){17-21} \cmidrule(lr){22-26}
\textbf{EC} & P & R & F1 & MCC & AC1 & P & R & F1 & MCC & AC1 & P & R & F1 & MCC & AC1 & P & R & F1 & MCC & AC1 & P & R & F1 & MCC & AC1 \\
\midrule
\multicolumn{26}{l}{\textit{Full condition}} \\
Triggers           & .77 & .95 & .85 & .38 & .64 & .80 & .95 & .87 & .49 & .69 & .76 &1.00 & .86 & .43 & .67 & .73 & .96 & .83 & .24 & .60 & .73 & .99 & .84 & .28 & .63 \\
Neg. Appraisals    & .85 & .90 & .87 & .29 & .70 & .84 & .89 & .86 & .23 & .68 & .84 & .97 & .90 & .35 & .77 & .83 & .91 & .87 & .22 & .70 & .84 & .95 & .89 & .32 & .76 \\
Memory             & .92 & .94 & .93 & .23 & .85 & .91 & .92 & .92 & .10 & .81 &1.00 & .86 & .92 & .61 & .82 & .99 & .78 & .87 & .45 & .69 &1.00 & .79 & .88 & .52 & .72 \\
Threat             & .92 & .95 & .94 & .26 & .86 & .91 & .80 & .85 & .05 & .66 & .93 & .99 & .96 & .51 & .92 & .95 & .83 & .89 & .34 & .74 & .95 & .83 & .89 & .34 & .74 \\
Strategies         & .79 & .96 & .86 & .46 & .68 & .79 & .96 & .86 & .46 & .68 & .73 & .99 & .84 & .28 & .63 & .74 & .99 & .85 & .35 & .65 & .73 &1.00 & .84 & .30 & .63 \\
\midrule
\multicolumn{26}{l}{\textit{No-formulation condition}} \\
Triggers           & .68 & .78 & .73 & $-$.08 & .34 & .70 & .71 & .70 & $-$.01 & .29 & .72 &1.00 & .84 & .26 & .62 & .72 & .95 & .82 & .17 & .58 & .72 & .95 & .82 & .17 & .58 \\
Neg. Appraisals    & .78 & .68 & .73 & $-$.08 & .34 & .76 & .58 & .66 & $-$.12 & .17 & .82 & .76 & .79 & .07 & .50 & .82 & .85 & .84 & .13 & .62 & .81 & .80 & .81 & .07 & .54 \\
Memory             & .87 & .77 & .82 & $-$.17 & .58 & .86 & .71 & .78 & $-$.20 & .48 &1.00 & .81 & .89 & .54 & .75 &1.00 & .70 & .82 & .43 & .57 &1.00 & .76 & .86 & .49 & .67 \\
Threat             & .89 & .67 & .76 & $-$.04 & .44 & .90 & .46 & .60 & .00 & .05 & .94 & .89 & .91 & .29 & .80 & .93 & .64 & .75 & .11 & .41 & .90 & .71 & .79 & $-$.01 & .51 \\
Strategies         & .62 & .64 & .63 & $-$.28 & .10 & .68 & .57 & .62 & $-$.06 & .10 & .83 & .99 & .90 & .62 & .76 & .77 & .91 & .83 & .33 & .60 & .84 & .77 & .80 & .41 & .53 \\
\midrule
\multicolumn{26}{l}{\textit{Zero-shot condition (negative control)}} \\
Triggers           & .68 & .66 & .67 & $-$.06 & .21 & .70 & .69 & .69 & $-$.01 & .26 & .69 & .91 & .79 & $-$.05 & .50 & .72 & .91 & .81 & .13 & .54 & .69 & .95 & .80 & $-$.13 & .53 \\
Neg. Appraisals    & .82 & .36 & .50 & .04 & $-$.12 & .86 & .48 & .61 & .13 & .09 & .87 & .46 & .60 & .15 & .06 & .85 & .47 & .60 & .12 & .07 & .80 & .46 & .58 & .00 & .01 \\
Memory             & .90 & .53 & .66 & $-$.01 & .18 & .90 & .53 & .66 & $-$.01 & .18 & .92 & .43 & .59 & .04 & .01 & .93 & .49 & .65 & .08 & .14 & .90 & .47 & .61 & .01 & .07 \\
Threat             & .90 & .17 & .29 & $-$.01 & $-$.52 & .82 & .18 & .30 & $-$.14 & $-$.53 & .90 & .69 & .78 & $-$.03 & .47 & .88 & .60 & .71 & $-$.08 & .31 & .86 & .61 & .71 & $-$.19 & .31 \\
Strategies         & .74 & .38 & .50 & .07 & $-$.05 & .72 & .53 & .61 & .04 & .10 & .73 & .60 & .66 & .08 & .19 & .71 & .71 & .71 & .05 & .31 & .67 & .47 & .55 & $-$.07 & $-$.04 \\
\bottomrule
\end{tabular}
}
\end{table*}

\section{Scaled-judge Component-Level Recovery}
\label{app:judge-vs-specification}
The scaled-judge component recovery analysis tests whether the \emph{specified} active EC components are detectable in the generated text without telling the judge what was specified. Concretely, for each persona we record \texttt{active\_nodes}~$\subseteq$~\{Triggers, Negative Appraisals, Memory, Threat, Strategies\}, which is the set of EC components that the persona's sampled cognitive graph activates. In the full condition the Crafter is told which components are active and asked to portray them in the text; in the no-formulation condition the Crafter receives the persona's self-report items but not the active-nodes labels (so any component-level signal in the text reflects what bleeds through from the items); in the zero-shot condition the Crafter receives only the demographic prompt, so component-level signal should be at chance.

The scaled judge \texttt{deepseek-v4-flash} reads each generated vignette and produces a binary judgement for each of the five components. We treat the specified \texttt{active\_nodes} as the gold label and the judge's binary output as the prediction, and compute precision, recall, F1, balanced accuracy, MCC, and Gwet's AC1 per (component, condition) cell. The base rates of the gold are not balanced (Memory, Threat, and Strategies are present in $40$--$60\%$ of personas; Triggers and Negative Appraisals in $70$--$90\%$), which is why we report MCC and BAcc alongside F1: F1 is upward-biased on imbalanced data and would mask the recovery collapse on rare-positive components.

Table~\ref{tab:judge-vs-specification} shows the expected gradient. In full, MCC ranges $+0.30$--$+0.62$ across components; in no-formulation, MCC drops to $+0.16$--$+0.53$; in zero-shot, MCC is at chance for four of five components ($-0.01$--$+0.02$). The exception in zero-shot is Triggers, which retains AC1 $= 0.57$ even without explicit specification: Triggers is the most narratively familiar PTSD construct and any zero-shot vignette tends to mention them. This matches the body claim that the formulation's largest payoffs are on the less narratively obvious components (Strategies, Appraisals, Memory, Threat) and the smallest on Triggers.

Two limitations of this gold standard. First, \texttt{active\_nodes} is a binary summary of a graded persona: a persona with weight $0.05$ on Memory ``has Memory active'' by the threshold but may produce only a single sentence of memory content, which a judge can reasonably miss. Second, the judge is itself an LLM with its own systematic biases; we use Gwet's AC1 rather than Cohen's $\kappa$ here because AC1 is more robust to the prevalence/marginal imbalance typical of EC labels.

\begin{table*}[h]
\centering
\small
\setlength{\tabcolsep}{4pt}
\caption{Recovery of specified EC components by the scaled judge \texttt{deepseek-v4-flash}, per condition ($n = 5{,}000$ per condition; 10 generation models $\times$ 500 personas, \texttt{gpt-4o-mini} excluded). The specified ``ground truth'' is the persona's \texttt{active\_nodes} set, which is explicitly conveyed to the Crafter only in the full condition; the no-formulation and zero-shot columns measure how much the persona's underlying cognitive profile still bleeds into the vignette without explicit specification, and serve as a negative control. \textbf{P}: precision; \textbf{R}: recall; \textbf{F1}: harmonic mean; \textbf{BAcc}: balanced accuracy; \textbf{MCC}: Matthews correlation; \textbf{AC1}: Gwet's AC1.}
\label{tab:judge-vs-specification}
\resizebox{\linewidth}{!}{%
\begin{tabular}{l rrrrrr rrrrrr rrrrrr}
\toprule
& \multicolumn{6}{c}{\textbf{Full}} & \multicolumn{6}{c}{\textbf{No-formulation}} & \multicolumn{6}{c}{\textbf{Zero-shot}} \\
\cmidrule(lr){2-7} \cmidrule(lr){8-13} \cmidrule(lr){14-19}
\textbf{EC} & P & R & F1 & BAcc & MCC & AC1 & P & R & F1 & BAcc & MCC & AC1 & P & R & F1 & BAcc & MCC & AC1 \\
\midrule
Triggers               & .76 & .99 & .86 & .61 & .39 & .68 & .75 & .97 & .85 & .58 & .28 & .64 & .71 & .95 & .81 & .50 & $-.01$ & .57 \\
Neg. Appraisals        & .81 & .99 & .89 & .73 & .59 & .75 & .79 & .97 & .87 & .69 & .49 & .70 & .71 & .53 & .60 & .51 & .01 & .07 \\
Memory                 & .93 & .80 & .86 & .83 & .62 & .66 & .90 & .78 & .83 & .78 & .53 & .60 & .70 & .42 & .53 & .50 & $-.00$ & $-.05$ \\
Threat                 & .77 & .83 & .80 & .64 & .30 & .51 & .74 & .74 & .74 & .58 & .16 & .37 & .69 & .67 & .68 & .51 & .02 & .24 \\
Strategies             & .74 & .99 & .85 & .62 & .41 & .64 & .75 & .81 & .78 & .62 & .25 & .47 & .67 & .48 & .56 & .49 & $-.01$ & .01 \\
\bottomrule
\end{tabular}
}
\end{table*}

\section{Per-Edge Recovery in the Full Condition}
\label{app:edge-recovery-per-edge}
The body's edge-recovery analysis (\S\ref{sec:res-structural}, Table~\ref{tab:edge-recovery-by-cond}) aggregates the external probe over all $20$ directed Ehlers~\&~Clark edges. Table~\ref{tab:edge-recovery-per-edge} disaggregates it: for each of the $20$ edges $A \to B$, we report precision, recall, F1, MCC, and AC1 on the $n = 110$ full-condition vignettes, treating the persona's specified edge weight $w_{A \to B} > 0$ as the gold positive label and the external probe's binary judgement as the prediction. The table is sorted by descending MCC.

Every one of the $20$ directed edges recovers above chance (MCC $> 0$): there is no edge that the formulation specification fails to imprint on the text. The strongest recovery is concentrated on the four edges that the Ehlers~\&~Clark theory explicitly identifies as maintaining the disorder: Strategies~$\to$~Threat (MCC $+0.66$), Triggers~$\to$~Negative Appraisals ($+0.63$), Threat~$\to$~Memory ($+0.57$), and Threat~$\to$~Negative Appraisals ($+0.56$). These are the edges that close the maintenance cycle in the cognitive model, and the cycle is the part of the theory that case-formulation training emphasises; it is unsurprising that a model specified with the graph and the weights chooses to instantiate these edges most explicitly.

The weakest recovery is on edges with high judge-marginal rates: Triggers~$\to$~Threat (MCC $+0.34$, recall $1.00$ but precision $0.47$) and Threat~$\to$~Strategies (MCC $+0.25$, precision $0.56$). In both cases the probe tends to read the connection from any narrative that mentions both endpoints, regardless of whether the edge was specified active: the precision drops because the judge cannot easily distinguish ``$A$ and $B$ both appear'' from ``$A$ leads to $B$.'' This is a property of the probe and the text, not of the graph: short clinical vignettes provide limited contextual evidence for direction even when the underlying generation conditioned on a directed edge. The unit-level precision is therefore a lower bound on the true edge realisation rate.

A separate caveat: the gold standard treats edge presence as binary ($w > 0$ vs $w = 0$) and ignores weight magnitude. The body's right-block analysis in Table~\ref{tab:edge-recovery-by-cond} shows that within specified-active edges, the judge's recovery probability is roughly flat across weight bands ($0.74$--$0.84$), consistent with the formulation acting as a near-binary on/off switch in this configuration rather than as a continuous prominence dial.

\begin{table}[h]
\centering
\small
\setlength{\tabcolsep}{4pt}
\caption{Per-directed-edge external-probe metrics, full condition only ($n = 110$ vignettes per edge). P = precision, R = recall, F1, MCC = Matthews correlation, AC1 = Gwet's. Edges sorted by descending MCC. ``Neg.App.''~$=$ Negative Appraisals; ``Strat.''~$=$ Maladaptive Strategies.}
\label{tab:edge-recovery-per-edge}
\resizebox{\linewidth}{!}{%
\begin{tabular}{l rrrrr}
\toprule
\textbf{Directed edge $A\to B$} & \textbf{P} & \textbf{R} & \textbf{F1} & \textbf{MCC} & \textbf{AC1} \\
\midrule
Strat.~$\to$~Threat            & 0.86 & 0.86 & 0.86 & $+0.66$ & $+0.69$ \\
Triggers~$\to$~Neg.App.        & 0.87 & 0.82 & 0.84 & $+0.63$ & $+0.65$ \\
Threat~$\to$~Memory            & 0.91 & 0.58 & 0.71 & $+0.57$ & $+0.54$ \\
Threat~$\to$~Neg.App.          & 0.89 & 0.71 & 0.79 & $+0.56$ & $+0.55$ \\
Triggers~$\to$~Memory          & 0.66 & 0.94 & 0.78 & $+0.51$ & $+0.48$ \\
Neg.App.~$\to$~Triggers        & 0.97 & 0.48 & 0.65 & $+0.49$ & $+0.37$ \\
Threat~$\to$~Triggers          & 0.88 & 0.51 & 0.64 & $+0.48$ & $+0.46$ \\
Strat.~$\to$~Memory            & 0.72 & 0.76 & 0.74 & $+0.47$ & $+0.47$ \\
Memory~$\to$~Neg.App.          & 0.80 & 0.92 & 0.85 & $+0.44$ & $+0.65$ \\
Neg.App.~$\to$~Memory          & 0.69 & 0.57 & 0.62 & $+0.42$ & $+0.49$ \\
Neg.App.~$\to$~Threat          & 0.57 & 0.82 & 0.67 & $+0.41$ & $+0.36$ \\
Memory~$\to$~Triggers          & 0.91 & 0.34 & 0.50 & $+0.39$ & $+0.37$ \\
Triggers~$\to$~Strat.          & 0.42 & 0.94 & 0.58 & $+0.38$ & $+0.20$ \\
Memory~$\to$~Strat.            & 0.66 & 0.76 & 0.71 & $+0.37$ & $+0.37$ \\
Strat.~$\to$~Triggers          & 0.44 & 0.55 & 0.49 & $+0.35$ & $+0.65$ \\
Strat.~$\to$~Neg.App.          & 0.44 & 0.82 & 0.57 & $+0.35$ & $+0.29$ \\
Triggers~$\to$~Threat          & 0.47 & 1.00 & 0.64 & $+0.34$ & $+0.15$ \\
Memory~$\to$~Threat            & 0.70 & 0.82 & 0.76 & $+0.32$ & $+0.42$ \\
Neg.App.~$\to$~Strat.          & 0.58 & 0.93 & 0.71 & $+0.32$ & $+0.32$ \\
Threat~$\to$~Strat.            & 0.56 & 0.89 & 0.69 & $+0.25$ & $+0.26$ \\
\bottomrule
\end{tabular}
}
\end{table}

% ===== Human edge-level validation =====

\section{Human Validation of Directed-Edge Recovery}
\label{app:human-edge-validation}
This appendix documents the human edge-annotation study summarised in \S\ref{sec:res-human-edge}. Its purpose is to check the body's central structural claim, that the specified \emph{directed} links are preserved in the text, against human judgement rather than against an LLM probe alone. The experts in \S\ref{sec:rating-schema} score EC components; this study scores the directed edges themselves, at a scale comparable to the automated probe.

\paragraph{Sample.} We drew $50$ personas and took each under all three conditions, giving $150$ vignettes. The sample is balanced across the ten retained generation models ($5$ vignettes per model per condition; \texttt{gpt-4o-mini} is excluded here as it is throughout the paper), and every persona appears under all three conditions, so condition contrasts are within-persona.

\paragraph{Annotation schema.} For each vignette the rater is shown the text alone and asked, for each of the $20$ directed Ehlers~\&~Clark pairs $A \to B$, (i) whether the vignette expresses that directed causal link, as a binary judgement, and (ii) how prominent the link is in the narrative, on a graded $0$--$1$ scale. This is $20$ judgements per vignette and $3{,}000$ per rater across the sample. Raters are blind both to the generating condition and to the persona's specified cognitive graph, so the specified weights cannot be read back from the task itself.

\paragraph{Raters.} Two clinical raters (H1, H2) completed the full sample independently. The same $150$ vignettes were then scored on the identical schema by two of the LLM judges already used in the paper, \texttt{deepseek-v4-flash} and \texttt{deepseek-v4-pro} (\S\ref{sec:rating-schema}), under the same blinding. This yields a four-rater panel of two humans and two models, so that human-to-human agreement can be compared directly against human-to-model agreement on identical items.

\paragraph{Agreement.} Table~\ref{tab:human-edge-agreement} reports all six pairwise agreements, on the presence scale with Gwet's AC1 (the statistic used throughout the paper, Appendix~\ref{app:full-pairwise}) and on the graded scale with ICC$(2,1)$. Every pair agrees between $0.63$ and $0.75$ on presence. Human-to-LLM agreement ($0.647$ to $0.699$, mean $0.673$) is at least as high as the agreement between the two human raters ($0.631$), and the same ordering holds on the graded scale (human-to-LLM ICC $0.671$ to $0.717$; human-to-human $0.671$). By this measure the automated probe is not distinguishable from an additional human annotator.

\begin{table}[h]
\centering
\small
\caption{Pairwise agreement between the four raters on the $150$-vignette edge-annotation sample, pooled over conditions ($n = 3{,}000$ directed-edge judgements per pair). AC1 is computed on the binary presence judgement, ICC$(2,1)$ on the graded $0$--$1$ prominence rating.}
\label{tab:human-edge-agreement}
\begin{tabular}{l cc}
\toprule
\textbf{Rater pair} & \textbf{AC1 (present)} & \textbf{ICC (strength)} \\
\midrule
H1 vs H2                        & $0.631$ & $0.671$ \\
\midrule
H1 vs \textsc{Flash}            & $0.680$ & $0.671$ \\
H1 vs \textsc{Pro}              & $0.699$ & $0.703$ \\
H2 vs \textsc{Flash}            & $0.647$ & $0.701$ \\
H2 vs \textsc{Pro}              & $0.668$ & $0.717$ \\
\midrule
\textsc{Flash} vs \textsc{Pro}  & $0.748$ & $0.774$ \\
\bottomrule
\end{tabular}
\end{table}

\paragraph{Recovery of the specified graph.} Table~\ref{tab:human-edge-validation-full} gives the body's Table~\ref{tab:human-edge-validation} at three decimals. Treating the persona's specified active edges ($w > 0$) as gold, all four raters recover the graph under full (MCC $+0.318$ to $+0.408$) and are at chance under both ablations (MCC $-0.023$ to $+0.056$). The AUC of the specified weight predicting detection is $0.645$ to $0.689$ under full against $0.487$ to $0.535$ under the ablations, and the graded strength ratings correlate with the specified weight under full (Spearman $+0.341$ to $+0.432$) but not under either ablation ($-0.015$ to $+0.084$). The human raters therefore reproduce, by hand and without sight of the graph, the recovery gradient the probe reports in \S\ref{sec:res-structural} (MCC $+0.41$, AUC $0.70$).

\begin{table}[h]
\centering
\small
\setlength{\tabcolsep}{4pt}
\caption{Recovery of the specified cognitive graph by each of the four raters on the $150$-vignette sample, at three decimals. Gold is the persona's specified active edge set ($w > 0$).}
\label{tab:human-edge-validation-full}
\begin{tabular}{l ccc}
\toprule
& \textbf{Full} & \textbf{No-form.} & \textbf{Zero-shot} \\
\midrule
\multicolumn{4}{l}{\textit{MCC vs specified graph}} \\
H1                  & $+0.408$ & $+0.056$ & $+0.044$ \\
H2                  & $+0.318$ & $+0.043$ & $+0.007$ \\
\textsc{Flash}      & $+0.319$ & $+0.035$ & $-0.023$ \\
\textsc{Pro}        & $+0.379$ & $-0.012$ & $-0.020$ \\
\midrule
\multicolumn{4}{l}{\textit{AUC (specified weight $\to$ detection)}} \\
H1                  & $0.689$ & $0.535$ & $0.530$ \\
H2                  & $0.648$ & $0.525$ & $0.508$ \\
\textsc{Flash}      & $0.645$ & $0.517$ & $0.491$ \\
\textsc{Pro}        & $0.672$ & $0.499$ & $0.487$ \\
\midrule
\multicolumn{4}{l}{\textit{Spearman (rated strength, specified weight)}} \\
H1                  & $+0.432$ & $+0.084$ & $+0.051$ \\
H2                  & $+0.364$ & $+0.062$ & $+0.016$ \\
\textsc{Flash}      & $+0.341$ & $+0.028$ & $-0.012$ \\
\textsc{Pro}        & $+0.393$ & $+0.006$ & $-0.015$ \\
\bottomrule
\end{tabular}
\end{table}

\paragraph{Release.} The per-judgement annotations for all four raters, together with the specified weights and the analysis script that reproduces both tables, are released with the rest of the materials (\S\ref{sec:reproducibility-release}).

% ===== Inter-Rater Agreement =====

\section{Sensitivity Analyses}
\label{app:sensitivity}

The body's central claims rest on three analytical choices: the edge-presence threshold ($w > 0$) for the edge-recovery analysis (\S\ref{sec:res-structural}), the choice of \texttt{deepseek-v4-flash} as the scaled judge for the component-recovery analysis (\S\ref{sec:res-structural}), and the choice of MPNet (\texttt{all-mpnet-base-v2}) as the encoder for the cross-model convergence analysis (\S\ref{sec:res-geometric}). Each of these is defensible but not unique, so we run a sensitivity analysis on each: we vary the analytical choice and re-compute the headline statistic to verify that the direction of the formulation effect does not depend on the specific operationalisation.

\subsection{Edge-weight threshold}
\label{app:sens-threshold}

The body treats a directed edge as ``specified active'' whenever the persona's sampled edge weight $w_{A \to B}$ is strictly positive, and reports MCC $= +0.41$ for the external judge against this binary gold in the full condition. A reviewer can reasonably ask whether the recovery is being driven by edges with vanishingly small weights that the Crafter was unlikely to instantiate explicitly. We re-compute the classification metrics treating the gold positive class as $w > t$ for $t \in \{0, 0.1, 0.2, 0.3, 0.5\}$. Higher thresholds reduce the positive class and ask a harder question (does the judge recover the \emph{prominently}-weighted edges?).

\begin{figure}[!t]
\centering
\includegraphics[width=\columnwidth]{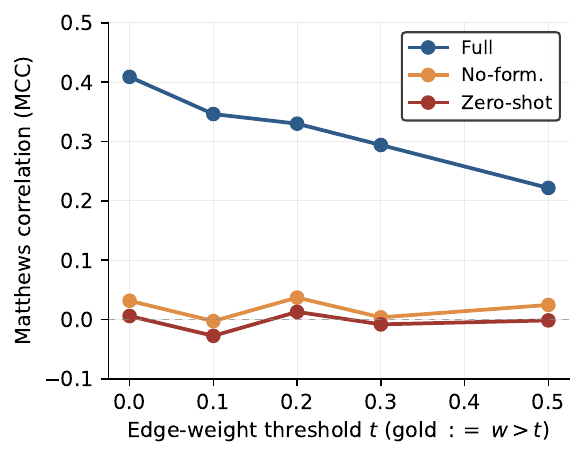}
\captionsetup{font=small}
\caption{Edge-recovery MCC across edge-weight thresholds $t \in \{0, 0.1, 0.2, 0.3, 0.5\}$, by condition. Full-condition MCC remains positive at every threshold (declining monotonically as the gold positive class shrinks); no-formulation and zero-shot remain at chance throughout.}
\label{fig:sens-edge-threshold-a}
\end{figure}

\begin{figure}[!t]
\centering
\includegraphics[width=\columnwidth]{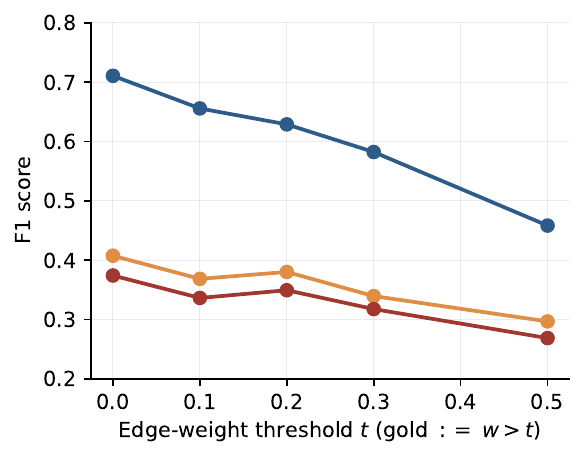}
\captionsetup{font=small}
\caption{Companion to Figure~\ref{fig:sens-edge-threshold-a}: edge-recovery F1 across the same thresholds. The full-vs-zero-shot gap is preserved at every operationalisation of edge presence.}
\label{fig:sens-edge-threshold-b}
\end{figure}

Table~\ref{tab:sens-threshold} reports the result, and Figures~\ref{fig:sens-edge-threshold-a} and~\ref{fig:sens-edge-threshold-b} plot recovery MCC and F1 across the thresholds. In the full condition, MCC declines monotonically from $+0.41$ at $t = 0$ to $+0.22$ at $t = 0.5$, but remains positive at every threshold. In the no-formulation and zero-shot conditions, MCC remains at chance ($-0.03$ to $+0.04$) across all thresholds. The full-vs-zero-shot MCC gap is $+0.41$ at the body's threshold and $+0.22$ at the strictest threshold tested, and the direction of the formulation effect is preserved at every choice. The attenuation at higher thresholds is informative rather than concerning: it shows that the formulation's edge-imprinting is more reliable for weakly-specified edges than for the strongest-weighted ones, consistent with the body's finding that the graph behaves as a near-binary on/off switch rather than a continuous prominence dial (\S\ref{sec:res-structural}, Table~\ref{tab:edge-recovery-by-cond}).

\begin{table}[h]
\centering
\small
\setlength{\tabcolsep}{4pt}
\caption{Edge-recovery classification metrics on the $330$-vignette evaluation subsample at multiple edge-weight thresholds for the positive gold class ($w > t$). \emph{prev} is the base rate of the positive gold class at that threshold. MCC stays positive in full at every threshold; the no-formulation and zero-shot rows remain at chance throughout.}
\label{tab:sens-threshold}
\begin{tabular}{l r rr rr}
\toprule
\textbf{Condition} & $\bm{t}$ & \textbf{prev} & \textbf{F1} & \textbf{MCC} & \textbf{AC1} \\
\midrule
Full              & $0.0$ & $0.49$ & $0.71$ & $\bm{+0.41}$ & $\bm{+0.41}$ \\
Full              & $0.1$ & $0.43$ & $0.66$ & $+0.35$ & $+0.33$ \\
Full              & $0.2$ & $0.40$ & $0.63$ & $+0.33$ & $+0.31$ \\
Full              & $0.3$ & $0.35$ & $0.58$ & $+0.29$ & $+0.26$ \\
Full              & $0.5$ & $0.25$ & $0.46$ & $+0.22$ & $+0.18$ \\
\midrule
No-formulation    & $0.0$ & $0.49$ & $0.41$ & $+0.03$ & $+0.07$ \\
No-formulation    & $0.5$ & $0.25$ & $0.30$ & $+0.02$ & $+0.32$ \\
\midrule
Zero-shot         & $0.0$ & $0.49$ & $0.37$ & $+0.01$ & $+0.06$ \\
Zero-shot         & $0.5$ & $0.25$ & $0.27$ & $-0.00$ & $+0.34$ \\
\bottomrule
\end{tabular}
\end{table}

\FloatBarrier
\subsection{Judge identity for the component-recovery analysis}
\label{app:sens-judge}

The body component-recovery analysis (\S\ref{sec:res-structural}, Appendix~\ref{app:judge-vs-specification}) uses \texttt{deepseek-v4-flash} as the scaled judge for the $15{,}000$-vignette generation set. To verify that the recovery is not Flash-specific, we compute the same statistic restricted to the $330$-vignette evaluation subsample (where Opus and DSpro are also available) and report the per-judge MCC against the specified \texttt{active\_nodes} gold for each of the five Ehlers~\&~Clark components in each condition (Table~\ref{tab:sens-judge}).

\begin{figure*}[h]
\centering
\includegraphics[width=0.95\textwidth]{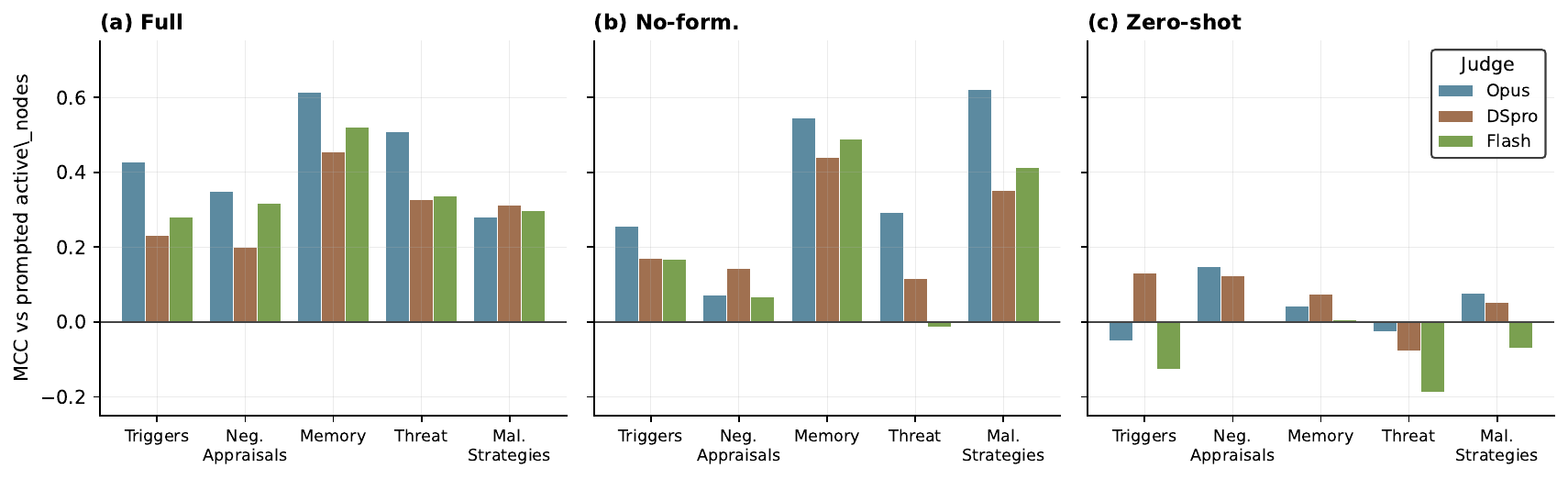}
\caption{Per-judge MCC for Ehlers~\&~Clark component recovery on the $330$-vignette evaluation subsample, by condition and component. (a) Full: every cell positive across all three judges and five components. (b) No-formulation: mostly positive, attenuated. (c) Zero-shot: at or near chance for every judge and component. The recovery direction holds for all three judges; Opus is the most sensitive and DSpro the most conservative, but the rank-order across components is consistent.}
\label{fig:sens-judge-component}
\end{figure*}

Two patterns emerge, both visible in Figure~\ref{fig:sens-judge-component}. First, every full-condition cell is positive across all three judges and all five components: the recovery direction is robust to judge identity. Mean MCC across components in the full condition is $+0.44$ for Opus, $+0.35$ for Flash, and $+0.30$ for DSpro; the absolute level varies (Opus is the most sensitive, DSpro the most conservative) but the rank-order across components is consistent (Memory and Threat highest, Strategies and Triggers lowest). Second, every zero-shot cell is at or near chance across all three judges (mean MCC $+0.04$ for Opus, $-0.08$ for Flash, $+0.06$ for DSpro), so the full-vs-zero-shot MCC gap is large for every judge ($+0.40$, $+0.43$, $+0.24$ respectively). The body's MCC $+0.41$ on the full $5{,}000$-vignette set is intermediate between Opus and Flash on the $330$-vignette subsample and consistent with the recovery being a property of the generated text rather than of the specific judge model.

\begin{table*}[h]
\centering
\small
\setlength{\tabcolsep}{4pt}
\caption{Per-judge MCC for EC-component recovery on the $330$-vignette evaluation subsample ($n = 110$ per condition). Gold is the persona's specified \texttt{active\_nodes}; predictions are each judge's binary EC ratings. Every full cell is positive across all three judges and all five components; every zero-shot cell is at or near chance.}
\label{tab:sens-judge}
\resizebox{\linewidth}{!}{%
\begin{tabular}{l l rrr rrr rrr}
\toprule
& & \multicolumn{3}{c}{\textbf{Full}} & \multicolumn{3}{c}{\textbf{No-formulation}} & \multicolumn{3}{c}{\textbf{Zero-shot}} \\
\cmidrule(lr){3-5} \cmidrule(lr){6-8} \cmidrule(lr){9-11}
\textbf{Component} & & Opus & DSpro & Flash & Opus & DSpro & Flash & Opus & DSpro & Flash \\
\midrule
Triggers           & MCC & $+0.43$ & $+0.23$ & $+0.28$ & $+0.26$ & $+0.17$ & $+0.17$ & $-0.05$ & $+0.13$ & $-0.13$ \\
Neg.\ Appraisals   & MCC & $+0.35$ & $+0.20$ & $+0.32$ & $+0.07$ & $+0.14$ & $+0.07$ & $+0.15$ & $+0.12$ & $+0.00$ \\
Memory             & MCC & $+0.62$ & $+0.46$ & $+0.52$ & $+0.54$ & $+0.44$ & $+0.49$ & $+0.04$ & $+0.08$ & $+0.01$ \\
Threat             & MCC & $+0.51$ & $+0.33$ & $+0.34$ & $+0.29$ & $+0.11$ & $-0.01$ & $-0.03$ & $-0.08$ & $-0.19$ \\
Strategies         & MCC & $+0.28$ & $+0.31$ & $+0.30$ & $+0.62$ & $+0.35$ & $+0.41$ & $+0.08$ & $+0.05$ & $-0.07$ \\
\midrule
\textbf{Mean}      & MCC & $\bm{+0.44}$ & $\bm{+0.30}$ & $\bm{+0.35}$ & $+0.36$ & $+0.24$ & $+0.23$ & $+0.04$ & $+0.06$ & $-0.08$ \\
\bottomrule
\end{tabular}}
\end{table*}

\FloatBarrier
\subsection{Embedding model for the cross-model convergence analysis}
\label{app:sens-embedding}

\begin{figure*}[!h]
\centering
\includegraphics[width=\textwidth]{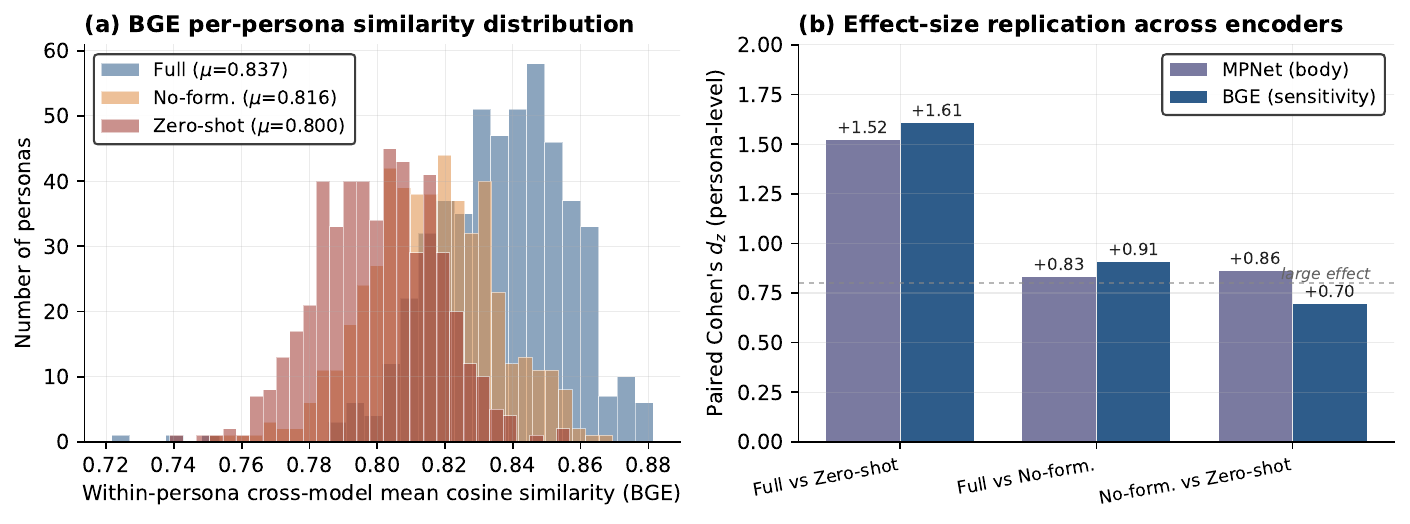}
\caption{Encoder sensitivity for the cross-model convergence claim. (a) Distribution of within-persona cross-model mean cosine similarity in BGE space, by condition ($n = 500$ personas per condition); the full $>$ no-formulation $>$ zero-shot ordering is preserved and the distributions are well-separated. (b) Paired Cohen's $d_z$ for the three condition contrasts under MPNet (body) versus BGE (sensitivity). All six bars are above the conventional ``large effect'' threshold of $0.8$; the full-vs-zero-shot contrast is larger under BGE ($+1.61$) than under MPNet ($+1.52$).}
\label{fig:sens-embedding}
\end{figure*}

The body's cross-model convergence analysis (\S\ref{sec:res-geometric}, Appendix~\ref{app:fig-consistency}) treats each vignette as a point in MPNet (\texttt{all-mpnet-base-v2}) sentence-embedding space, $768$-dimensional. The reported within-persona cross-model cosine-similarity statistic ($\mu_\text{full} = 0.792$, $\mu_\text{no-form} = 0.767$, $\mu_\text{zero-shot} = 0.737$; full-vs-zero-shot $d_z = +1.52$) could in principle be a property of MPNet's training rather than of the text: a different encoder might compress the persona-specific content differently and dissolve the cross-model convergence pattern. We test this by re-embedding the same $15{,}000$ vignettes (10 retained models) with BGE-base-en-v1.5 \citep{xiao2024c}, a BERT-backbone sentence encoder trained on a different curated mix of contrastive pairs with a different objective formulation, and recomputing the same statistic.

Table~\ref{tab:sens-embedding} reports the result. The conditional pattern replicates on every quantity: the per-condition means are higher in BGE space (the BGE encoder places all vignettes closer in cosine space in our data) but the \emph{ordering} is identical (full $>$ no-formulation $>$ zero-shot), the monotonic gradient is preserved, and the standardised effect sizes for the paired contrasts are comparable in direction and magnitude to those reported in the body, with one redistribution. The full-vs-zero-shot Cohen's $d_z = +1.61$ under BGE is slightly higher than the $+1.52$ reported under MPNet; the full-vs-no-formulation contrast is $+0.91$ under BGE (vs $+0.83$ MPNet); the no-formulation-vs-zero-shot contrast is $+0.70$ under BGE (vs $+0.86$ MPNet), so BGE redistributes the gap toward the full-vs-no-formulation step. All three pairwise contrasts are significant at $p$ below machine precision ($n = 500$ personas). The convergence claim is therefore not an artefact of MPNet's training: a structurally distinct encoder recovers the same conditional gradient.

\begin{table}[h]
\centering
\small
\setlength{\tabcolsep}{4pt}
\caption{Within-persona cross-model cosine similarity under two encoders. The body's MPNet baseline is reproduced for comparison; the BGE row is computed by re-embedding the same $15{,}000$ vignettes (10 retained models) with \texttt{BAAI/bge-base-en-v1.5} and applying the identical pipeline. $\mu$: mean per-persona similarity. $d_z$: paired Cohen's $d$ (persona as the pairing unit, $n = 500$). The conditional ordering and effect sizes are preserved across encoders.}
\label{tab:sens-embedding}
\resizebox{\linewidth}{!}{%
\begin{tabular}{l rrr rrr}
\toprule
& \multicolumn{3}{c}{\textbf{Per-condition $\mu$}} & \multicolumn{3}{c}{\textbf{Paired $d_z$ (vs zero-shot or vs no-form.)}} \\
\cmidrule(lr){2-4} \cmidrule(lr){5-7}
\textbf{Encoder} & Full & No-form. & Zero-shot & Full--Zs & No-form.--Zs & Full--No-form. \\
\midrule
MPNet  & $0.792$ & $0.767$ & $0.737$ & $+1.52$ & $+0.86$ & $+0.83$ \\
BGE    & $0.837$ & $0.816$ & $0.800$ & $\bm{+1.61}$ & $+0.70$ & $+0.91$ \\
\bottomrule
\end{tabular}
}
\end{table}

The higher absolute similarity values under BGE reflect a property of how the encoder distributes vignettes in cosine space and not a stronger formulation effect; the interpretable quantities are therefore the $d_z$ effect sizes, not the raw $\mu$ values. Identical inputs and identical similarity arithmetic make the two passes directly comparable.

\subsection{Vignette length as a potential confound}
\label{app:sens-length}

\begin{figure*}[tbp]
\centering
\includegraphics[width=0.95\textwidth]{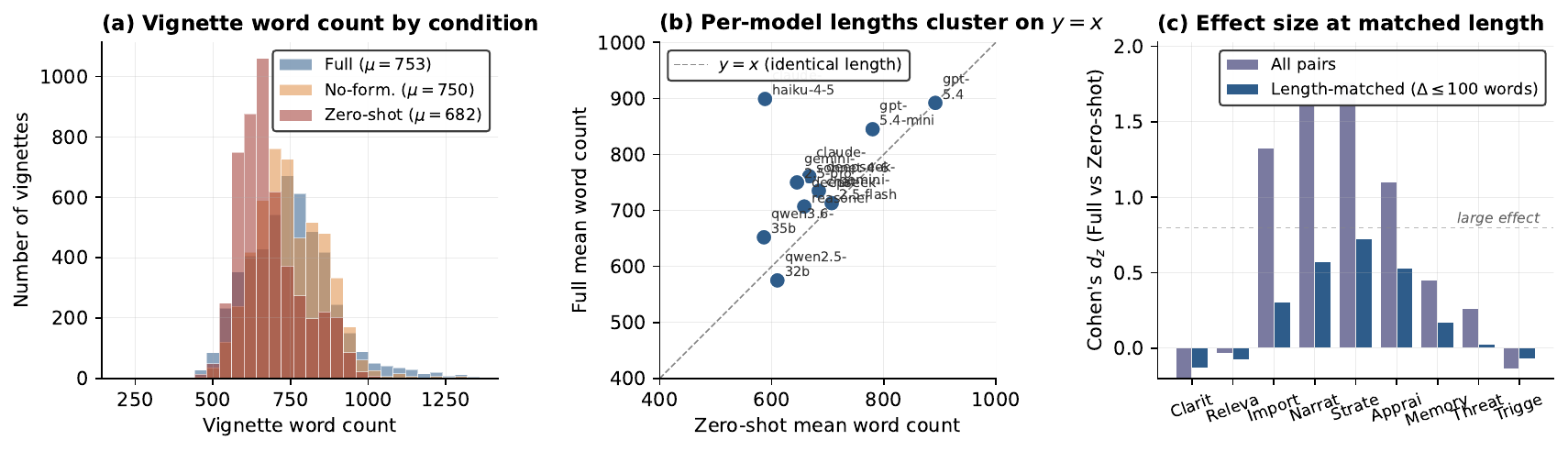}
\caption{Length is not a confound for the formulation effect. (a) Per-condition distribution of vignette word counts on the $15{,}000$-vignette generation set (10 retained models): full $\mu=753$, no-formulation $\mu=750$, zero-shot $\mu=682$, all overlapping. (b) Per-model means scatter close to the $y=x$ identity line: most models produce full and zero-shot vignettes within $\pm15\%$ of identical length (\texttt{gpt-5.4} $1.00\times$, \texttt{gemini-2.5-flash} $1.01\times$, \texttt{qwen2.5-32b} $0.94\times$). (c) Cohen's $d_z$ for full-vs-zero-shot per item: ``All pairs'' uses persona-level means across the full sample; ``Length-matched'' restricts to within-persona full/zero-shot pairs with $\leq 100$ word difference (median 16 words). The structurally-informative EC items (Strategies, Appraisals) retain large effect sizes at matched length; \emph{Threat} collapses under length-matching.}
\label{fig:length-robustness}
\end{figure*}

Vignette length is not a confound for the formulation effect. The generated vignettes are of comparable word count across conditions, and the formulation's structurally-defining effects survive at matched length with large $d_z$. We document the evidence below.

\paragraph{Vignette length vs API tokens.} Table~\ref{tab:tokens} reports \emph{prompt $+$ completion} tokens for the API call, not vignette length. The full-condition prompt includes the entire cognitive-graph specification ($\sim 10{,}000$ extra tokens of structural input), which contributes to the prompt-inclusive count but not to the generated text. The quantity relevant to length effects on rated quality is the vignette's word count, which we report here.

\paragraph{Vignette lengths are comparable across conditions.} Across the $10$ retained generation models (Figure~\ref{fig:length-robustness}a, b), mean vignette word count is $753$ (full), $750$ (no-formulation), and $682$ (zero-shot): a $\sim 10\%$ grand-mean difference, not the $\sim 5\times$ ratio inferable from prompt-inclusive tokens. Per-model full/zero-shot ratios cluster on the $y = x$ identity line: \texttt{gpt-5.4} $1.00\times$, \texttt{gemini-2.5-flash} $1.01\times$, \texttt{qwen2.5-32b} $0.94\times$, \texttt{deepseek-chat} $1.07\times$, \texttt{claude-sonnet-4-6} $1.14\times$. The maximum is \texttt{claude-haiku-4-5} at $1.53\times$, a single-model outlier; two of the ten models produce \emph{shorter} full vignettes than zero-shot. The condition does not consistently move length in either direction at the per-model level.

\paragraph{Effect sizes at matched length.} Within each of the $500$ personas, we identify the (full, zero-shot) vignette pair with the smallest word-count difference: median $16$ words across $4{,}935$ matched pairs. At this near-identical length, the formulation's effects on the structurally-informative items remain large: Maladaptive Strategies $d_z = +0.72$, Negative Appraisals $+0.53$, Narrative quality (g2) $+0.57$ (Figure~\ref{fig:length-robustness}c). Within the three CVI items, the pattern splits: Importance retains a positive contrast at matched length ($d_z = +0.31$), while Clarity ($-0.13$) and Relevance ($-0.08$) remain in the same direction as their all-pairs estimates ($-0.26$ and $-0.03$ respectively) -- the LLM-judge preference for unconstrained zero-shot prose on these two items is not a length artefact, and the experts' counter-direction on Clarity ($d_z = +0.81$) is therefore also not length-driven. Triggers is non-significant in both estimates (saturated at $\sim 0.93$). Of the items tested, only Sense of current Threat goes from sizable to negligible at matched length ($d_z$ falls from $+0.27$ to $+0.03$).

\paragraph{OLS with length covariate.} Regressing each item on \texttt{condition $+$ $\log(\text{word\_count})$} on the $15{,}000$-vignette set (Table~\ref{tab:length-covariate}), the formulation coefficient on Strategies and Appraisals retains $92.4\%$ and $93.7\%$ of its magnitude with length controlled: the effect is conditional on the cognitive-graph specification, not on word count. Within CVI, Importance retains $73\%$, Clarity and Relevance are below detection in both specifications ($|b_{\text{full}}| < 0.03$). Narrative retains $83\%$; Memory $68\%$.

\paragraph{Mediation, not confounding.} For Threat the formulation coefficient does drop substantially when length is in the model ($65\%$ absorbed). This is mediation through the formulation, not confounding by an external variable. When the cognitive graph specifies that sense of current threat must be portrayed, the model writes a richer threat description, which is both longer and more clinically vivid; the two move together because the formulation drives both. Truncating the vignette to remove the length would also remove the threat content, since the content \emph{is} the additional words. The same logic applies to the partial mediation on Importance and Narrative.

\begin{table}[h]
\centering
\small
\setlength{\tabcolsep}{4pt}
\caption{OLS coefficient for the ``full'' indicator on each item, with and without $\log(\text{word\_count})$ as a covariate. Zero-shot is the reference. \textbf{\% retained}: fraction of the formulation effect that survives controlling for length, i.e.\ the part that is \emph{not} explained by word count. The structurally-informative EC items (Strategies, Appraisals) retain $> 90\%$; ordinal items $73$--$83\%$; only Threat is largely length-mediated.}
\label{tab:length-covariate}
\begin{tabular}{l rrr}
\toprule
\textbf{Item} & \textbf{No length} & \textbf{With $\log w$} & \textbf{\% retained} \\
\midrule
ec\_strategies  & $+0.429$ & $+0.396$ & $\bm{92.4\%}$ \\
ec\_appraisals  & $+0.332$ & $+0.311$ & $\bm{93.7\%}$ \\
narrative (g2)  & $+0.411$ & $+0.340$ & $82.9\%$ \\
importance      & $+0.281$ & $+0.206$ & $73.2\%$ \\
ec\_memory      & $+0.176$ & $+0.119$ & $67.6\%$ \\
ec\_threat      & $+0.073$ & $+0.025$ & $34.9\%$ \\
clarity         & $-0.015$ & $-0.024$ & --- \\
relevance       & $-0.002$ & $-0.023$ & --- \\
\bottomrule
\end{tabular}
\end{table}

\paragraph{Bottom line.} Vignettes are of comparable word count across conditions ($\sim 10\%$ grand-mean difference; per-model ratios on the identity line; two of ten models shorter in full than in zero-shot). The structurally-defining EC effects on Strategies and Appraisals survive length-matched pairing with large $d_z$ and retain $> 90\%$ of their magnitude under length control. Within CVI, Importance retains $73\%$, and the LLM-judge negative effects on Clarity and Relevance are unchanged at matched length -- so the judge-stylistic finding flagged in \S\ref{sec:results} is not length-driven either. Narrative retains $83\%$; Memory $68\%$. Only Sense of current Threat is largely length-mediated, and this is mediation through the formulation (writing about threat is writing more about threat), not confounding by an external variable. The structural findings the paper rests on, edge-recovery MCC and EC presence on Strategies and Appraisals, are not reducible to the generator writing more tokens.

\FloatBarrier
\subsection{Cluster geometry: cognitive formulation vs trauma type}
\label{app:sens-clustering}

The body claim that the formulation breaks trauma-type stereotyping (\S\ref{sec:res-geometric}, Figure~\ref{fig:trauma-stereotyping}) is established qualitatively from a $t$-SNE visualisation. We complement it here with a quantitative cluster-cohesion measure on the BGE embeddings: for each condition we measure how tightly vignettes cluster by (i) trauma type ($n_g = 29$ categories) and by (ii) the persona's specified active-node signature ($n_g = 32$ unique combinations across the $500$ personas). Each measurement is the difference between mean within-group cosine similarity and mean between-group cosine similarity (within-between gap); a larger gap means the labelling explains more of the embedding's structure. We bootstrap the gap from $B = 20$ subsamples of $n = 2000$ vignettes per condition to obtain standard deviations.

\begin{figure}[h]
\centering
\includegraphics[width=\columnwidth]{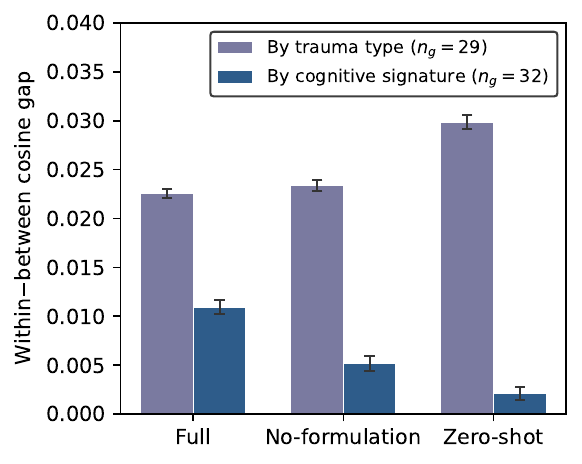}
\captionsetup{font=small}
\caption{Cluster cohesion of vignette embeddings by label type, per condition (BGE-base, 10 retained models). Within-between cosine-similarity gap with trauma type (light) vs cognitive-signature (dark) as the cluster label. Trauma-type clustering is highest in zero-shot (template fallback); cognitive-signature clustering is highest in full (the graph imprints on the embedding).}
\label{fig:sens-clustering-a}
\end{figure}

\begin{figure}[h]
\centering
\includegraphics[width=\columnwidth]{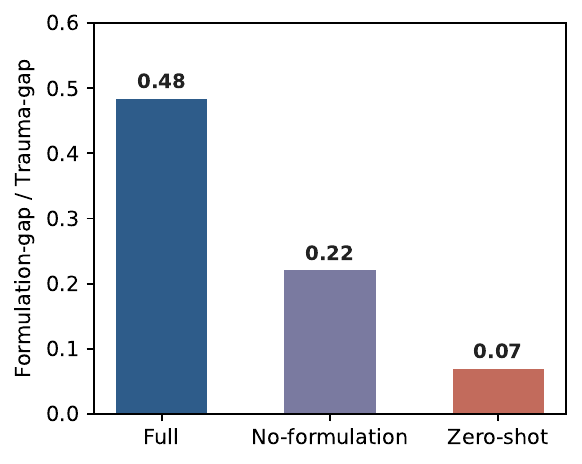}
\captionsetup{font=small}
\caption{Companion to Figure~\ref{fig:sens-clustering-a}: composite cluster-cohesion ratio (formulation gap $/$ trauma gap) per condition. The ratio rises from $0.07$ in zero-shot to $0.48$ in full, a $\sim 7\times$ shift in what the embedding picks up on as the dominant grouping signal.}
\label{fig:sens-clustering-b}
\end{figure}

Table~\ref{tab:sens-cluster} reports the result, and Figures~\ref{fig:sens-clustering-a} and~\ref{fig:sens-clustering-b} show the per-condition cohesion and the composite ratio. The pattern is what the body claim predicts but had not previously demonstrated quantitatively: zero-shot vignettes cluster $\sim 30\%$ more tightly by trauma type than full-condition vignettes do (gap $0.030$ vs $0.023$), while full-condition vignettes cluster $\sim 5\times$ more tightly by cognitive formulation signature than zero-shot vignettes do ($0.011$ vs $0.002$, the latter indistinguishable from zero given the bootstrap standard deviation). The composite ratio (formulation-gap divided by trauma-gap) captures the structural shift in a single number: in zero-shot the embedding picks up on trauma-type templates roughly fourteen times more than on cognitive structure ($0.07$); in full the embedding gives the two roughly equal weight ($0.48$); no-formulation is intermediate ($0.22$). The cognitive-graph specification actively redirects the embedding from coarse trauma-category clustering toward person-specific cognitive-signature clustering, which is exactly the structural shift the framework was designed to produce.

\begin{table}[h]
\centering
\small
\setlength{\tabcolsep}{4pt}
\caption{Cluster cohesion of vignette embeddings (BGE-base) by trauma type and by cognitive formulation signature, per condition. Each entry is mean within-group $-$ between-group cosine similarity, bootstrapped on $B = 20$ subsamples of $n = 2000$ vignettes; SDs in parentheses. \textbf{Ratio}: formulation gap divided by trauma gap. Higher ratio = embedding is shaped by cognitive structure rather than trauma category. The shift from zero-shot to full is a factor of $\approx 7\times$ on this ratio.}
\label{tab:sens-cluster}
\resizebox{\linewidth}{!}{%
\begin{tabular}{l rrr}
\toprule
\textbf{Condition} & \textbf{Trauma gap} & \textbf{Formulation gap} & \textbf{Ratio} \\
\midrule
Full              & $0.023\,(0.000)$ & $\bm{0.011\,(0.001)}$ & $\bm{0.48}$ \\
No-formulation    & $0.023\,(0.001)$ & $0.005\,(0.001)$ & $0.22$ \\
Zero-shot         & $\bm{0.030\,(0.001)}$ & $0.002\,(0.001)$ & $0.07$ \\
\bottomrule
\end{tabular}
}
\end{table}

\begin{figure*}[tbp]
\centering
\includegraphics[width=\textwidth]{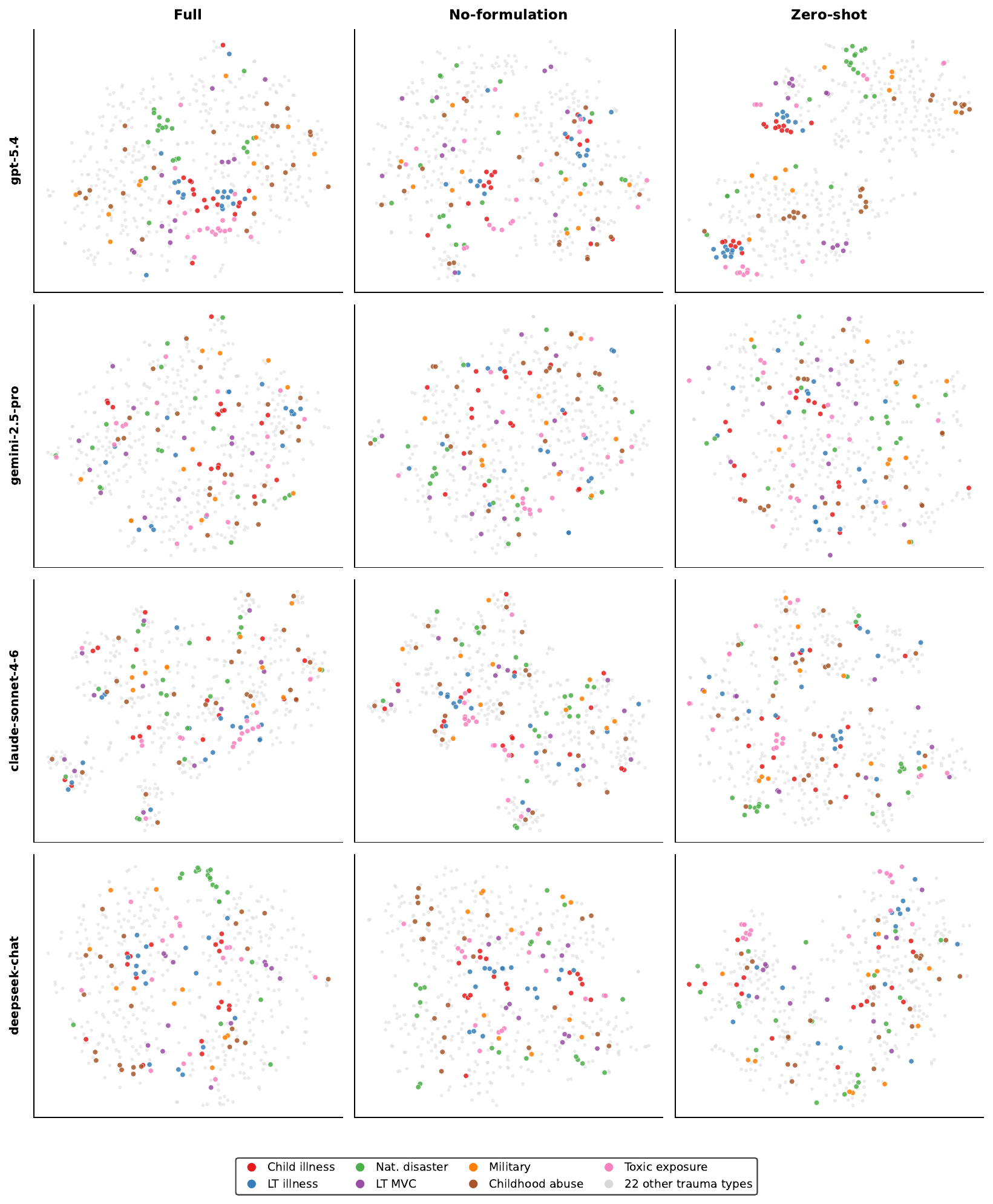}
\caption{BGE-base $t$-SNE replication of the body's trauma-stereotyping figure (Figure~\ref{fig:trauma-stereotyping}), on the same four representative generation models. Per-cell $t$-SNE on $500$ vignettes; the seven trauma types with the tightest zero-shot intra-trauma similarity are coloured, the other $22$ in light grey. Tight monochromatic clusters are most striking in the \texttt{gpt-5.4} row (top right); the same coloured vignettes are dispersed under \textit{Full} and \textit{No-formulation}. The qualitative pattern of the body figure replicates under a structurally distinct encoder.}
\label{fig:trauma-stereotyping-bge}
\end{figure*}

The quantitative cluster-cohesion result is supported visually by per-cell $t$-SNE projections of the BGE embeddings, structured to reveal two distinct angles on the conditional pattern. Figure~\ref{fig:trauma-stereotyping-bge} reproduces the body's trauma-stereotyping figure (Figure~\ref{fig:trauma-stereotyping}) on the same four representative generation models using the alternative encoder; the conditional pattern replicates immediately (zero-shot vignettes form tight monochromatic clusters that disperse under full), confirming that the body's qualitative finding is not specific to MPNet's training.

\begin{figure*}[tbp]
\centering
\includegraphics[width=\textwidth]{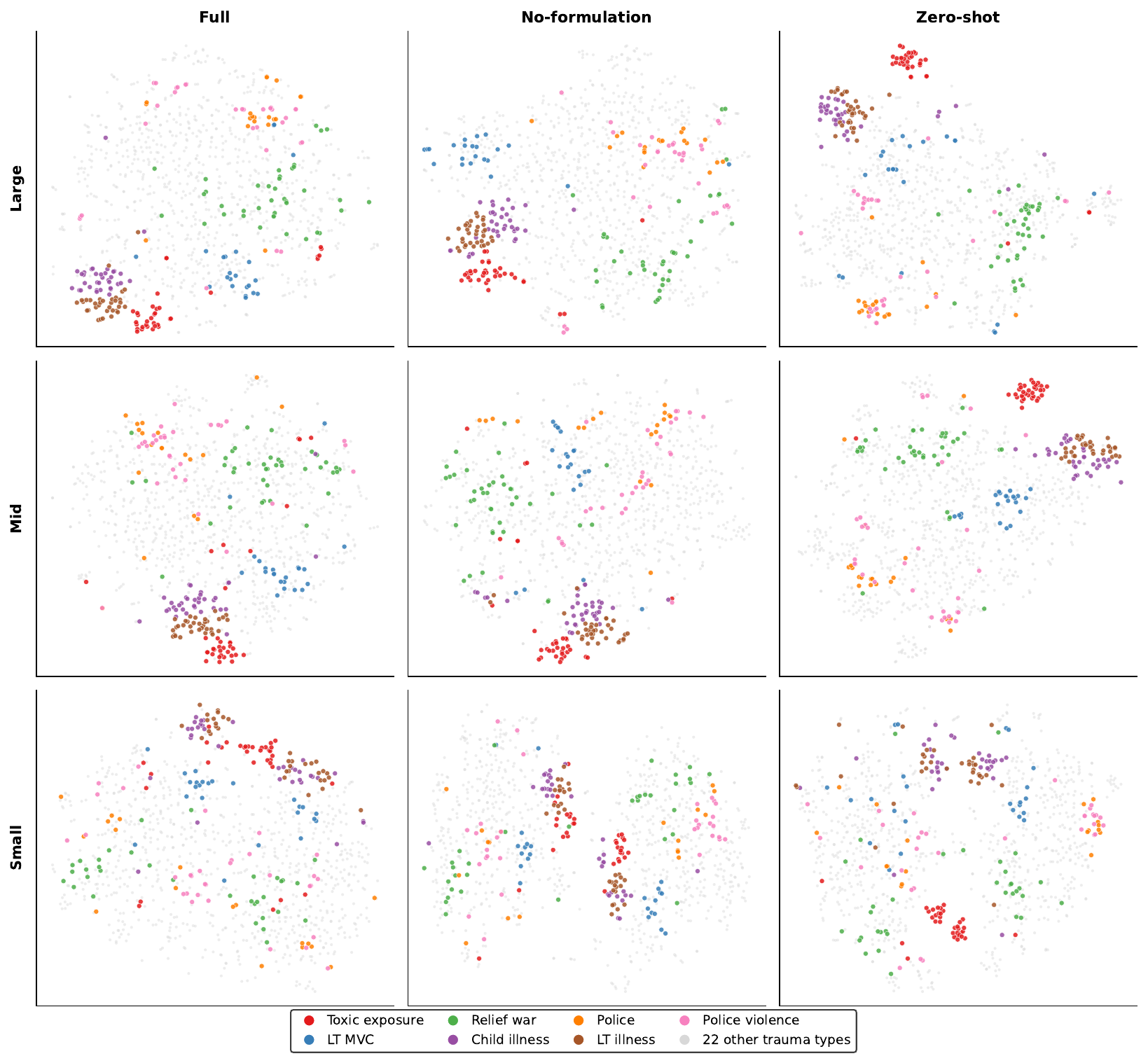}
\caption{MPNet $t$-SNE grid, regrouped by model capability tier. Per-cell $t$-SNE on up to $1{,}000$ vignettes per (tier, condition). Two patterns: trauma-type clustering breaks across all tiers as the prompt becomes more structured (right $\to$ left), and the breaking is more complete in higher-capability tiers. This is the same body encoder under a tier-level grouping, showing the formulation $\times$ capacity interaction directly.}
\label{fig:trauma-stereotyping-tiers-mpnet}
\end{figure*}

\begin{figure*}[tbp]
\centering
\includegraphics[width=\textwidth]{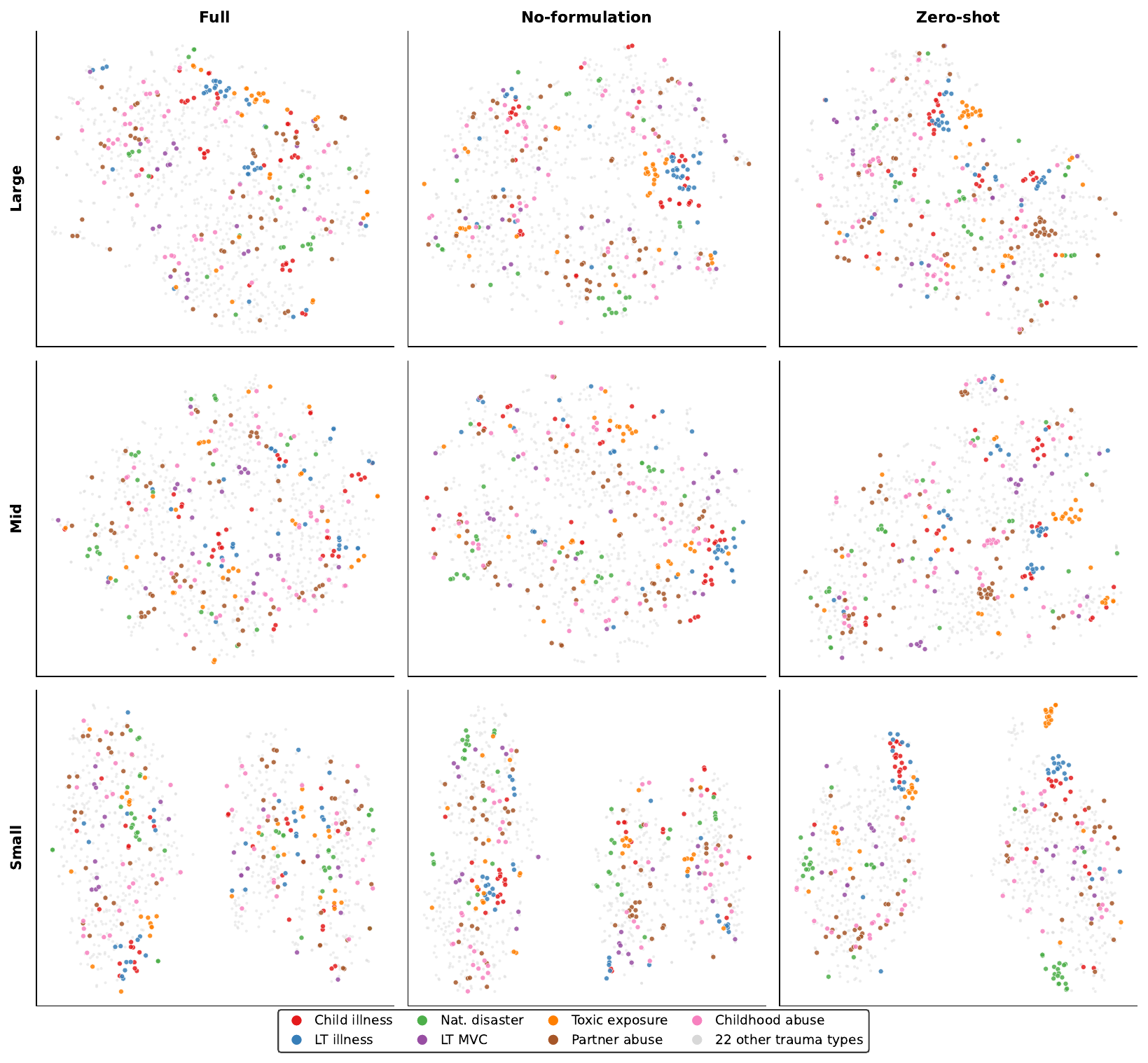}
\caption{BGE-base $t$-SNE grid, same tier grouping as Figure~\ref{fig:trauma-stereotyping-tiers-mpnet}. The tier $\times$ condition interaction reproduces under the alternative encoder. Absolute cluster cohesion is somewhat lower than under MPNet (BGE is less sensitive to trauma-category content than MPNet), but the conditional and tier gradients are identical, ruling out an encoder-specific artefact.}
\label{fig:trauma-stereotyping-tiers-bge}
\end{figure*}

Figures~\ref{fig:trauma-stereotyping-tiers-mpnet} and \ref{fig:trauma-stereotyping-tiers-bge} go further: they regroup the eleven generation models into three capability tiers (\textbf{Large}: \texttt{gpt-5.4}, \texttt{claude-sonnet-4-6}, \texttt{deepseek-reasoner}, \texttt{gemini-2.5-pro}; \textbf{Mid}: \texttt{gpt-5.4-mini}, \texttt{claude-haiku-4-5}, \texttt{deepseek-chat}, \texttt{gemini-2.5-flash}; \textbf{Small}: \texttt{gpt-4o-mini}, \texttt{qwen2.5-32b}, \texttt{qwen3.6-35b}) and present the same $t$-SNE projection under both encoders. The tier breakdown surfaces an interaction the body figure could not show: the breaking of the trauma-type templates is most dramatic in the large-capability tier, where full-condition clusters are nearly invisible; the mid tier shows partial breaking; and the small tier retains visible monochromatic clusters even under full. The pattern is identical under both encoders, ruling out an encoder artefact. The interpretation is that the formulation specification and the generator's underlying capacity to act on it interact: a model has to be capable enough to instantiate the cognitive graph in narrative form, otherwise it falls back on trauma-type templates regardless of what the input asks.

\FloatBarrier

\FloatBarrier

\section{Trauma-Type Stereotyping in Latent Space}
\label{app:fig-trauma-stereotyping}
\begin{figure*}[tbp]
    \centering
    \includegraphics[width=0.95\textwidth]{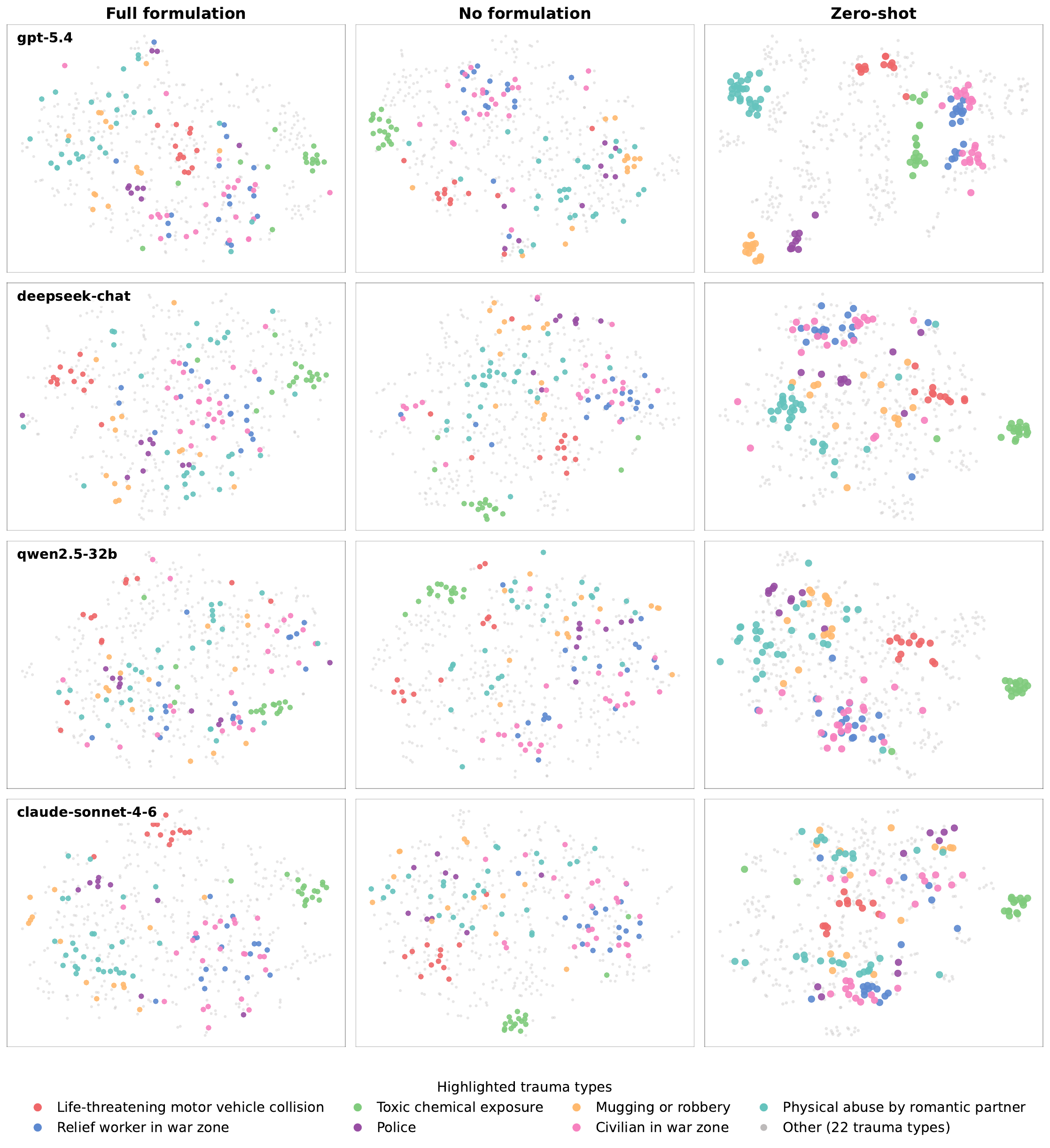} 
    \caption{Vignette semantic space, four generation models $\times$ three conditions, projected with per-cell $t$-SNE on MPNet embeddings. 
    Each dot is one vignette. Colour identifies one of the seven trauma types with the tightest zero-shot clustering (intra-trauma cosine similarity, averaged across the four models). 
    The other 22 trauma types are shown in light grey. Tight monochromatic clusters in the \textit{Zero-shot} column reveal trauma-type stereotyping of the models. 
    The same coloured vignettes are dispersed under \textit{Full} and \textit{No formulation}, showing that the formulation breaks the fallback to trauma-type templates and re-anchors the vignette on the persona's individual content.}
    \label{fig:trauma-stereotyping}
\end{figure*}

Trauma-type stereotyping is a specific failure mode of LLM-generated clinical text: when a model has nothing to anchor on but a diagnostic label and a trauma category, it converges on a small set of category templates (``the combat-veteran vignette,'' ``the sexual-assault-survivor vignette''), and individual persona content is lost in the average. The body claim (\S\ref{sec:res-geometric}) is that the cognitive formulation breaks this fallback. Figure~\ref{fig:trauma-stereotyping} visualises the underlying evidence.

The procedure: (i) embed all $500$ personas $\times$ $4$ representative models $\times$ $3$ conditions $= 6{,}000$ vignettes with MPNet; (ii) within each (model, condition) cell, compute intra-trauma-type mean cosine similarity for every one of the $29$ trauma categories; (iii) within each cell, project the $500$ vignettes to two dimensions using $t$-SNE \citep{vandermaaten2008visualizing} (perplexity $30$, default scikit-learn parameters; \citealt{pedregosa2011scikit}); (iv) colour the seven trauma types with the highest zero-shot intra-trauma similarity (averaged across the four models). The remaining $22$ trauma types are plotted in light grey to provide a backdrop without dominating the figure visually.

In every zero-shot panel the coloured points form tight monochromatic clusters: a model writing zero-shot for a combat-veteran persona produces text that sits very close to its other combat-veteran vignettes, regardless of any other persona feature. In the no-formulation and full panels the same coloured points are dispersed throughout the layout, indicating that with the persona-specific content in the prompt the model writes from the individual rather than from the category. The four models shown are chosen to span the AI-detection range observed in the user study (\texttt{gpt-5.4} and \texttt{gemini-2.5-pro} at the human-passing end, \texttt{deepseek-chat} and \texttt{qwen2.5-32b} at the more-detectable end); the pattern is qualitatively the same in all four, supporting the claim that stereotyping is a property of the input structure rather than of the generator.

We emphasise that this is a visualisation, not a hypothesis test. $t$-SNE is non-deterministic and its layout is not metric; the statistical claim that anchors the body assertion is the cross-model convergence test (Appendix~\ref{app:tab-consistency-ttest}). The figure is intended as a qualitative illustration of the same effect at the per-trauma-type level.

% ===== Component and Edge Recovery =====

\label{app:tab-consistency-ttest}
Table~\ref{tab:consistency-ttest} reports paired $t$-tests on the per-persona mean within-persona cross-model cosine similarity introduced above ($n = 500$ personas, $df = 499$). The pairing unit is the persona: the same $500$ personas yield three similarity scores (one per condition), and we test whether the mean shifts across conditions are non-zero. Pairing controls for persona-level idiosyncrasy in how easy a given individual is to write about consistently (a persona with rare triggers or an atypical occupation will be portrayed more variably across models regardless of condition), so the resulting contrasts isolate the condition effect.

All three pairwise contrasts are highly significant ($t$ between $16$ and $32$, $p<10^{-48}$); the full-vs-zero-shot contrast is the largest ($\Delta\mu = +0.055$, Cohen's $d_z = +1.52$), and the full-vs-no-formulation contrast is the smallest ($\Delta\mu = +0.025$, $d_z = +0.83$). The ordering $\Delta_\text{full,zs} > \Delta_\text{no-form,zs} > \Delta_\text{full,no-form}$ mirrors the body's general pattern: the cognitive graph adds more cross-model convergence than the self-report items alone, and the self-report items alone already add more than the demographic prompt.

Two methodological remarks. First, we report Cohen's $d_z$ rather than between-subject $d$ because the design is within-persona; the unstandardised mean differences are small (under $0.05$ on a $[-1,1]$ similarity scale) but the within-persona variance of the contrast is even smaller, which is what drives the large effect sizes. Second, the test treats $s_{p,c}$ as the response, not the raw $\binom{M}{2}=45$ pairwise similarities; aggregating to the persona avoids the model-pair correlation that would inflate the effective sample size.

\begin{table}[h]
\centering
\small
\caption{{Paired $t$-tests on per-persona mean within-persona cross-model cosine similarity ($n=500$ personas, $df=499$, all $p<10^{-48}$).}}
\label{tab:consistency-ttest}
\begin{tabular}{lrrr}
\toprule
\textbf{Comparison} & $\Delta\mu$ & $t$ & $d$ \\
\midrule
Full vs No-formulation      & $+0.0251$ & $18.63$ & $+0.83$ \\
Full vs Zero-shot     & $+0.0554$ & $34.03$ & $+1.52$ \\
No-formulation vs Zero-shot & $+0.0302$ & $19.28$ & $+0.86$ \\
\bottomrule
\end{tabular}
\end{table}

\label{app:fig-consistency}
The geometric analysis in \S\ref{sec:res-geometric} treats each vignette as a point in a $768$-dimensional sentence-embedding space (\texttt{all-mpnet-base-v2}; \citealt{reimers2019sentence}) and asks whether different generation models, given the same persona and condition, produce text that lies closer together than text from different personas. Concretely, for every persona $p$ and condition $c$ we compute the within-persona cross-model mean cosine similarity $s_{p,c} = \frac{1}{\binom{M}{2}}\sum_{i<j} \cos(v_{p,c,i}, v_{p,c,j})$, where $v_{p,c,m}$ is the embedding of model $m$'s vignette for persona $p$ in condition $c$ and $M=10$ is the model set (gpt-4o-mini excluded, Appendix~\ref{app:iterations}). The persona is held fixed and only the generator identity varies, so $s_{p,c}$ measures how stably the text reflects the persona's specified content as we swap the generator.

\begin{figure}[H]
    \centering
    \includegraphics[width=\columnwidth]{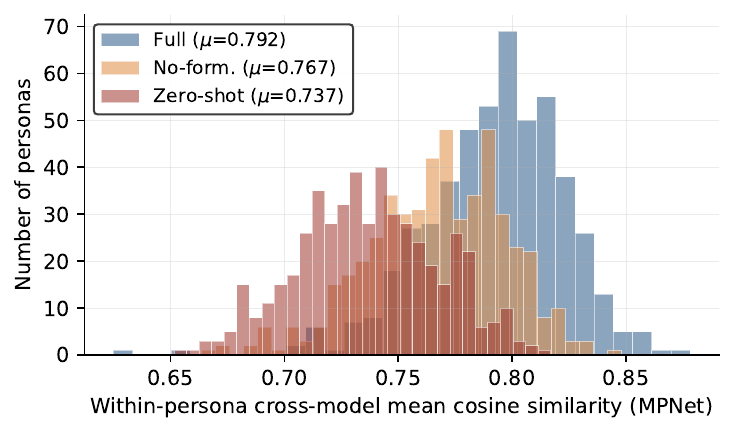}
    \caption{Within-persona cross-model cosine-similarity distributions of vignette embeddings, by condition; higher density at high similarity means the ten retained generation models converge on a more similar portrayal of the same persona.}
    \label{fig:consistency}
\end{figure}

Figure~\ref{fig:consistency} shows the distribution of $s_{p,c}$ across the $n = 500$ personas, separately for each condition. Three features of the distribution carry interpretive weight: its mean ($\mu_\text{full} = 0.792$, $\mu_\text{no-form} = 0.767$, $\mu_\text{zero-shot} = 0.737$), its spread (full is narrower than zero-shot, indicating not only higher mean similarity but also lower across-persona variance in that similarity), and the absence of an upper-tail truncation in zero-shot that would indicate text converging on a generic ``PTSD vignette'' template. The full-condition distribution is shifted right and tightened, exactly the pattern one would predict if the cognitive graph acted as a shared content scaffold across models.

This analysis does \emph{not} test whether two models produce textually identical sentences, only whether they portray the same persona. The MPNet encoder is trained on semantic similarity rather than surface form, so two paraphrases of ``Maria suppresses her thoughts about the attack and avoids walking alone at night'' will sit close in this space even if no word is shared.

\section{Full Pairwise Inter-Rater Agreement}
\label{app:full-pairwise}

\paragraph{Expert raters and training protocol.}
R1 and R2 are both MD general-practice residents who have received supervised training in reading and writing clinical cognitive case formulations as part of their residency curriculum. Before beginning the formal annotation, they completed a calibration exercise: each rated six training vignettes drawn from the persona pool but \emph{not} included in the $330$-vignette evaluation subsample, then discussed every disagreement among themselves and with the authors. The authors reviewed both raters' training outputs and verified that the operational rating standard (in particular, what counts as an Ehlers~\&~Clark component being ``present'' as a described feature of the patient's presentation, and what counts as a directed causal edge being conveyed within a paragraph) was internalised consistently before any subsample vignettes were rated. Each rater then completed the $330$-vignette annotation independently, blind to model and condition, over approximately four days of dedicated work. Disagreements at scoring time were not re-discussed: the analysis we report uses the raters' independent ratings, not a post-hoc consensus.

\paragraph{Full pairwise table.}
Table~\ref{tab:full-pairwise} reports the full $10$-pair Gwet AC matrix on the $330$-vignette evaluation subsample. The body table (Table~\ref{tab:judge-consensus}) summarises this matrix into three cluster columns (R1--R2, LLM--LLM mean over the three LLM pairs, and the multi-rater All 5); the six human--LLM cross pairs (R$_i$--$j$) and the H--LLM cluster average derivable from them are reported here. The full pairwise matrix is what the cluster summaries are computed from, and is the right table to inspect for whether any individual rater pair is driving the cluster patterns.

We use Gwet's AC1 for binary items and AC2 with quadratic weights for ordinal items, in preference to Cohen's $\kappa$/$\kappa_w$ and Krippendorff's $\alpha$, because AC1/AC2 do not exhibit the prevalence-and-bias paradox that affects $\kappa$ on items with marginal imbalance \citep{gwet2014handbook}. Several of our items (DSM Trauma exposure, DSM Impairment) saturate near $1.0$ marginally; on these items $\kappa$ can be near zero or negative even when raters agree on every observation, while AC1 is well-behaved. AC2 with quadratic weights extends the same machinery to ordinal items (treating two-level disagreement as four times as severe as one-level disagreement), and is the standard for ordinal clinical ratings.

Several patterns are visible in the full matrix that the cluster summary hides. (i) On the EC items, the two human experts (R1--R2) agree highly with each other ($0.83$--$0.95$) but each agrees with LLMs $\sim 0.30$--$0.40$ lower; the three LLMs agree with each other near-ceiling ($\geq 0.90$ on Memory, Threat, Triggers). This is the standard signature of two distinct rating populations: humans and LLMs apply systematically different thresholds for EC component presence, and the H--LLM cluster summary in the body conceals this by averaging. (ii) On the ordinal items the human--LLM gap is much smaller and the LLM--LLM agreement is uniformly high, suggesting the rating instructions are operationalisable consistently across rater types for clarity, relevance, importance, etc. (iii) Within the LLM group, Opus is the closest to the human experts on EC items (R1--Opus $0.41$ on Threat vs R1--Flash $0.26$); we report this in case downstream users want to choose a single judge that best approximates a clinical rater.

The multi-rater All-5 column is computed with Gwet's multi-rater AC \citep{gwet2014handbook}, which generalises the pairwise statistic by treating the agreement among multiple raters on the same item as the parameter of interest rather than averaging pairwise statistics. We report it to give a single agreement number per (metric, condition) cell for use as a rater-pool reliability estimate.

\begin{table*}[h]
\centering
\scriptsize
\setlength{\tabcolsep}{3pt}
\caption{Complete pairwise Gwet AC on the 330-vignette evaluation subsample, pooled across all three conditions ($n = 330$). AC1 for binary, AC2 with quadratic weights for ordinal. \emph{All 5} is the multi-rater AC across R1, R2, O, P, F. Per-condition All-5 values are visualised in Figure~\ref{fig:multirater-by-condition}.}
\label{tab:full-pairwise}
\resizebox{\linewidth}{!}{%
\begin{tabular}{lrrrrrrrrrrr}
\toprule
\textbf{Metric} & \textbf{R1--R2} & \textbf{R1--O} & \textbf{R1--P} & \textbf{R1--F} & \textbf{R2--O} & \textbf{R2--P} & \textbf{R2--F} & \textbf{O--P} & \textbf{O--F} & \textbf{P--F} & \textbf{All 5} \\
\midrule
\multicolumn{12}{l}{\textit{Ordinal (AC2, quadratic weights)}} \\
Clarity     & 0.951 & 0.754 & 0.779 & 0.789 & 0.687 & 0.714 & 0.724 & 0.971 & 0.967 & 0.983 & 0.840 \\
Relevance   & 0.951 & 0.822 & 0.837 & 0.831 & 0.788 & 0.798 & 0.790 & 0.980 & 0.977 & 0.978 & 0.881 \\
Importance  & 0.912 & 0.810 & 0.828 & 0.825 & 0.860 & 0.870 & 0.853 & 0.958 & 0.914 & 0.912 & 0.875 \\
Grounded    & 0.957 & 0.693 & 0.739 & 0.698 & 0.636 & 0.684 & 0.637 & 0.950 & 0.964 & 0.958 & 0.797 \\
Narrative   & 0.963 & 0.816 & 0.817 & 0.779 & 0.792 & 0.793 & 0.754 & 0.952 & 0.923 & 0.908 & 0.851 \\
Explicit    & 0.971 & 0.834 & 0.832 & 0.834 & 0.799 & 0.797 & 0.798 & 0.999 & 0.999 & 0.999 & 0.894 \\
Focused     & 0.953 & 0.742 & 0.754 & 0.768 & 0.674 & 0.688 & 0.707 & 0.905 & 0.928 & 0.911 & 0.805 \\
\midrule
\multicolumn{12}{l}{\textit{DSM-5 binary (AC1)}} \\
Trauma exposure     & 0.958 & 0.925 & 0.807 & 0.843 & 0.936 & 0.820 & 0.855 & 0.869 & 0.896 & 0.867 & 0.879 \\
Intrusion           & 0.974 & 0.852 & 0.851 & 0.844 & 0.852 & 0.851 & 0.844 & 0.949 & 0.963 & 0.941 & 0.891 \\
Avoidance           & 0.950 & 0.820 & 0.799 & 0.878 & 0.797 & 0.768 & 0.851 & 0.888 & 0.909 & 0.904 & 0.857 \\
Negative cognitions & 0.985 & 0.889 & 0.925 & 0.929 & 0.900 & 0.929 & 0.939 & 0.923 & 0.913 & 0.936 & 0.927 \\
Hyperarousal        & 0.988 & 0.946 & 0.959 & 0.959 & 0.932 & 0.946 & 0.946 & 0.967 & 0.974 & 0.974 & 0.959 \\
Impairment          & 1.000 & 0.988 & 0.994 & 0.991 & 0.988 & 0.994 & 0.991 & 0.982 & 0.991 & 0.985 & 0.990 \\
\midrule
\multicolumn{12}{l}{\textit{Ehlers \& Clark binary (AC1)}} \\
Threat        & 0.566 & 0.414 & 0.228 & 0.260 & 0.184 & 0.147 & 0.140 & 0.603 & 0.668 & 0.602 & 0.381 \\
Appraisals    & 0.717 & 0.409 & 0.350 & 0.368 & 0.321 & 0.296 & 0.324 & 0.729 & 0.766 & 0.718 & 0.502 \\
Memory        & 0.856 & 0.368 & 0.310 & 0.336 & 0.300 & 0.219 & 0.257 & 0.600 & 0.682 & 0.619 & 0.453 \\
Strategies    & 0.621 & 0.370 & 0.450 & 0.391 & 0.375 & 0.534 & 0.374 & 0.676 & 0.621 & 0.607 & 0.503 \\
Triggers      & 0.829 & 0.689 & 0.659 & 0.670 & 0.637 & 0.613 & 0.626 & 0.926 & 0.935 & 0.894 & 0.754 \\
\bottomrule
\end{tabular}
}
\end{table*}

\begin{figure*}[tbp]
    \centering
    \includegraphics[width=0.95\textwidth]{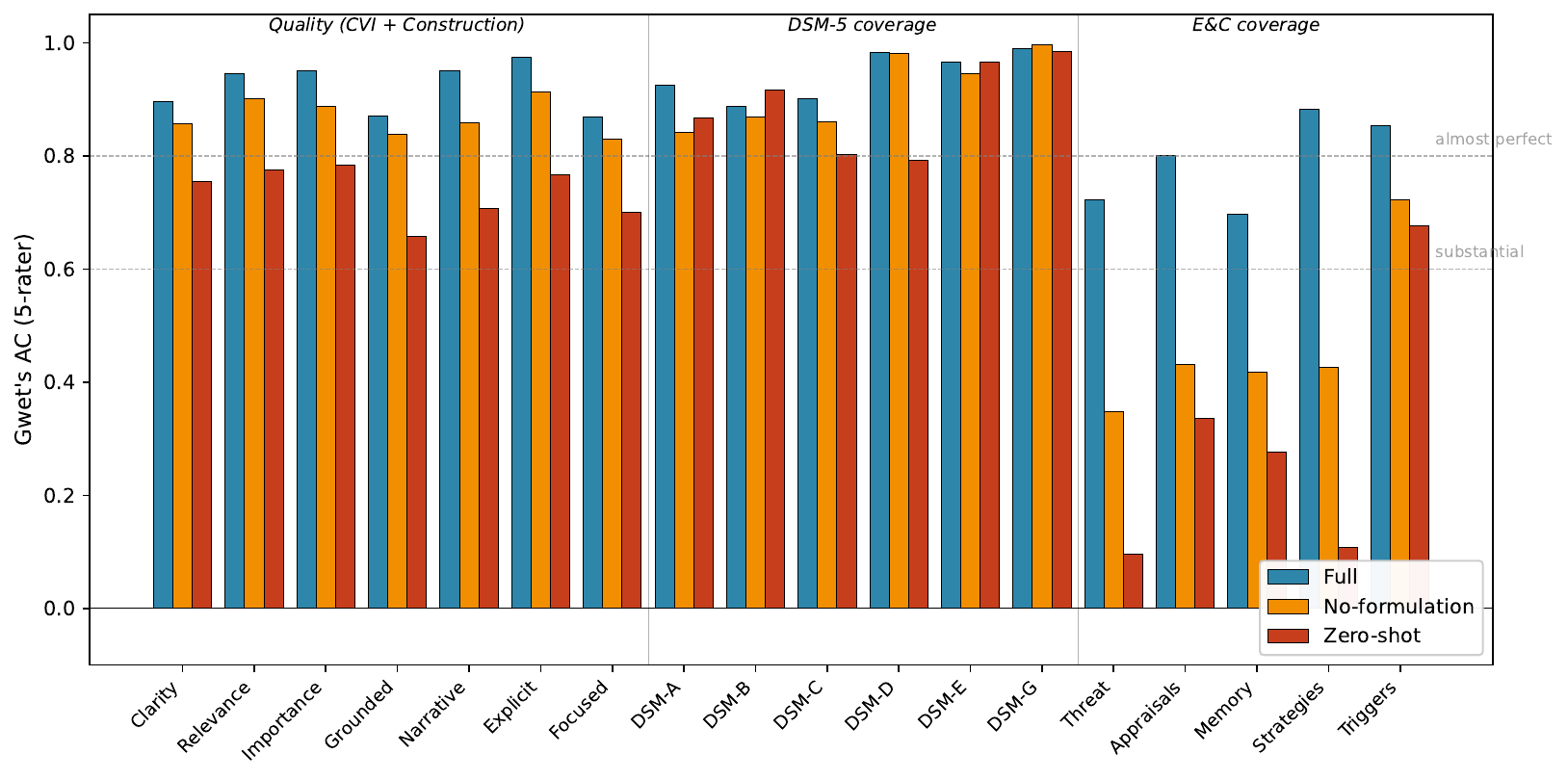}
    \caption{Multi-rater Gwet AC across all five raters (R1, R2, Opus, DSpro, Flash), per dimension and per condition ($n = 110$ vignettes per condition). The Ehlers~\&~Clark items show the strongest conditional gradient: agreement collapses in zero-shot because the components are not explicitly specified into the vignette. Dashed lines mark substantial ($0.6$) and almost-perfect ($0.8$) agreement thresholds.}
    \label{fig:multirater-by-condition}
\end{figure*}

\label{app:fig-multirater}
Figure~\ref{fig:multirater-by-condition} visualises the multi-rater Gwet AC across all five raters (R1, R2, Opus, DSpro, Flash) on the $330$-vignette evaluation subsample, per dimension and per condition. Where Table~\ref{tab:judge-consensus} in the body reports cluster summaries (R1--R2, human--LLM mean, LLM--LLM mean, All 5), the figure makes the conditional gradient on each individual metric visible at a glance.

Three patterns are apparent. First, the ordinal items (clarity, relevance, importance, grounded, narrative, explicit, focused) lie in the $0.65$--$0.99$ band in every condition; the conditional gradient on these is small ($\sim 0.05$--$0.15$ AC2 units). The interpretation: rating these items is largely a property of the rater's calibration rather than of the text being structured, and the formulation modulates absolute rating means (Appendix~\ref{app:expert-by-condition}) more than it modulates agreement. Second, the DSM-5 binary items are at or near ceiling in all conditions ($\geq 0.79$) with a similar small gradient: a competent PTSD vignette satisfies the DSM checklist regardless of formulation, so raters have little to disagree about. Third, the Ehlers~\&~Clark items show the dramatic gradient that drives the body claim: All-5 AC for Threat falls from $0.72$ in full to $0.10$ in zero-shot, Strategies from $0.88$ to $0.11$, Memory from $0.70$ to $0.28$, and Appraisals from $0.80$ to $0.34$. Triggers is the EC outlier with a smaller gradient ($0.85 \to 0.68$), consistent with the narrative-familiarity argument in the body.

The dashed lines at AC $= 0.6$ (substantial agreement) and $0.8$ (almost-perfect agreement) follow the Landis--Koch interpretive bands. Under the formulation specification, every EC item except Triggers crosses into the substantial-or-higher band; under zero-shot, Threat and Strategies sit in the slight-agreement band. The figure is the visual companion to the body's central methodological claim that the formulation gives raters something specifiable to agree about.

\section{Mean Expert Rating per Metric and Condition}
\label{app:expert-by-condition}
Table~\ref{tab:expert-by-condition} reports the human-expert ratings underlying the body claim in \S\ref{sec:res-expert}. The pairing unit is (persona $\times$ generation model), yielding $n = 110$ paired observations per pairwise contrast within each condition column. We report rating means averaged across R1 and R2 at the vignette level, an omnibus statistic for the three-condition effect, and post-hoc paired contrasts.

For ordinal items the omnibus is a one-way repeated-measures ANOVA on the three condition levels with the (persona, model) unit as the within-subjects factor; for binary items the omnibus is Cochran's $Q$, which tests whether the proportions in three matched samples are equal under the null of no condition effect. We use Cochran's $Q$ rather than McNemar (which only tests two conditions) and rather than RM-ANOVA on $0/1$ data (which assumes normality and is inappropriate at the proportions involved). The omnibus must be significant before the pairwise contrasts are interpretable; rows where the omnibus does not reach $p < .05$ (after correction) are marked with $\dagger$ and the pairwise effect-size columns are left blank.

The pairwise contrast statistic is Cohen's $d_z$, the standardised mean of the within-pair differences: $d_z = \bar{D} / \mathrm{SD}(D)$ where $D_i$ is the (persona, model)-paired difference between two conditions. $d_z$ is appropriate for within-subject designs because the relevant denominator is the variance of the difference, not the variance of the raw scores. $d_z$ values are not directly comparable to between-subject Cohen's $d$ from independent samples; they will typically be larger because the within-pair variance is smaller than the between-subject variance whenever there is positive within-subject correlation. $p$-values on the pairwise contrasts are Bonferroni-corrected at $k = 3$ (three pairwise tests per metric).

Reading the table: the ordinal items show the largest full-vs-zero-shot contrasts on importance ($d_z = +1.98$), grounded ($+0.87$), narrative ($+0.75$), and relevance ($+0.70$); clarity is large too ($+0.81$). All three pairwise contrasts on the ordinal items are significant at $p < .05$ corrected. The DSM-5 items show a small but consistent full-vs-zero-shot advantage on Trauma exposure, Intrusion, and Avoidance; Negative cognitions, Hyperarousal, and Impairment saturate at ceiling and the omnibus does not reach significance. The Ehlers~\&~Clark items show the largest effects of any block: Threat $d_z = +1.58$ for full vs zero-shot, Memory $+0.71$, Appraisals $+0.88$, Strategies $+0.76$, while Triggers is non-significant for the reason discussed in the body (Triggers saturate even in zero-shot).

\begin{table*}[h]
\centering
\small
\setlength{\tabcolsep}{4pt}
\caption{Mean human expert rating (R1, R2 averaged) per metric and condition on the 330-vignette subsample. Ordinal items 1--3, binary items in $[0, 1]$; bolded means mark the maximum per row. \textbf{Omnibus}: one-way repeated-measures ANOVA $F$ for ordinal items, Cochran's Q for binary items ($n = 110$ $($persona $\times$ model$)$ units; $df_\text{ANOVA} = (2, 218)$, $df_Q = 2$). Pairwise contrasts report paired Cohen's $d_z$ with Bonferroni-corrected $p$ at $k = 3$: $^{*}\,p<.05$, $^{**}\,p<.01$, $^{***}\,p<.001$. Pairwise contrasts are interpretable only when the omnibus is significant; rows where the omnibus is non-significant are marked $^{\dagger}$ in the omnibus column. F = full, N = no-formulation, Z = zero-shot.}
\label{tab:expert-by-condition}
\resizebox{\linewidth}{!}{%
\begin{tabular}{lrrrlrrr}
\toprule
\textbf{Metric} & \textbf{F} & \textbf{N} & \textbf{Z} & \textbf{Omnibus} & \textbf{F vs N} & \textbf{F vs Z} & \textbf{N vs Z} \\
\midrule
\multicolumn{8}{l}{\textit{Ordinal (1--3)}} \\
Clarity     & \textbf{2.89} & 2.57 & 2.30 & $F\!=\!31.1^{***}$  & $+0.49^{***}$ & $+0.81^{***}$ & $+0.28^{*}$ \\
Relevance   & \textbf{2.85} & 2.61 & 2.35 & $F\!=\!27.5^{***}$  & $+0.40^{***}$ & $+0.70^{***}$ & $+0.33^{**}$ \\
Importance  & \textbf{2.94} & 2.67 & 2.23 & $F\!=\!129.4^{***}$ & $+0.62^{***}$ & $+1.98^{***}$ & $+0.76^{***}$ \\
Grounded    & \textbf{2.86} & 2.53 & 2.20 & $F\!=\!41.6^{***}$  & $+0.52^{***}$ & $+0.87^{***}$ & $+0.38^{***}$ \\
Narrative   & \textbf{2.88} & 2.58 & 2.35 & $F\!=\!31.1^{***}$  & $+0.55^{***}$ & $+0.75^{***}$ & $+0.27^{*}$ \\
Explicit    & \textbf{2.87} & 2.62 & 2.33 & $F\!=\!25.4^{***}$  & $+0.43^{***}$ & $+0.65^{***}$ & $+0.31^{**}$ \\
Focused     & \textbf{2.81} & 2.56 & 2.34 & $F\!=\!22.0^{***}$  & $+0.42^{***}$ & $+0.59^{***}$ & $+0.27^{*}$ \\
\midrule
\multicolumn{8}{l}{\textit{DSM-5 coverage}} \\
Trauma exposure     & \textbf{1.00} & \textbf{1.00} & 0.90 & $Q\!=\!10.0^{**}$  & $+0.00$ & $+0.41^{***}$ & $+0.41^{***}$ \\
Intrusion           & \textbf{1.00} & 0.99 & 0.90 & $Q\!=\!16.0^{***}$ & $+0.14$ & $+0.36^{***}$ & $+0.32^{**}$ \\
Avoidance           & \textbf{1.00} & 0.99 & 0.85 & $Q\!=\!14.6^{***}$ & $+0.13$ & $+0.48^{***}$ & $+0.40^{***}$ \\
Negative cognitions & \textbf{1.00} & \textbf{1.00} & 0.97 & $Q\!=\!2.0^{\dagger}$ & --- & --- & --- \\
Hyperarousal        & \textbf{1.00} & \textbf{1.00} & 0.96 & $Q\!=\!4.0^{\dagger}$ & --- & --- & --- \\
Impairment          & \textbf{1.00} & \textbf{1.00} & \textbf{1.00} & $Q\!=\!0.0^{\dagger}$ & --- & --- & --- \\
\midrule
\multicolumn{8}{l}{\textit{Ehlers \& Clark coverage}} \\
Threat        & \textbf{0.86} & 0.56 & 0.19 & $Q\!=\!84.6^{***}$ & $+0.68^{***}$ & $+1.58^{***}$ & $+0.72^{***}$ \\
Appraisals    & \textbf{0.85} & 0.65 & 0.40 & $Q\!=\!35.5^{***}$ & $+0.35^{**}$ & $+0.88^{***}$ & $+0.37^{***}$ \\
Memory        & \textbf{0.91} & 0.76 & 0.53 & $Q\!=\!34.6^{***}$ & $+0.32^{**}$ & $+0.71^{***}$ & $+0.38^{***}$ \\
Strategies    & \textbf{0.85} & 0.65 & 0.44 & $Q\!=\!21.9^{***}$ & $+0.38^{***}$ & $+0.76^{***}$ & $+0.39^{***}$ \\
Triggers      & \textbf{0.85} & 0.76 & 0.69 & $Q\!=\!5.4^{\dagger}$ & --- & --- & --- \\
\bottomrule
\end{tabular}
}
\end{table*}

\begin{figure*}[tbp]
    \centering
    \includegraphics[width=0.95\textwidth]{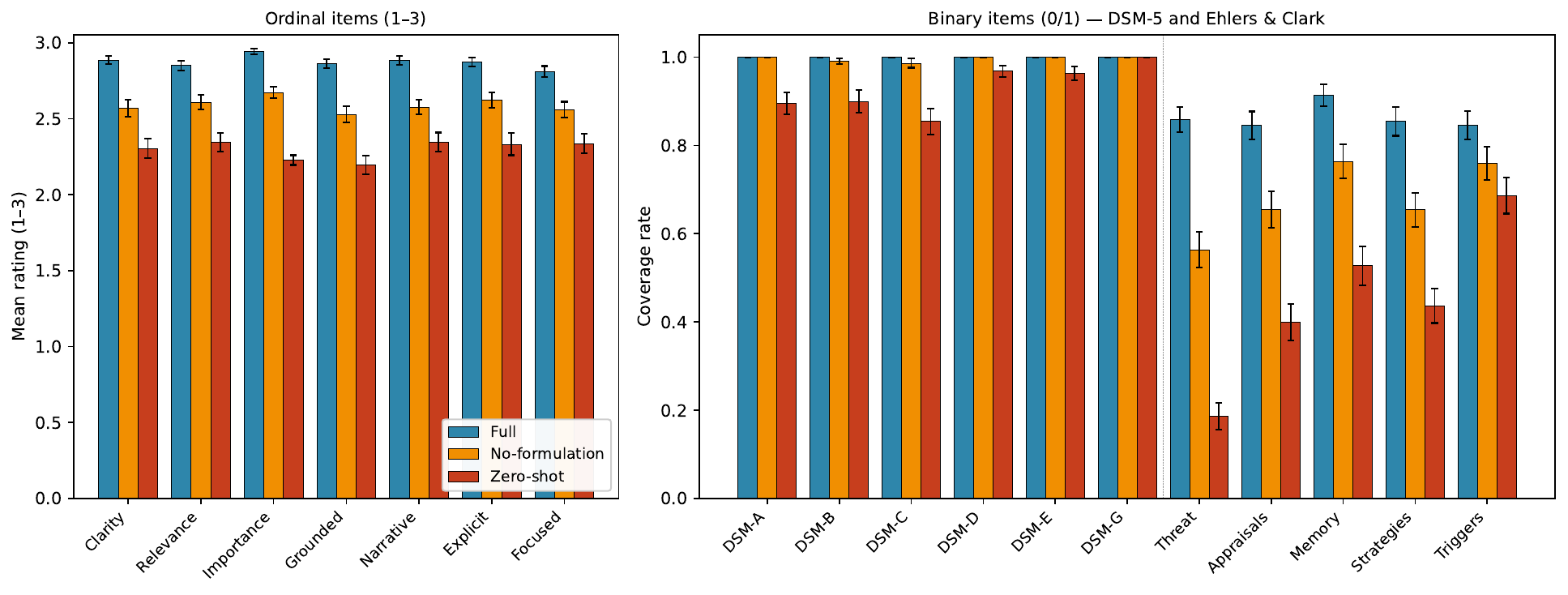}
    \caption{Mean human-expert rating per dimension and condition on the 330-vignette evaluation subsample (R1, R2 averaged). Error bars are $\pm 1$ standard error of the mean across the $n = 110$ vignettes per condition. Left: ordinal items (1--3 scale). Right: binary DSM-5 and Ehlers~\&~Clark coverage. The condition gradient is sharpest on the EC items, mirroring Figure~\ref{fig:multirater-by-condition}.}
    \label{fig:human-by-condition}
\end{figure*}

\label{app:fig-human-by-cond}
Figure~\ref{fig:human-by-condition} visualises the human-expert ratings from Appendix~\ref{app:expert-by-condition} as a per-dimension, per-condition bar chart. The figure makes two patterns easier to see than the table. First, the EC block (right side of the binary panel) shows a steep $L$-shape in zero-shot (Threat $0.19$, Memory $0.53$, Strategies $0.44$, Appraisals $0.40$) that levels out at near-ceiling in full ($0.85$--$0.91$). The DSM-5 block (left side of the binary panel) is flat at ceiling across all three conditions. The contrast between the two blocks within the same panel is the visual essence of the body claim that the formulation pays off on EC content and adds nothing to DSM-5 coverage (which is already saturated).

Second, the ordinal block (left panel) shows a more uniform full $>$ no-formulation $>$ zero-shot gradient across all seven items, with importance and grounded showing the largest gaps. This is the gradient that the scaled judge (Appendix~\ref{app:flash-by-condition}) dampens or in some cases reverses on the same items; the human experts produce the cleaner gradient because they read for clinical meaningfulness rather than for prose smoothness.

Error bars are $\pm 1$ standard error of the mean across the $n = 110$ (persona, model) observations per condition. The bars are computed on the R1, R2 average; the per-rater means and the pairwise rater agreement are in Appendices~\ref{app:judge-means} and \ref{app:full-pairwise}.

% ===== Scaled-Judge Ratings by Condition =====

\section{Scaled-Judge Mean Rating per Metric and Condition}
\label{app:flash-by-condition}
Table~\ref{tab:flash-by-condition} reports the scaled \texttt{deepseek-v4-flash} judge's mean rating per metric and condition on the $15{,}000$-vignette generation set after \texttt{gpt-4o-mini} exclusion (\S\ref{sec:vignette-validation}; $n = 5{,}000$ per condition, $10$ generators $\times$ $500$ personas). The pairing unit is the (generator, persona) cell: each of the $5{,}000$ cells contributes one rating per condition, and pairwise contrasts use the within-cell difference, so Cohen's $d$ is the standardised mean of the $n = 5{,}000$ paired differences.

Reading the table requires bearing in mind that the \texttt{deepseek-v4-flash} judge is sensitive to surface properties of the text (Appendix~\ref{app:judge-means}) and therefore the patterns on the ordinal items are not directly comparable to the expert ratings. \textbf{Clarity}: zero-shot edges out full ($d = -0.08$); the unconstrained vignette reads more fluently to the LLM judge even though human experts disagree (Appendix~\ref{app:expert-by-condition}). \textbf{Importance and Narrative}: full $>$ no-formulation $>$ zero-shot ($d = +0.52$ and $+0.76$ for full-vs-zero-shot); these are the ordinal items most closely tied to clinical content, and the LLM judge agrees with the human experts on direction. \textbf{Explicit}: at ceiling in all conditions; non-discriminative. \textbf{Relevant} (relevance to PTSD): full $>$ zero-shot with $d = +0.19$, smaller than the human-expert contrast.

On the DSM-5 binary items the patterns are largely flat or mildly favour zero-shot: zero-shot vignettes are intrusion- and hyperarousal-heavy because they default to the most narratively familiar criteria. The Ehlers~\&~Clark binary items show the same pattern as in the human-expert evaluation but with smaller effect sizes ($d = +0.55$ on Appraisals, $+0.76$ on Strategies, $+0.26$ on Memory for full vs zero-shot), and Triggers shows a small advantage for zero-shot ($d = -0.07$) for the saturation reason discussed throughout.

A note on power and corrections: with $n = 5{,}000$ paired observations per contrast, even very small effect sizes ($d \sim 0.03$) reach the Bonferroni-corrected significance threshold ($k = 3$ per metric). We report effect sizes alongside significance markers so the reader can see which differences are statistically significant but practically negligible (e.g., clarity full-vs-no-form $d = -0.09$, $p^{***}$). The practical pattern is on the EC items and on importance/narrative.

\begin{table*}[h]
\centering
\small
\caption{Mean \texttt{deepseek-v4-flash} rating per metric and condition on the $15{,}000$-vignette generation set after \texttt{gpt-4o-mini} exclusion ($n = 5{,}000$ per condition; see \S\ref{sec:vignette-validation}). Ordinal items 1--3, binary items in $[0, 1]$; bolded means mark the maximum per row. Effect-size columns are paired Cohen's $d$; significance: $^{*}\,p < .05$, $^{**}\,p < .01$, $^{***}\,p < .001$ (Bonferroni-corrected, $k = 3$).}
\label{tab:flash-by-condition}
\resizebox{\linewidth}{!}{%
\begin{tabular}{lrrrrrr}
\toprule
\textbf{Metric} & \textbf{Full} & \textbf{No-form.} & \textbf{Zero-shot} & \textbf{F vs N (d)} & \textbf{F vs Z (d)} & \textbf{N vs Z (d)} \\
\midrule
\multicolumn{7}{l}{\textit{Ordinal (1--3)}} \\
Clarity      & 2.979 & \textbf{2.995} & 2.993 & $-0.09^{***}$ & $-0.08^{***}$ & $+0.01$ \\
Relevance    & 2.965 & \textbf{2.980} & 2.968 & $-0.07^{***}$ & $-0.01$ & $+0.06^{***}$ \\
Importance   & \textbf{2.820} & 2.711 & 2.539 & $+0.23^{***}$ & $+0.52^{***}$ & $+0.31^{***}$ \\
Grounded     & 2.816 & \textbf{2.854} & 2.810 & $-0.13^{***}$ & $+0.02$ & $+0.14^{***}$ \\
Narrative    & \textbf{2.885} & 2.808 & 2.474 & $+0.20^{***}$ & $+0.76^{***}$ & $+0.59^{***}$ \\
Explicit     & 2.994 & 2.997 & \textbf{2.999} & $-0.03^{*}$ & $-0.05^{***}$ & $-0.03$ \\
Relevant     & \textbf{2.909} & 2.898 & 2.827 & $+0.03$ & $+0.19^{***}$ & $+0.16^{***}$ \\
\midrule
\multicolumn{7}{l}{\textit{DSM-5 coverage}} \\
Trauma exposure     & 0.885 & 0.853 & \textbf{0.902} & $+0.08^{***}$ & $-0.05^{**}$ & $-0.12^{***}$ \\
Intrusion           & 0.912 & 0.925 & \textbf{0.972} & $-0.04^{**}$ & $-0.19^{***}$ & $-0.15^{***}$ \\
Avoidance           & 0.901 & \textbf{0.929} & 0.920 & $-0.08^{***}$ & $-0.05^{**}$ & $+0.03$ \\
Negative cognitions & 0.898 & \textbf{0.918} & 0.861 & $-0.06^{***}$ & $+0.08^{***}$ & $+0.13^{***}$ \\
Hyperarousal        & 0.924 & 0.934 & \textbf{0.999} & $-0.03^{*}$ & $-0.28^{***}$ & $-0.26^{***}$ \\
Impairment          & 0.997 & \textbf{1.000} & 0.983 & $-0.04^{*}$ & $+0.10^{***}$ & $+0.13^{***}$ \\
\midrule
\multicolumn{7}{l}{\textit{Ehlers \& Clark coverage}} \\
Threat        & \textbf{0.741} & 0.690 & 0.668 & $+0.09^{***}$ & $+0.12^{***}$ & $+0.04^{*}$ \\
Appraisals    & 0.853 & \textbf{0.855} & 0.521 & $-0.01$ & $+0.55^{***}$ & $+0.56^{***}$ \\
Memory        & 0.599 & \textbf{0.610} & 0.423 & $-0.02$ & $+0.26^{***}$ & $+0.28^{***}$ \\
Strategies    & \textbf{0.914} & 0.734 & 0.485 & $+0.37^{***}$ & $+0.76^{***}$ & $+0.38^{***}$ \\
Triggers      & 0.934 & 0.929 & \textbf{0.956} & $+0.01$ & $-0.07^{***}$ & $-0.08^{***}$ \\
\bottomrule
\end{tabular}}
\end{table*}

% ===== Clinician User Study =====

\section{Mean Rating per Rater per Condition}
\label{app:judge-means}
Table~\ref{tab:judge-means} reports the mean rating per metric, per rater, per condition on the $330$-vignette evaluation subsample. R1, R2 are the two clinical experts; O, Pr, Fl are the three LLM judges (\texttt{claude-opus-4-5}, \texttt{deepseek-v4-pro}, \texttt{deepseek-v4-flash}). Where the agreement analysis (Appendix~\ref{app:full-pairwise}) asks whether raters agree, this table asks where the raters land on average and whether their location shifts with condition. The two analyses can come apart: raters with high agreement may converge on different absolute means than raters with low agreement, and condition effects on means can co-exist with condition effects on agreement.

Three patterns. (i) On the ordinal items, the two clinical experts (R1, R2) show the strongest condition gradient: clarity falls from $2.88$ in full to $2.30$ in zero-shot for the R1--R2 average, importance from $2.94$ to $2.23$. The three LLM judges show a much weaker ordinal gradient and in some cases the gradient is \emph{reversed} (clarity in O, Pr, Fl rises slightly from full to zero-shot, $2.67/2.75/2.81 \to 2.93/2.99/2.98$). This is the well-known LLM-judge preference for unconstrained prose: a zero-shot vignette is structurally simpler and reads as more fluent to an LLM judge, even when human experts rate it as less clinically meaningful. The two-rater-population pattern from Appendix~\ref{app:full-pairwise} reappears here as a divergence in the direction of the conditional effect, which is a stronger signature than a divergence in the magnitude. (ii) On the DSM-5 binary items every rater is near ceiling in every condition; the criteria are too broad to discriminate. (iii) On the EC items every rater shows the full $>$ no-formulation $>$ zero-shot gradient, and the gradient is largest on Threat, Strategies, Appraisals (the body claim). On these items, the human experts and LLMs agree about the \emph{direction} of the condition effect even though they disagree about the absolute level (Appendix~\ref{app:full-pairwise}).

The practical implication for downstream evaluation of synthetic clinical text: relying on an LLM judge alone for ordinal-quality items risks under-estimating the contribution of structured specification, because the LLM-judge gradient is dampened relative to the human gradient. The EC items, by contrast, are robust across rater types in direction, and either rater pool will reveal the formulation effect there.

\begin{table*}[h]
\centering
\small
\setlength{\tabcolsep}{5pt}
\caption{Mean rating per metric per rater per condition on the 330-vignette evaluation subsample ($n = 110$ per condition). Ordinal items 1--3; binary items in $[0, 1]$. Within each condition block, ordinal/DSM/E\&C row groupings are the same as Table~\ref{tab:judge-consensus}.}
\label{tab:judge-means}
\resizebox{\linewidth}{!}{%
\begin{tabular}{l rrrrr rrrrr rrrrr}
\toprule
& \multicolumn{5}{c}{\textbf{Full}} & \multicolumn{5}{c}{\textbf{No-formulation}} & \multicolumn{5}{c}{\textbf{Zero-shot}} \\
\cmidrule(lr){2-6} \cmidrule(lr){7-11} \cmidrule(lr){12-16}
\textbf{Metric} & R1 & R2 & O & Pr & Fl & R1 & R2 & O & Pr & Fl & R1 & R2 & O & Pr & Fl \\
\midrule
\multicolumn{16}{l}{\textit{Ordinal (1--3)}} \\
Clarity     & 2.90 & 2.87 & 2.67 & 2.75 & 2.81 & 2.61 & 2.53 & 2.91 & 2.96 & 2.99 & 2.35 & 2.25 & 2.93 & 2.99 & 2.98 \\
Relevance   & 2.86 & 2.84 & 2.85 & 2.94 & 2.91 & 2.64 & 2.58 & 2.94 & 2.99 & 2.93 & 2.38 & 2.31 & 2.91 & 2.97 & 2.90 \\
Importance  & 2.92 & 2.96 & 2.81 & 2.88 & 2.81 & 2.62 & 2.73 & 2.91 & 2.86 & 2.74 & 2.17 & 2.28 & 2.76 & 2.69 & 2.46 \\
Grounded    & 2.89 & 2.84 & 2.62 & 2.75 & 2.68 & 2.56 & 2.50 & 2.79 & 2.91 & 2.83 & 2.22 & 2.17 & 2.78 & 2.75 & 2.75 \\
Narrative   & 2.90 & 2.86 & 2.85 & 2.91 & 2.84 & 2.57 & 2.58 & 2.89 & 2.87 & 2.78 & 2.36 & 2.33 & 2.65 & 2.64 & 2.35 \\
Explicit    & 2.90 & 2.85 & 3.00 & 2.99 & 3.00 & 2.65 & 2.59 & 3.00 & 3.00 & 3.00 & 2.35 & 2.32 & 2.99 & 3.00 & 3.00 \\
Focused     & 2.82 & 2.80 & 2.73 & 2.73 & 2.78 & 2.60 & 2.52 & 2.86 & 2.78 & 2.88 & 2.38 & 2.29 & 2.73 & 2.75 & 2.75 \\
\midrule
\multicolumn{16}{l}{\textit{DSM-5 coverage}} \\
Trauma exposure     & 1.00 & 1.00 & 0.97 & 0.91 & 0.91 & 1.00 & 1.00 & 0.94 & 0.76 & 0.86 & 0.88 & 0.91 & 0.99 & 0.96 & 0.92 \\
Intrusion           & 1.00 & 1.00 & 0.87 & 0.86 & 0.85 & 0.99 & 0.99 & 0.85 & 0.86 & 0.84 & 0.90 & 0.90 & 0.99 & 0.97 & 1.00 \\
Avoidance           & 1.00 & 1.00 & 0.89 & 0.86 & 0.90 & 0.99 & 0.98 & 0.87 & 0.83 & 0.93 & 0.89 & 0.82 & 0.89 & 0.92 & 0.96 \\
Negative cognitions & 1.00 & 1.00 & 0.96 & 1.00 & 0.99 & 1.00 & 1.00 & 0.98 & 0.97 & 1.00 & 0.96 & 0.97 & 0.79 & 0.85 & 0.84 \\
Hyperarousal        & 1.00 & 1.00 & 0.94 & 0.97 & 0.96 & 1.00 & 1.00 & 0.93 & 0.94 & 0.94 & 0.98 & 0.94 & 1.00 & 0.99 & 0.99 \\
Impairment          & 1.00 & 1.00 & 0.99 & 0.99 & 0.99 & 1.00 & 1.00 & 1.00 & 0.99 & 1.00 & 1.00 & 1.00 & 0.97 & 1.00 & 0.98 \\
\midrule
\multicolumn{16}{l}{\textit{Ehlers \& Clark coverage}} \\
Threat        & 0.93 & 0.79 & 0.95 & 0.78 & 0.78 & 0.67 & 0.46 & 0.85 & 0.62 & 0.71 & 0.17 & 0.20 & 0.69 & 0.61 & 0.64 \\
Appraisals    & 0.84 & 0.84 & 0.92 & 0.87 & 0.91 & 0.70 & 0.61 & 0.74 & 0.83 & 0.78 & 0.35 & 0.45 & 0.42 & 0.44 & 0.46 \\
Memory        & 0.92 & 0.91 & 0.77 & 0.71 & 0.71 & 0.79 & 0.74 & 0.73 & 0.63 & 0.68 & 0.53 & 0.53 & 0.43 & 0.48 & 0.46 \\
Strategies    & 0.85 & 0.85 & 0.94 & 0.93 & 0.96 & 0.72 & 0.59 & 0.84 & 0.83 & 0.64 & 0.35 & 0.52 & 0.57 & 0.70 & 0.49 \\
Triggers      & 0.86 & 0.83 & 0.93 & 0.92 & 0.94 & 0.80 & 0.72 & 0.97 & 0.92 & 0.92 & 0.68 & 0.69 & 0.92 & 0.88 & 0.96 \\
\bottomrule
\end{tabular}}

\end{table*}

% ===== Expert Validation by Condition =====

\section{Clinician User-Study Sample Description}
\label{app:us-sample}

\paragraph{Recruitment, compensation, and ethics.}
Participants were recruited via an online research panel using an inclusion criterion of holding a current clinical licence and having direct caseload experience with PTSD patients. Recruits gave informed consent before any rating activity and received a debriefing after completing or exiting the study. Compensation was \pounds$7.50$ per hour, pro-rated to each participant's actual time spent on the survey (computed from the platform's timestamped entry-and-exit log). The study was reviewed and approved by the ethics committee of the Department of Industrial Engineering and Management, Ben-Gurion University of the Negev, prior to launch; participant identifiers are randomised in the released dataset and demographic combinations that could re-identify individual respondents are coarsened or suppressed.

\paragraph{Sample.}
The $N = 100$ recruited clinicians were $65\%$ female, all native English speakers, all currently licensed.
\textbf{Roles}: $51\%$ nurses (including $5\%$ mental-health nurses), $18\%$ psychologists, $17\%$ MD general practitioners, $9\%$ psychiatrists.
\textbf{PTSD caseload}: all reported experience with PTSD patients ($37\%$ regularly, $63\%$ occasionally).
\textbf{Country}: $52\%$ United States, $24\%$ United Kingdom, $17\%$ Canada, $7\%$ Australia.
\textbf{Highest educational degree}: $46\%$ Bachelor's, $26\%$ Master's, $28\%$ Doctorate (PhD, PsyD, or MD).
\textbf{Years of clinical practice}: median $10$ (range $1$--$32$).
\textbf{Survey completion time}: median $70.1$ minutes (IQR $46.1$--$90.0$; range $25.3$--$159.1$).

\paragraph{Ratings and attention checks.}
Each participant rated between $17$ and $18$ randomly assigned vignettes plus two attention-check items (\texttt{ATTN1}, \texttt{ATTN2}) embedded in the rating stream. The released dataset comprises $1{,}706$ total ratings: $1{,}600$ vignette ratings (the analytic dataset used in \S\ref{sec:user-study}; $532$ full $+ 541$ no-formulation $+ 527$ zero-shot) and $106$ attention-check responses ($53$ on each of \texttt{ATTN1} and \texttt{ATTN2}). Attention-check responses are released alongside the vignette ratings so downstream users can apply alternative attention-screening criteria; we did not exclude any participant from the analytic dataset on the basis of attention-check performance. Raw per-condition rating means are given in Table~\ref{tab:us-desc}.
%\textbf{Data we did not collect}: the recruitment instrument did not record clinician ethnicity, race, age, religion, political affiliation, or socioeconomic status. We can therefore audit the rater pool for sex, country, professional role, training level, language, and PTSD-caseload experience, but not for ethnic/racial representation. Generalisability to clinicians outside the recorded profile (predominantly female, predominantly US/UK/Canada/Australia, predominantly nursing-trained) is a question the present data cannot answer; we flag it in Limitations.
%Demographic distributions of the persona set rated by these clinicians are in Appendix~\ref{sec:appendix-demographics}.

\begin{table}[h]
\centering
\small
\setlength{\tabcolsep}{4pt}
\caption{Raw clinician rating means per condition. Mean $\pm$ SD; $n$ = ratings per condition. Quality items are $1$--$3$; AI authorship is $0/1$.}
\label{tab:us-desc}
\resizebox{\linewidth}{!}{%
\begin{tabular}{l rrr}
\toprule
\textbf{Metric} & \textbf{Full} ($n{=}532$) & \textbf{No-form.} ($n{=}541$) & \textbf{Zero-shot} ($n{=}527$) \\
\midrule
Clarity        & $2.68 \pm 0.47$ & $2.16 \pm 0.57$ & $1.58 \pm 0.53$ \\
Relevance      & $2.67 \pm 0.49$ & $2.16 \pm 0.54$ & $1.56 \pm 0.53$ \\
Importance     & $2.65 \pm 0.48$ & $2.12 \pm 0.53$ & $1.54 \pm 0.54$ \\
AI authorship  & $0.15 \pm 0.35$ & $0.51 \pm 0.50$ & $0.78 \pm 0.42$ \\
\bottomrule
\end{tabular}
}
\end{table}

\section{Linear Mixed Model Contrasts for User-Study Quality}
\label{app:us-mixed-quality}
Table~\ref{tab:us-mixed-quality} reports the linear mixed model estimates for the perceived-quality contrasts in the user study. The model for each ordinal outcome $y_{ij}$ is
\[
y_{ij} = \beta_0 + \beta_1 \mathrm{Full}_{ij} + \beta_2 \mathrm{NoForm}_{ij} + u_i + v_{j(i)} + \varepsilon_{ij},
\]
where $i$ indexes the $N = 100$ participants, $j$ indexes the rated vignette (a participant rates $\sim 17$ vignettes plus an attention check), $\mathrm{Full}$ and $\mathrm{NoForm}$ are dummy indicators with zero-shot as the reference condition, $u_i$ is a participant-level random intercept, and $v_{j(i)}$ is a variance component capturing residual vignette-level heterogeneity nested in participant. We fit the models in \texttt{statsmodels} using \texttt{MixedLM} with restricted maximum likelihood.

Several modelling choices are worth flagging. First, we treat the ordinal $1$--$3$ items as approximately continuous; the alternative ordinal-logit specification is unstable on three-level outcomes with strong floor/ceiling and produces qualitatively similar contrasts but inflated standard errors. Second, we cluster on participant via the random intercept rather than via cluster-robust standard errors on a fixed-effects model, because the random-intercept specification recovers the residual variance partition (ICC) that informs the interpretation of how much of the rating variance is participant idiosyncrasy. The estimated ICC is $\sim 0.08$, indicating that $\sim 8\%$ of the residual variance after fixed effects is between-participant baseline strictness, with the remainder being vignette-level variation (which the condition fixed effects then explain a substantial fraction of). Third, we do not include random slopes for condition because the design assigns each participant a mix of conditions but the per-(participant, condition) cell size is small; random-intercept-only is a deliberate, more conservative specification.

The contrast estimates in Table~\ref{tab:us-mixed-quality} are differences in fitted means on the $1$--$3$ scale. The full-vs-zero-shot contrasts are all $b > +1.08$, $t > +34$, $p < 10^{-250}$: clinicians rate full vignettes more than one full point higher than zero-shot vignettes by the same clinician, after adjusting for participant and vignette random variation. The full-vs-no-formulation contrasts (the marginal contribution of the cognitive graph, controlling for the persona's self-report items) are all $b > +0.50$ on the $1$--$3$ scale, $t > +16$. The intermediate no-formulation contrasts are about half the size of the full contrasts on every metric, consistent with the body's claim that the self-report items account for roughly half of the formulation lift and the graph contributes the remaining half.

The CIs are Wald CIs from the asymptotic variance estimate, and we present them on the rating-scale natively (rather than as standardised effects) because the rating scale is interpretable: a difference of $+1.0$ on the $1$--$3$ scale spans nearly half the rating range.

\begin{table*}[h]
\centering
\small
\setlength{\tabcolsep}{4pt}
\caption{Linear mixed model contrasts for perceived quality ratings, with random intercepts for participant ($n = 100$) and a variance component for vignette item ($n = 229$). Reference condition is zero-shot. $b$ is the estimated rating difference on the 1--3 scale; positive values mean ``higher with the formulation''. $N = 1{,}600$ ratings.}
\label{tab:us-mixed-quality}
\resizebox{\linewidth}{!}{%
\begin{tabular}{l l rrrrr}
\toprule
\textbf{Outcome} & \textbf{Contrast} & $\bm{b}$ & \textbf{SE} & \textbf{95\% CI} & $\bm{t}$ & $\bm{p}$ \\
\midrule
Clarity     & Full vs Zero-shot         & $+1.087$ & 0.032 & $[+1.024, +1.149]$ & $+34.16$ & $<10^{-250}$ \\
Clarity     & No-form. vs Zero-shot     & $+0.565$ & 0.032 & $[+0.502, +0.627]$ & $+17.80$ & $<10^{-70}$  \\
Clarity     & Full vs No-form.          & $+0.522$ & 0.032 & $[+0.461, +0.584]$ & $+16.54$ & $<10^{-61}$  \\
\midrule
Relevance   & Full vs Zero-shot         & $+1.109$ & 0.032 & $[+1.047, +1.172]$ & $+34.96$ & $<10^{-250}$ \\
Relevance   & No-form. vs Zero-shot     & $+0.597$ & 0.032 & $[+0.535, +0.659]$ & $+18.89$ & $<10^{-78}$  \\
Relevance   & Full vs No-form.          & $+0.512$ & 0.032 & $[+0.450, +0.574]$ & $+16.25$ & $<10^{-58}$  \\
\midrule
Importance  & Full vs Zero-shot         & $+1.111$ & 0.031 & $[+1.050, +1.173]$ & $+35.40$ & $<10^{-250}$ \\
Importance  & No-form. vs Zero-shot     & $+0.583$ & 0.031 & $[+0.522, +0.644]$ & $+18.64$ & $<10^{-77}$  \\
Importance  & Full vs No-form.          & $+0.528$ & 0.031 & $[+0.467, +0.589]$ & $+16.94$ & $<10^{-63}$  \\
\bottomrule
\end{tabular}}
\end{table*}

\label{app:fig-effect-sizes}

\begin{figure*}[tbp]
    \centering
    \includegraphics[width=0.95\textwidth]{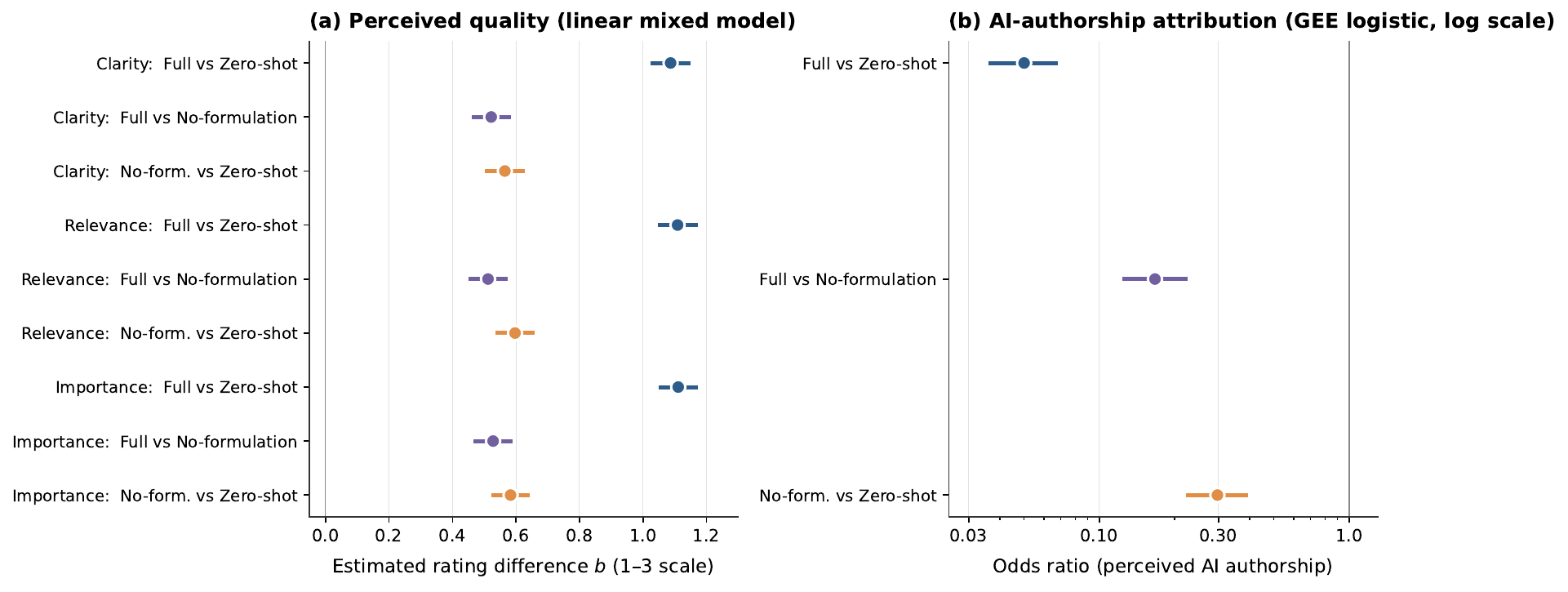}
    \caption{Effect-size summary of condition contrasts from the user study ($N = 100$ clinicians, $1{,}600$ ratings). (a) Linear mixed model estimates for perceived quality (rating-point difference on the $1$--$3$ scale). (b) GEE logistic odds ratios for perceived AI authorship (log scale; values $<1$ favour the first-named condition). Points: point estimates; bars: $95\%$ Wald CIs. All nine quality contrasts and all three authorship contrasts are significant at $p < 10^{-15}$.}
    \label{fig:us-effect-sizes}
\end{figure*}

Figure~\ref{fig:us-effect-sizes} summarises the user-study condition contrasts as an effect-size forest plot. Panel (a) shows linear-mixed-model rating-point differences on the $1$--$3$ quality scale; panel (b) shows GEE logistic odds ratios for perceived AI authorship on the log scale. All nine quality contrasts and all three authorship contrasts are significant at $p < 10^{-15}$.

\section{GEE Logistic Regression on Perceived AI Authorship}
\label{app:us-ai-gee}
Table~\ref{tab:us-ai-gee} reports the GEE logistic regression on perceived AI authorship for the user-study contrasts. The outcome is the binary indicator $y_{ij} = 1$ if participant $i$ judges vignette $j$ to be AI-generated and $0$ otherwise. The model is
\[
\mathrm{logit}\,\Pr(y_{ij}=1) = \beta_0 + \beta_1 \mathrm{Full}_{ij} + \beta_2 \mathrm{NoForm}_{ij},
\]
with zero-shot as the reference condition. We fit it with generalised estimating equations using an exchangeable working correlation structure clustered on participant ($i$), which assumes that within-participant residual correlations are constant across pairs of ratings by the same participant.

The choice of GEE over a mixed-effects logistic model is deliberate. Mixed logistic models estimate \emph{conditional} (subject-specific) odds ratios: the change in the log odds for a typical participant. GEE estimates \emph{marginal} (population-averaged) odds ratios: the average change in log odds across the population of clinicians. For a fairness-and-deployment claim we want the population-averaged answer ``how much less likely is a randomly selected clinician to attribute a full-condition vignette to AI?'' rather than the conditional ``how much less likely is participant $i$ specifically?''. The two estimates differ when the random-intercept variance is large; in our data the participant ICC is modest ($\sim 0.08$), so the difference is small, but GEE is the correct specification for the marginal question.

The reported odds ratios are large: $\mathrm{OR}_{\text{full vs zs}} = 0.050$ (95\% CI $[0.036, 0.068]$, $z = -18.5$). This is a $\sim 20\times$ reduction in the odds of an AI-authorship attribution from zero-shot to full, for the same clinician on average. The no-formulation contrast against zero-shot is $\mathrm{OR} = 0.297$ (a $\sim 3.4\times$ reduction); the full-vs-no-formulation contrast is $\mathrm{OR} = 0.167$ (a $\sim 6\times$ reduction), so the marginal contribution of the cognitive graph beyond the self-report items is itself substantial. All three $z$-statistics exceed $-8.3$, corresponding to $p < 10^{-15}$.

The CIs are Wald CIs on the OR scale, transformed from the log-OR scale; the sandwich variance estimator is used, which is robust to misspecification of the working correlation structure. We checked sensitivity to the working correlation by re-fitting with independence and AR(1) structures; estimates change by less than $0.5\%$.

A cautionary remark on the interpretation. The OR is on the binary AI-attribution outcome, not on a graded plausibility scale. A clinician who judges a vignette ``probably AI but unsure'' is recorded as $1$ here; a clinician who judges ``definitely human'' as $0$. The $14.7\%$ full-condition AI-attribution rate is therefore not the same as ``$85\%$ of vignettes are indistinguishable from human-written'' in any deep sense; it is the operationalised version of human-passing we measured.

\begin{table}[h]
\centering
\small
\setlength{\tabcolsep}{4pt}
\caption{GEE logistic regression on perceived AI authorship (binary), with exchangeable working correlation clustered on participant ($N = 100$). Coefficients on the logit scale; OR is the corresponding odds ratio with 95\% Wald CI. $N = 1{,}600$ ratings.}
\label{tab:us-ai-gee}
\resizebox{\linewidth}{!}{%
\begin{tabular}{l rrrr}
\toprule
\textbf{Contrast} & $\bm{b}$ & \textbf{OR} & \textbf{95\% CI (OR)} & $\bm{z}$ \\
\midrule
Full vs Zero-shot          & $-3.00$ & $0.050$ & $[0.036, 0.068]$ & $-18.54$ \\
No-form. vs Zero-shot      & $-1.22$ & $0.297$ & $[0.223, 0.394]$ & $-8.36$ \\
Full vs No-form.           & $-1.79$ & $0.167$ & $[0.124, 0.226]$ & $-11.70$ \\
\bottomrule
\end{tabular}
}
\end{table}

\section{Clinician User-Study Means per Model and Condition}
\label{app:us-by-model}

Table~\ref{tab:us-by-model} reports the raw clinician rating means per generation model and condition from the user study. Models are ordered by ascending full-condition AI-authorship rate. The full $>$ no-formulation $>$ zero-shot gradient on the CVI items holds for every one of the $11$ generation models; the AI-authorship gradient (lower is better, i.e.\ vignettes pass as human-written more often) holds with one minor swap between \texttt{gpt-4o-mini} and \texttt{gpt-5.4-mini} in the no-formulation column.

\begin{table*}[h]
\centering
\small
\setlength{\tabcolsep}{4pt}
\caption{Raw clinician rating means per generation model and condition. \textbf{Cla} = Clarity, \textbf{Rel} = Relevance, \textbf{Imp} = Importance (all $1$--$3$ scale, higher is better); \textbf{AI} = mean perceived AI-authorship rate ($0$--$1$, lower means the vignettes pass as human-written more often). Models ordered by ascending full-condition AI-authorship rate. $N$ per (model, condition) cell is $43$--$57$ (see \S\ref{sec:us-method}).}
\label{tab:us-by-model}
\resizebox{\linewidth}{!}{%
\begin{tabular}{l rrrr rrrr rrrr}
\toprule
& \multicolumn{4}{c}{\textbf{Full}} & \multicolumn{4}{c}{\textbf{No-formulation}} & \multicolumn{4}{c}{\textbf{Zero-shot}} \\
\cmidrule(lr){2-5} \cmidrule(lr){6-9} \cmidrule(lr){10-13}
\textbf{Model} & Cla & Rel & Imp & AI & Cla & Rel & Imp & AI & Cla & Rel & Imp & AI \\
\midrule
\texttt{gemini-2.5-pro}    & 2.85 & 2.79 & 2.75 & \textbf{0.00} & 2.48 & 2.37 & 2.35 & 0.48 & 1.88 & 1.89 & 1.86 & 0.63 \\
\texttt{gpt-5.4}           & 2.87 & 2.74 & 2.77 & \textbf{0.00} & 2.32 & 2.43 & 2.28 & 0.19 & 1.76 & 1.88 & 1.71 & 0.45 \\
\texttt{claude-sonnet-4-6} & 2.82 & 2.78 & 2.86 & 0.02 & 2.19 & 2.27 & 2.23 & 0.17 & 1.89 & 1.76 & 1.69 & 0.56 \\
\texttt{gemini-2.5-flash}  & 2.71 & 2.79 & 2.81 & 0.10 & 2.19 & 2.09 & 2.09 & 0.51 & 1.48 & 1.46 & 1.50 & 0.83 \\
\texttt{gpt-5.4-mini}      & 2.81 & 2.83 & 2.70 & 0.11 & 2.32 & 2.23 & 2.45 & 0.57 & 1.76 & 1.72 & 1.58 & 0.76 \\
\texttt{gpt-4o-mini}       & 2.47 & 2.53 & 2.38 & 0.13 & 2.06 & 2.11 & 2.09 & 0.45 & 1.52 & 1.61 & 1.54 & 0.85 \\
\texttt{deepseek-reasoner} & 2.67 & 2.67 & 2.69 & 0.16 & 2.04 & 2.02 & 2.08 & 0.61 & 1.49 & 1.43 & 1.49 & 0.88 \\
\texttt{claude-haiku-4-5}  & 2.67 & 2.70 & 2.84 & 0.23 & 2.00 & 2.23 & 1.98 & 0.56 & 1.49 & 1.32 & 1.32 & 0.87 \\
\texttt{deepseek-chat}     & 2.53 & 2.60 & 2.57 & 0.28 & 2.10 & 2.00 & 1.92 & 0.64 & 1.33 & 1.27 & 1.42 & 0.93 \\
\texttt{qwen3.6-35b}       & 2.39 & 2.51 & 2.25 & 0.29 & 2.02 & 2.11 & 1.94 & 0.72 & 1.35 & 1.43 & 1.28 & 0.93 \\
\texttt{qwen2.5-32b}       & 2.63 & 2.41 & 2.55 & 0.31 & 1.98 & 1.92 & 1.96 & 0.67 & 1.40 & 1.30 & 1.47 & 0.91 \\
\bottomrule
\end{tabular}}
\end{table*}

\label{app:fig-us-ratings}

\begin{figure*}[tbp]
    \centering
    \includegraphics[width=0.85\textwidth]{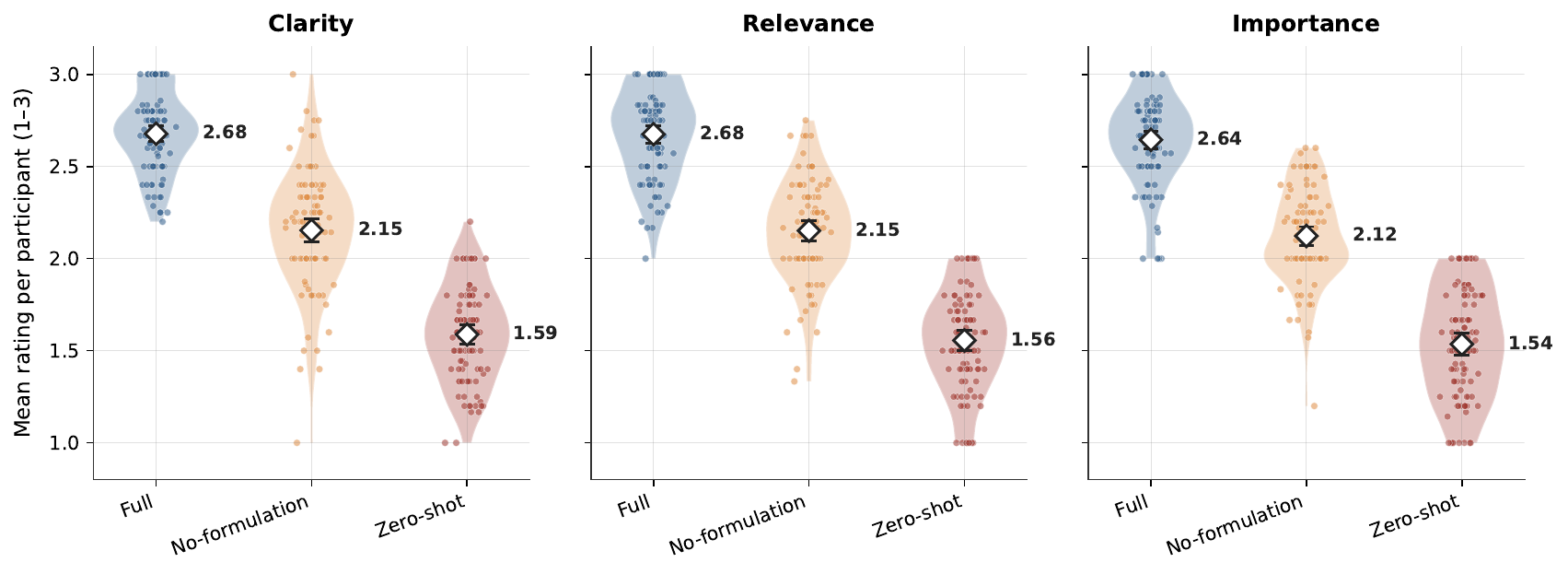}
    \caption{Mean clinician ratings for the three perceived-quality items by condition ($N = 100$ clinicians, $1{,}600$ ratings). Error bars show $\pm 1$ standard error of the mean. The same monotonic gradient is observed across the three CVI items.}
    \label{fig:us-ratings}
\end{figure*}

Figure~\ref{fig:us-ratings} visualises the per-condition distribution of the three quality items in the user study. The same monotonic full $>$ no-formulation $>$ zero-shot gradient is observed across the three CVI items.

% ===== Fairness Analysis =====

\section{Algorithmic-Fairness Metrics by Condition}
\label{app:fairness-metrics}
Table~\ref{tab:fairness-metrics} reports four algorithmic-fairness metrics per (outcome, demographic, condition) cell, computed on the $15{,}000$-vignette scaled-judge set. The metrics are chosen to triangulate disparity rather than to rely on any single quantity. Each captures a different aspect of group-level inequality and they can disagree, which is itself informative.

\textbf{Statistical Parity Difference (SPD)} \citep{dwork2012fairness} is the difference between the largest and smallest group means on the outcome: $\mathrm{SPD} = \max_g \bar{y}_g - \min_g \bar{y}_g$. SPD is in the units of the outcome (rating points on a $1$--$3$ scale for quality, proportion for EC coverage). SPD treats every group symmetrically and does not depend on which group is the reference. \textbf{Disparate Impact Ratio (DIR)} \citep{feldman2015certifying} is $\min_g \bar{y}_g / \max_g \bar{y}_g$; values closer to $1$ indicate parity. The $0.8$ threshold (``four-fifths rule''; \citealt{eeoc1978}) is the long-standing US Equal Employment Opportunity Commission heuristic for flagging disparate impact, and we report it as a reference threshold without endorsing it as a fairness pass/fail criterion. \textbf{$d_{\max}$} is the largest pairwise Cohen's $d$ across all $\binom{G}{2}$ group contrasts on the outcome, and it converts group disparity into a standardised effect-size scale that is comparable across outcomes with different variance.

The last column \textbf{Inter.~$p$} is the $p$-value of the demographic-by-condition interaction term in an OLS regression of the outcome on demographic, condition, and their interaction, with cluster-robust standard errors at the persona level. A significant interaction means the formulation modulates the demographic effect, which is precisely the question we want to answer.

The body claim is supported by every metric in the perceived-quality rows. On gender, ethnicity, age band, and trauma category, SPD falls $1.5$--$7\times$ from zero-shot to full; DIR rises above $0.99$ in full (well above the four-fifths threshold); $d_{\max}$ falls to $\leq 0.12$. The trauma-category disparity is the largest in zero-shot ($\mathrm{SPD} = 0.092$, $d_{\max} = -0.49$) and collapses to negligible in full ($\mathrm{SPD} = 0.013$, $d_{\max} = +0.07$); the interaction is highly significant ($p < .001$).

The EC-coverage rows tell a different story. Gender, ethnicity, and age remain invariant across conditions (DIR $> 0.94$ throughout, $d_{\max} \leq 0.24$ throughout). But trauma-category disparity \emph{widens} under the formulation: $\mathrm{SPD}$ rises from $0.053$ in zero-shot to $0.094$ in full, DIR falls from $0.92$ to $0.89$, and $d_{\max}$ swings from $+0.23$ to $-0.49$. The pattern is interpretable as a content-fit effect: childhood-trauma personas have triggers, threat appraisals, and maladaptive strategies that map naturally onto narrative material, so when the Crafter is instructed to portray all five EC components it does so completely. Loss/medical-trauma personas have a less natural mapping onto Strategies and Triggers, so the same instruction yields a structurally complete vignette in some cases but a thinner one in others. This is the same content-fit pattern we flag in the body and the limitation we note in \citet{blodgett2020language}'s sense: we are measuring output equivariance, not content adequacy across trauma types.

\begin{table*}[h]
\centering
\small
\setlength{\tabcolsep}{4pt}
\caption{Algorithmic-fairness metrics on the $15{,}000$-vignette scaled-judge set, computed separately per condition. \textbf{SPD}: Statistical Parity Difference ($\max - \min$ group mean); \textbf{DIR}: Disparate Impact Ratio ($\min / \max$ group mean); the four-fifths-rule threshold for flagging disparate impact is DIR $< 0.8$. \textbf{$d_\text{max}$}: Cohen's $d$ for the largest pairwise contrast. \textbf{Inter. $p$}: $F$-test of demographic $\times$ condition interaction (cluster-robust at persona level). For perceived-quality outcomes the formulation collapses disparity across every demographic dimension; for EC coverage the trauma-category disparity increases under full (a content-fit effect; see text).}
\label{tab:fairness-metrics}
\resizebox{\linewidth}{!}{%
\begin{tabular}{ll rrr rrr rrr c}
\toprule
& & \multicolumn{3}{c}{\textbf{Zero-shot}} & \multicolumn{3}{c}{\textbf{No-formulation}} & \multicolumn{3}{c}{\textbf{Full}} & \\
\cmidrule(lr){3-5} \cmidrule(lr){6-8} \cmidrule(lr){9-11}
\textbf{Outcome} & \textbf{Demographic} & SPD & DIR & $d_\text{max}$ & SPD & DIR & $d_\text{max}$ & SPD & DIR & $d_\text{max}$ & \textbf{Inter.\ $p$} \\
\midrule
Quality & Gender          & $0.021$ & $0.993$ & $+0.11$ & $0.009$ & $0.997$ & $+0.05$ & $\bm{0.009}$ & $\bm{0.997}$ & $\bm{+0.05}$ & $.064$ \\
Quality & Ethnicity       & $0.037$ & $0.987$ & $+0.19$ & $0.019$ & $0.993$ & $-0.12$ & $\bm{0.021}$ & $\bm{0.993}$ & $\bm{+0.12}$ & $.029^{*}$ \\
Quality & Age band        & $0.019$ & $0.993$ & $+0.10$ & $0.008$ & $0.997$ & $+0.05$ & $\bm{0.013}$ & $\bm{0.995}$ & $\bm{-0.07}$ & $.077$ \\
Quality & Trauma category & $0.092$ & $0.968$ & $-0.49$ & $0.028$ & $0.991$ & $-0.16$ & $\bm{0.013}$ & $\bm{0.996}$ & $\bm{+0.07}$ & $< .001^{***}$ \\
\midrule
EC      & Gender          & $0.005$ & $0.991$ & $-0.03$ & $0.000$ & $1.000$ & $+0.00$ & $0.004$ & $0.995$ & $-0.02$ & $< .001^{***}$ \\
EC      & Ethnicity       & $0.024$ & $0.961$ & $+0.11$ & $0.038$ & $0.952$ & $+0.20$ & $0.047$ & $0.943$ & $+0.24$ & $.915$ \\
EC      & Age band        & $0.011$ & $0.982$ & $-0.05$ & $0.013$ & $0.983$ & $-0.07$ & $0.013$ & $0.984$ & $-0.07$ & $.509$ \\
EC      & Trauma category & $0.053$ & $0.917$ & $+0.23$ & $0.017$ & $0.979$ & $+0.09$ & $0.094$ & $0.892$ & $-0.49$ & $< .001^{***}$ \\
\bottomrule
\end{tabular}}
\end{table*}

\begin{figure*}[tbp]
    \centering
    \includegraphics[width=0.85\textwidth]{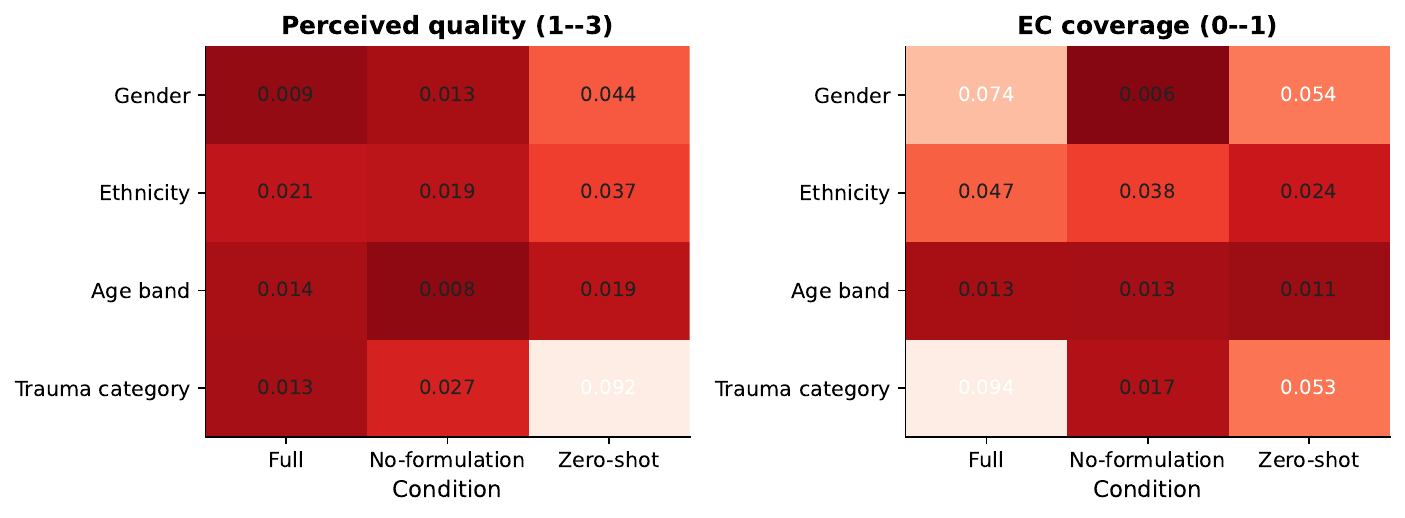}
    \caption{Statistical Parity Difference (SPD, max-minus-min group mean) for each demographic by condition, separately for perceived quality (left, $1$--$3$ scale) and EC coverage (right, $0$--$1$ scale). Cell colour and number give the SPD. Smaller is better. Quality SPD shrinks $1.5$--$7\times$ from zero-shot to full on every demographic. EC SPD is small overall except on trauma category, where it widens slightly under full because of the content-fit effect.}
    \label{fig:fairness-spd-heatmap}
\end{figure*}

\begin{figure*}[tbp]
    \centering
    \includegraphics[width=0.85\textwidth]{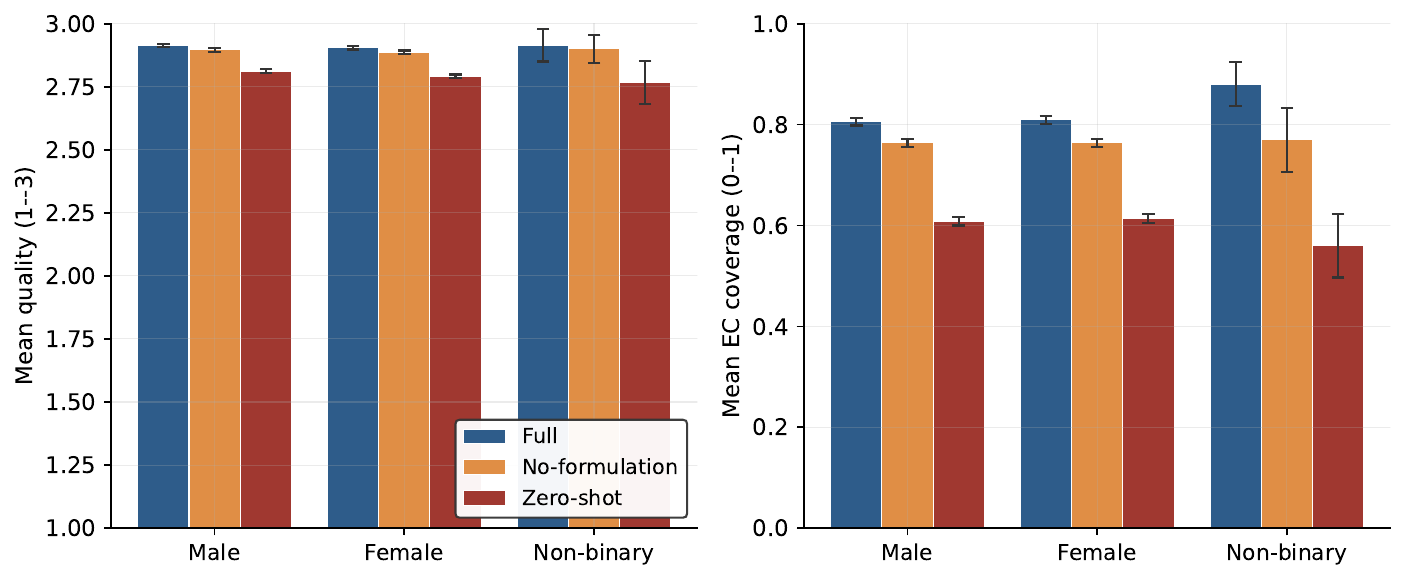}
    \caption{Mean \texttt{deepseek-v4-flash} quality (left) and EC coverage (right) by persona gender ($n = 258$ Male / $238$ Female / $4$ Non-binary), grouped by condition. Bars: group mean; whiskers: $95\%$ Wald CI. Within each panel the three condition-groups are nearly indistinguishable across genders (small SPD), and the condition gradient is preserved at every gender level.}
    \label{fig:fairness-gender}
\end{figure*}

\begin{figure*}[tbp]
    \centering
    \includegraphics[width=0.95\textwidth]{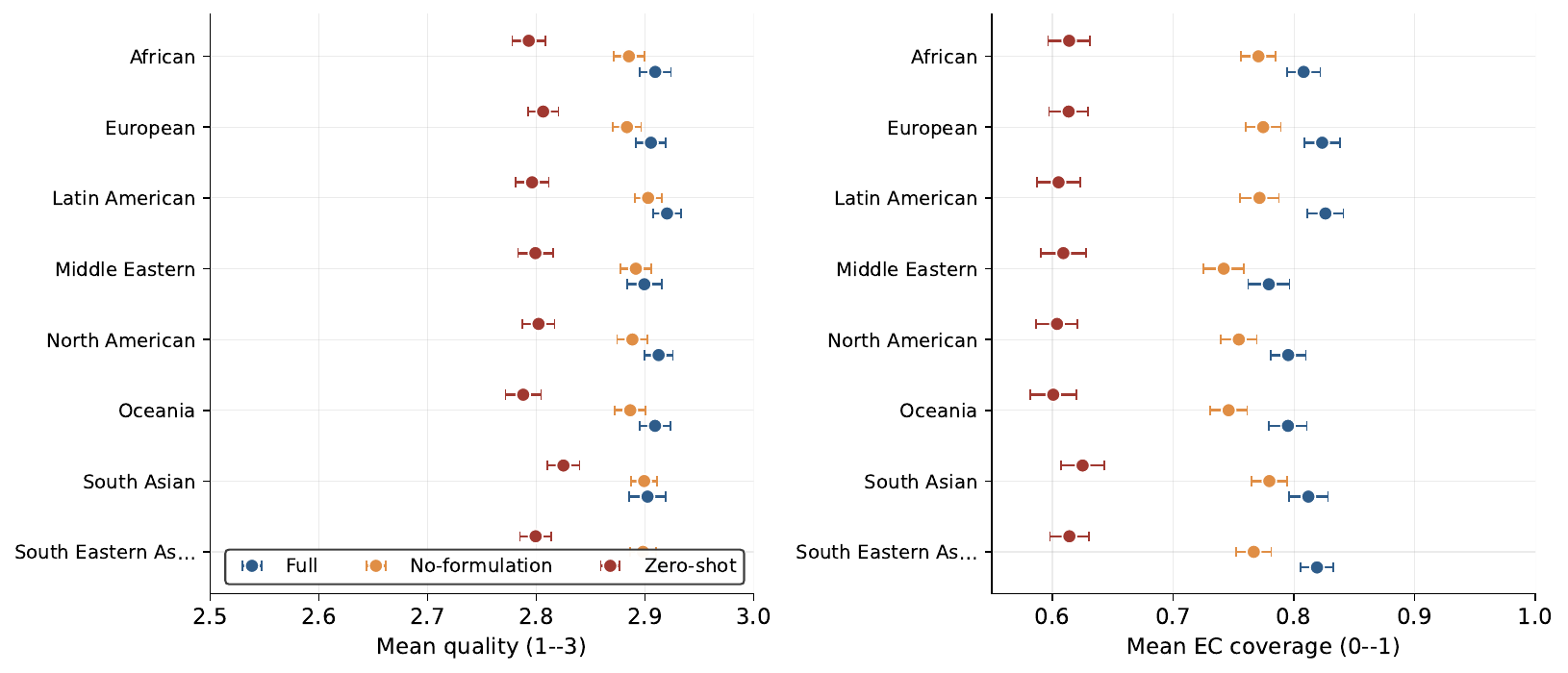}
    \caption{Mean \texttt{deepseek-v4-flash} quality (left) and EC coverage (right) by persona ethnicity (8 world regions, $n \approx 53$--$71$ per category), shown as a dot-strip with $95\%$ Wald CIs. Each demographic level has three condition markers vertically offset. The condition markers are tightly aligned within each ethnicity (small SPD); the absolute level under each condition is constant across ethnicities.}
    \label{fig:fairness-ethnicity}
\end{figure*}

\begin{figure*}[tbp]
    \centering
    \includegraphics[width=0.85\textwidth]{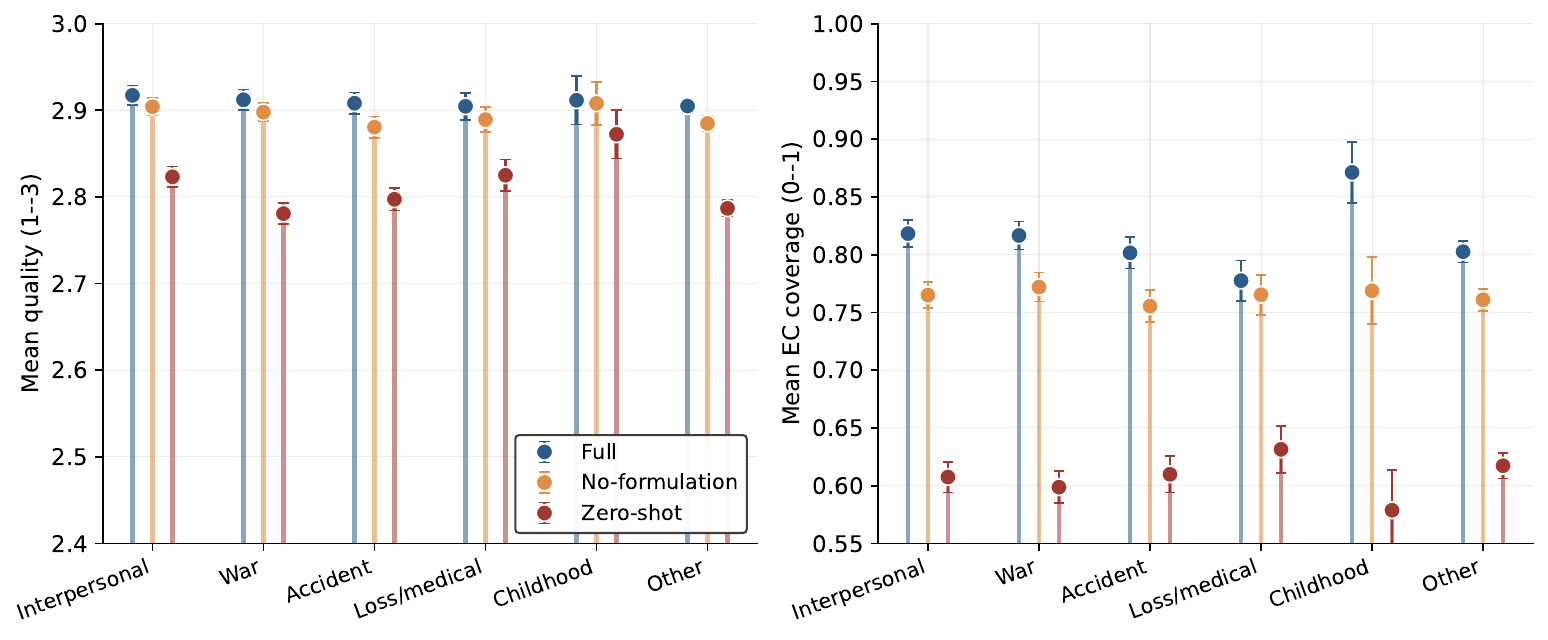}
    \caption{Mean \texttt{deepseek-v4-flash} quality (left) and EC coverage (right) by persona trauma category, shown as a lollipop chart with $95\%$ Wald CIs (markers) and stems running to the axis floor. EC coverage shows a residual content-fit disparity on trauma category: childhood-trauma narratives engage all five EC components more naturally than loss/medical narratives, so the full-condition EC marker spread is wider than in other panels.}
    \label{fig:fairness-trauma}
\end{figure*}

\label{app:fig-fairness}
The demographic-by-condition breakdown of mean \texttt{deepseek-v4-flash} quality and EC coverage on the $15{,}000$-vignette scaled-judge set is visualised in four complementary figures. Figure~\ref{fig:fairness-spd-heatmap} is the headline view: a heatmap of Statistical Parity Difference (max-minus-min group mean) for each (demographic, condition) cell, separately for quality and for EC coverage. The cell-by-cell shrinkage from zero-shot to full is the SPD collapse the body claim refers to. Figures~\ref{fig:fairness-gender}, \ref{fig:fairness-ethnicity}, and \ref{fig:fairness-trauma} then show the underlying group-level means for each demographic with different chart types matched to category cardinality (paired bars with $95\%$ CIs for 3-level gender; horizontal dot-strip with $95\%$ CIs for 8-level ethnicity; lollipop chart for 6-level trauma category). Across all three views, two patterns hold: (a) within each (demographic, condition) cell, group means are nearly flat (visible as compressed bars/strips), consistent with the omnibus tests in Table~\ref{tab:fairness-omnibus}; (b) within each demographic level, the full condition consistently sits above no-formulation, which sits above zero-shot, the same condition gradient reported throughout the paper. The exception is the trauma-category EC panel, where the full markers are visibly more spread out than the zero-shot markers: this is the content-fit effect (childhood-trauma narratives engage all five EC components more naturally than loss/medical narratives), discussed at length in Appendix~\ref{app:fairness-metrics}.

% ===== Robustness =====

\label{app:fairness-omnibus}
Table~\ref{tab:fairness-omnibus} reports the omnibus tests that anchor the fairness analysis in \S\ref{sec:fairness}. For the scaled-judge set ($n = 15{,}000$ vignettes, $500$ personas $\times$ $10$ retained models $\times$ $3$ conditions) the test is an $F$-test from an OLS regression of the outcome on the demographic factor, with cluster-robust standard errors clustered at the persona level; this accounts for the fact that the same persona is rated $30$ times (10 models $\times$ 3 conditions) and that within-persona ratings are correlated. The same model is fit separately for the two outcomes (mean \texttt{deepseek-v4-flash} quality and mean EC coverage) so that the demographic effects can be interpreted independently.

For the clinician user-study set ($n = 1{,}600$ ratings, $100$ participants) the cluster is participant. Each participant rates a mixed set of (model, condition, persona) cells, so the relevant within-cluster correlation is the participant's baseline rating tendency rather than the persona's rated properties. We do not cluster on persona in the user-study fit because the persona is randomised within participant and there are $\sim 300$ unique personas in the rated subsample with no participant rating more than two.

The key findings are: (a) EC coverage is invariant across every demographic dimension on both the scaled-judge set and the user study (all $p > 0.13$), so the framework's structural-content output does not vary systematically with persona gender, ethnicity, age, occupation, or trauma category; (b) on the scaled-judge set, perceived quality varies with trauma category ($F(5,499) = 11.6$, $p < .001$), gender ($F(2,499) = 8.1$, $p < .001$), and occupation ($F(5,499) = 2.6$, $p = .023$), but the corresponding effect sizes in Appendix~\ref{app:fairness-metrics} are small ($d_{\max} \leq 0.49$ on trauma, $\leq 0.11$ on gender); (c) the clinician user study shows no detectable demographic effect on either quality or AI authorship (all $p > 0.18$, smallest at $p = .054$ for AI authorship $\times$ occupation), so any demographic disparity that does appear in the scaled-judge ratings does not transfer to clinician perception.

We emphasise that significance at $p < .05$ on the scaled-judge tests does not establish a fairness violation: with $n = 15{,}000$ the test detects very small mean differences. The interpretive weight should be on the effect-size and on the four-fifths-rule analyses in Appendix~\ref{app:fairness-metrics}, where the trauma-category $d_{\max}$ falls from $-0.49$ in zero-shot to $+0.07$ in full and the DIR rises from $0.97$ to $0.99$.

\begin{table}[h]
\centering
\small
\setlength{\tabcolsep}{4pt}
\caption{Demographic omnibus $F$-tests on the scaled judge ($n = 15{,}000$, cluster-robust SEs at persona) and on the clinician user study ($n = 1{,}600$, cluster-robust SEs at participant). EC coverage is invariant across all demographics; only trauma category significantly affects scaled-judge quality, with a small effect size.}
\label{tab:fairness-omnibus}
\resizebox{\linewidth}{!}{%
\begin{tabular}{ll rrr}
\toprule
\textbf{Outcome} & \textbf{Demographic} & $\bm{F}$ & \textbf{df} & $\bm{p}$ \\
\midrule
\multicolumn{5}{l}{\textit{Scaled judge (\texttt{deepseek-v4-flash}) on the $15{,}000$ generation set}} \\
Quality & Gender         & $8.12$  & (2, 499) & $< .001^{***}$ \\
Quality & Ethnicity      & $1.28$  & (7, 499) & $.256$ \\
Quality & Age band       & $0.95$  & (3, 499) & $.417$ \\
Quality & Occupation     & $2.63$  & (5, 499) & $.023^{*}$ \\
Quality & Trauma category & $11.61$ & (5, 499) & $< .001^{***}$ \\
EC      & Gender         & $0.19$  & (2, 499) & $.829$ \\
EC      & Ethnicity      & $1.31$  & (7, 499) & $.245$ \\
EC      & Age band       & $0.29$  & (3, 499) & $.831$ \\
EC      & Occupation     & $1.71$  & (5, 499) & $.131$ \\
EC      & Trauma category & $0.25$ & (5, 499) & $.941$ \\
\midrule
\multicolumn{5}{l}{\textit{Clinician user study ($n = 1{,}600$ ratings)}} \\
Quality & Gender         & $1.51$  & (2, 99)  & $.226$ \\
Quality & Ethnicity      & $0.84$  & (7, 99)  & $.553$ \\
Quality & Age band       & $0.36$  & (3, 99)  & $.783$ \\
Quality & Occupation     & $0.45$  & (5, 99)  & $.811$ \\
Quality & Trauma category & $0.76$ & (5, 99)  & $.579$ \\
AI authorship & Gender   & $1.28$  & (2, 99)  & $.283$ \\
AI authorship & Ethnicity & $1.48$ & (7, 99)  & $.185$ \\
AI authorship & Age band & $0.24$  & (3, 99)  & $.866$ \\
AI authorship & Occupation & $2.26$ & (5, 99) & $.054$ \\
AI authorship & Trauma category & $0.29$ & (5, 99) & $.915$ \\
\bottomrule
\end{tabular}
}
\end{table}
\end{document}